%% file: main.tex
\documentclass[sigconf, nonacm]{acmart}

\newcommand\vldbdoi{10.14778/3849423.3849436}
\newcommand\vldbpages{229 - 242}
\newcommand\vldbvolume{20}
\newcommand\vldbissue{2}
\newcommand\vldbyear{2026}
\newcommand\vldbauthors{\authors}
 
\newcommand\vldbavailabilityurl{https://github.com/IBM/table-representation-evals}
\newcommand\vldbpagestyle{plain} 

\usepackage{listings}
\usepackage{xfrac}
\usepackage{tabularx}
\usepackage{caption}
\usepackage{subcaption}
\usepackage{csquotes}
\usepackage{bm}
\usepackage[para]{threeparttable}
\usepackage{enumitem}
\usepackage{hyperref}
\usepackage{multirow}
\usepackage{array}
\usepackage{makecell}
\usepackage{siunitx}
\usepackage{graphicx}
\usepackage{pifont}
\usepackage{placeins}
\usepackage[table]{xcolor}

\begin{document}

\newcommand{\todo}[1]{\textcolor{red}{#1}}
\newcommand{\revised}[1]{\textcolor{blue}{#1}}
\newcommand\benchmark{TEmBed}

\newcommand{\cmark}{\textcolor{green}{\ding{51}}} 
\newcommand{\xmark}{\textcolor{red}{\ding{55}}}   

\title{Towards Universal Tabular Embeddings: \\ A Benchmark Across Data Tasks}
\subtitle{With Extended Technical Report}
\author{Liane Vogel}
\affiliation{%
  \institution{Technical University of Darmstadt}
  \country{Germany}
}
\authornote{
Correspondence goes to \href{mailto:liane.vogel@tu-darmstadt.de}{liane.vogel@tu-darmstadt.de}}

\author{Kavitha Srinivas}
\affiliation{%
  \institution{IBM Research}
  \country{USA}
}

\author{Niharika D'Souza}
\affiliation{%
  \institution{IBM Research}
  \country{USA}
}

\author{Sola Shirai}
\affiliation{%
  \institution{IBM Research}
  \country{USA}
}

\author{Oktie Hassanzadeh}
\affiliation{%
  \institution{IBM Research}
  \country{USA}
}

\author{Horst Samulowitz}
\affiliation{%
  \institution{IBM Research}
  \country{USA}
}

\newcommand{\notesize}{\fontsize{7.5}{9}\selectfont}
\renewcommand\theadfont{\bfseries\notesize}
\renewcommand{\paragraph}[1]{\vspace{0.8ex}\noindent\textbf{#1}$\:$}
\newcommand{\newparagraph}{\vspace{0.8ex}\noindent}
\newcommand{\newparagraphwithindent}{\vspace{0.8ex}}

\begin{abstract}
Tabular foundation models aim to learn universal representations of tabular data that transfer across tasks and domains, enabling applications such as  table retrieval, semantic search and table-based prediction.
Despite the growing number of such models, it remains unclear which approach works best in practice, as existing methods are often evaluated under task-specific settings that make direct comparison difficult.
To address this, we introduce \benchmark, the \emph{\underline{T}abular \underline{E}mbedding Test \underline{B}ed}, a unified benchmark for systematically evaluating tabular embeddings across four representation levels: cell, row, column, and table.
Evaluating a diverse set of tabular representation learning models, we show that which model to use depends on the task and representation level.
Our results offer practical guidance for selecting tabular embeddings in real-world applications and lay the groundwork for developing more general-purpose tabular representation models.
\end{abstract}

\maketitle

\pagestyle{\vldbpagestyle}
\begingroup\small\noindent\raggedright\textbf{PVLDB Reference Format:}\\
\vldbauthors. Towards Universal Tabular Embeddings: A Benchmark Across Data Tasks. PVLDB, \vldbvolume(\vldbissue): \vldbpages, \\ \vldbyear.
\href{https://doi.org/\vldbdoi}{doi:\vldbdoi}
\endgroup
\begingroup
\renewcommand\thefootnote{}\footnote{\noindent
This work is licensed under the Creative Commons BY-NC-ND 4.0 International License. Visit \url{https://creativecommons.org/licenses/by-nc-nd/4.0/} to view a copy of this license. For any use beyond those covered by this license, obtain permission by emailing \href{mailto:info@vldb.org}{info@vldb.org}. Copyright is held by the owner/author(s). Publication rights licensed to the VLDB Endowment. \\
\raggedright Proceedings of the VLDB Endowment, Vol. \vldbvolume, No. \vldbissue\ %
ISSN 2150-8097. \\
\href{https://doi.org/\vldbdoi}{doi:\vldbdoi} \\
}\addtocounter{footnote}{-1}\endgroup

\ifdefempty{\vldbavailabilityurl}{}{
\vspace{.3cm}
\begingroup\small\noindent\raggedright\textbf{PVLDB Artifact Availability:}\\
The source code, data, and/or other artifacts have been made available at \url{\vldbavailabilityurl}.
\endgroup
}

\input{sections_01_introduction}

\input{sections_02_background}

\input{sections_03_benchmark_overview}
\input{sections_04_row_embeddings}

\input{sections_05_column_embeddings}
\input{sections_06_table_embeddings}

\input{sections_07_cell_embeddings}
\input{sections_08_discussion}
\input{sections_09_conclusion}

\begin{acks}
This work has been supported by the BMBF and the state of Hesse as part of the NHR program. 
We thank Jan-Micha Bodensohn for his support.
\end{acks}

\balance{}
\newpage

\bibliographystyle{ACM-Reference-Format}
\bibliography{references}

\newpage
\nobalance

\input{sections_10_appendix}
\end{document}

%% file: sections_01_introduction.tex
\section{Introduction}
\label{sec:introduction}

\begin{figure}[t]
    \centering
    \includegraphics[width=0.8\linewidth]{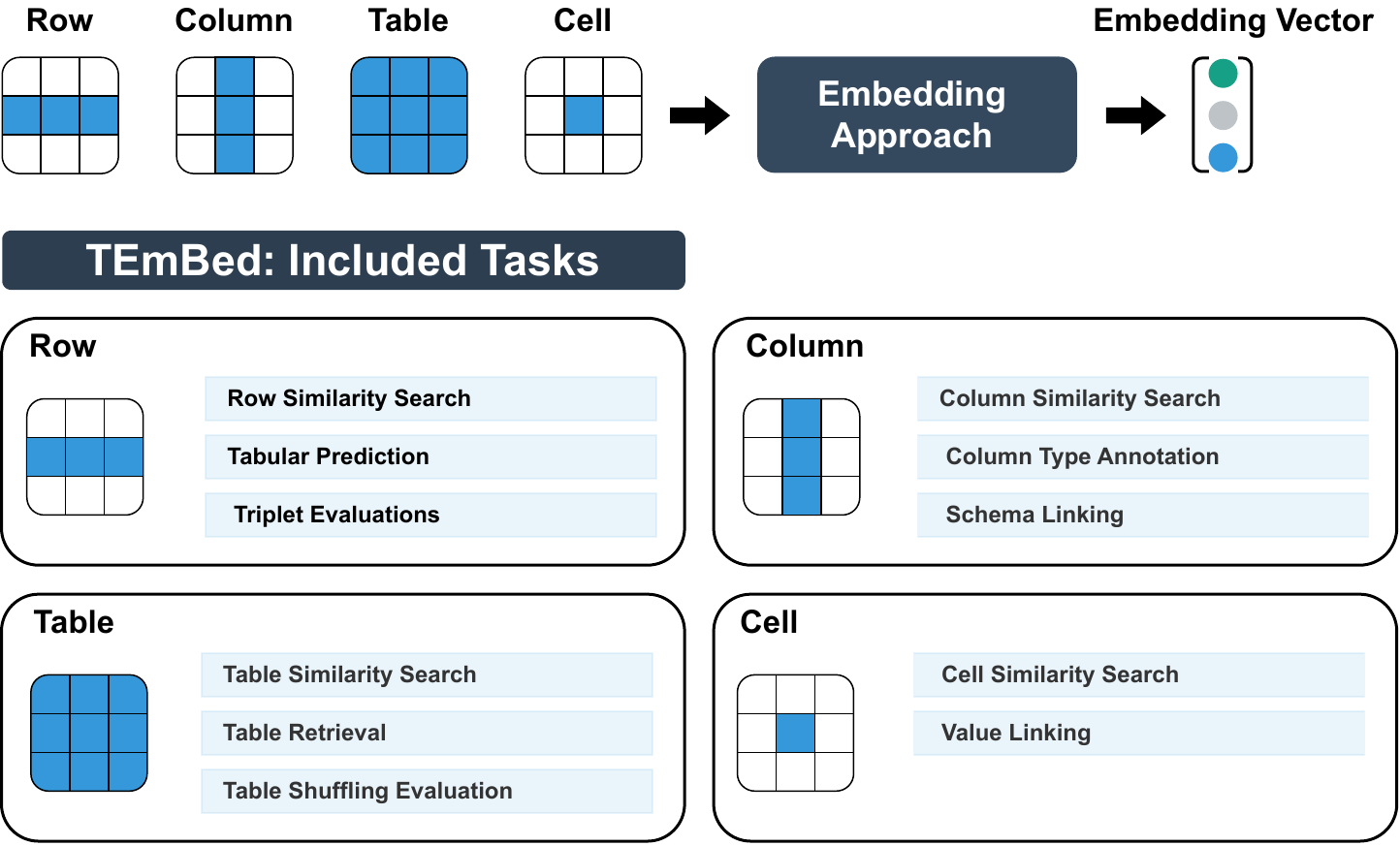}
    \caption{Overview of \benchmark. A tabular embedding approach maps a table to representations at one or more of four representation levels: cell, row, column and table. We systematically evaluate tabular embeddings on a diverse set of tasks across 87 datasets and benchmark 13 approaches.}
    \label{fig:header}
    \vspace{-2em}
\end{figure}

\newparagraphwithindent \textbf{Representation Learning on Relational Data.}
Tabular data is a core data modality in database systems and remains the dominant format for structured data management in practice. 
Consequently, representation learning for relational and tabular data has become an important research direction at the intersection of databases and machine learning. 
The goal is to embed tables in a way that captures semantic, structural, and statistical properties such that the resulting embeddings are reusable for downstream applications.
As shown in Figure \ref{fig:header}, tabular embeddings must capture four different levels of structural information, each of which is useful for different tasks; e.g., duplicate detection requires row-level embeddings, while table discovery requires table-level embeddings. 

\newparagraphwithindent \textbf{Tabular Foundation Models.}
Recently, a growing number of table foundation models \cite{qu2025tabicl, hollmann2025tabpfn, chen2023hytrel} have been introduced with the goal of learning universal representations of tabular data that transfer across tasks, domains, and datasets without requiring additional training.
These models leverage large-scale pretraining on heterogeneous data sources, enabling them to produce cell-, row-, column-, or table-level embeddings for any new table.

\newparagraphwithindent
Similar developments in natural language processing have been accompanied by standardized evaluation frameworks, most notably the Massive Text Embedding Benchmark (MTEB) \cite{muennighoff2023mteb}, which enables systematic comparison of general-purpose text embedding models across diverse tasks and is the de facto standard for the field.
However, \textbf{no analogous benchmark exists for tabular embeddings}. As a result, existing evaluations are narrow, task-specific, and use different embedding levels with incompatible protocols, making it difficult to comprehensively compare models \cite{badaro2023trl_survey}.

\newparagraphwithindent
Assessing the semantic and structural quality of learned embeddings is inherently challenging, as it often requires ground truth signals that vary across tasks and domains. 
For tabular models, this challenge is further complicated by the different granularity levels of embeddings required for tables, which may need fundamentally distinct evaluation criteria: a good table-level embedding must capture global schema and content, while a good cell-level embedding must reflect local value semantics in context. 

\newparagraphwithindent \textbf{Contributions.}
As our core contribution, we introduce \textbf{\benchmark}, a unified benchmarking framework for tabular embeddings.
Our framework evaluates models across cell-, row-, column-, and table-level tasks using a diverse collection of datasets, including some newly developed benchmarks. Further, \benchmark~ provides an open-source toolkit to easily extend both the supported models and benchmark datasets, enabling consistent measurements of model performance, time efficiency, and resource consumption across diverse tasks. 
By providing a common evaluation framework, \benchmark~ enables meaningful cross-task comparison, supports informed model selection, and establishes a foundation for developing more general-purpose tabular representation models.

Our main contributions are: 
\begin{enumerate}
    \item We introduce the first unified benchmarking framework for tabular embeddings that spans multiple datasets, tasks, and embedding levels (cell, row, column, table). We incorporate existing public datasets and complete this selection by introducing multiple new datasets.
    \item We provide an extensible open-source test bed with efficiency measurements that works across a large diversity of different embedding models. 
    \item We perform a comprehensive empirical study of state-of-the-art tabular embedding models, assessing their performance, computational cost, generalization across tasks, and offer practical guidance for model selection.
\end{enumerate}

%% file: sections_02_background.tex
\section{Background}
\label{sec:background}

\subsection{Tabular Data Representations}
\label{sec:background_tabular_data_representations}

\newparagraphwithindent \textbf{Table Encoders.}
Learning representations of tabular data has been an active research area over the past years, as surveyed by \citet{badaro2023trl_survey}, yet no comprehensive comparison across approaches exists due to the lack of a unified benchmark.
Broadly, two design philosophies have emerged. 
The first develops tabular-specific architectures and pretraining objectives \cite{deng2020turl, iida2021tabbie, yin2020tabert}, often targeting specific downstream tasks, such as table question answering \cite{herzig2020tapas} or column type annotation \cite{suhara2022doduo}.
More recent models such as HyTrel \cite{chen2023hytrel} take a more universal perspective, modeling tables as hypergraphs to produce representations at all four granularity levels we consider - cells, rows, columns and tables.
The second philosophy foregoes tabular-specific pretraining altogether: table content is serialized into a flat text sequence, typically by concatenating column headers and cell values, and passed to a general-purpose text embedding model \cite{ji2025target_benchmark, balaka2025pneuma}. 
We evaluate representative models from both philosophies on \benchmark, as they differ in how directly the evaluated tasks match their original training objective.

\newparagraphwithindent \textbf{Tabular Prediction.}
A distinct line of work focuses on tabular prediction: predicting a target column from the remaining columns of a table, analogous to classical supervised learning on structured data.
Traditionally dominated by gradient-boosted tree methods~\cite{chen2016xgboost}, the field is increasingly challenged by foundation models such as TabPFN \cite{hollmann2025tabpfn} and TabICL \cite{qu2025tabicl}, which perform in-context learning on new tables without gradient updates.
Other approaches fine-tune large language models on serialized tables \cite{gardner2024tabula} or incorporate semantic signals via pretrained text or graph-based encoders \cite{spinaci2025contexttab, kim2024carte, kim2025tarte}.
Although optimized for prediction, these models produce intermediate row-level representations that can in principle serve as general-purpose embeddings, which we investigate by including multiple tabular prediction models in our evaluations.

\newparagraphwithindent \textbf{LLMs for Generative Table Reasoning.}
A parallel line of work adapts LLMs to tabular data for generative and conversational reasoning tasks. 
Models such as TableLlama \cite{zhang2024tablellama} and Table-GPT~\cite{li2024table-gpt} are fine-tuned for generative, prompt-driven use cases like question answering, SQL generation and summarization, making them complementary to the task-agnostic embeddings we evaluate.

\subsection{Benchmarking Table Foundation Models}
\newparagraphwithindent \textbf{Existing task-specific benchmarks.}
Existing benchmarks for tabular data are typically task specific, rather than spanning multiple tasks and representation levels, a gap that \benchmark~ aims to fill. 
The TARGET benchmark \cite{ji2025target_benchmark} focuses on table retrieval for generative tasks such as TableQA and Text-to-SQL.
For data discovery tasks over data lakes, LakeBench \cite{srinivas2023lakebench} provides benchmarks for Table Unionability and Joinability search. 
In the field of tabular prediction, frameworks such as TabArena \cite{erickson2025tabarena}, TALENT \cite{liu2025talent}, and TabReD \cite{rubachev2025tabred} offer extensive model comparisons, but remain limited to predictive tasks. 
In contrast, our benchmark spans all four embedding levels and both retrieval and prediction tasks.

\newparagraphwithindent \textbf{Broader Embedding Evaluations.}
\citet{hoppe2025tabembedbench} benchmark task-agnostic embeddings from tabular foundation models on outlier detection and tabular prediction, but consider only row-level embeddings and a small set of models.
Closest to our work, the concurrent TRL-Bench \cite{pang2026trlbench} probes row-, column-, and table-level embeddings with trained heads, whereas \benchmark~also covers cells and evaluates embeddings directly via similarity search.

\newparagraphwithindent \textbf{Analytical Studies.}
Other efforts analyze embedding properties directly rather than downstream task performance, making them orthogonal to our application-oriented benchmark.
Observatory \cite{cong2023observatory} measures properties like sensitivity to column order and sample fidelity, while TabEE \cite{copul2024tabee} clusters embeddings and generates explanations for the resulting groups. 
In contrast, our benchmark measures embedding quality through performance on concrete tasks with well-defined ground truth labels.

\newparagraphwithindent \textbf{Evaluating LLMs on Table Tasks.}
A related line of benchmarks evaluates LLM capabilities on tables directly, rather than the quality of learned representations.
MMTU \cite{xing2025mmtu} spans dozens of tasks such as entity matching and table transformations, while \cite{sui2024meetsllm} evaluates structural understanding tasks like cell lookup and row retrieval. 
Some tasks, like semantic-join detection, overlap with \benchmark. 
Still, embeddings and conversational LLMs serve fundamentally different use cases, making the two approaches complementary.

\newparagraphwithindent
Overall, systematic evaluation frameworks analogous to text embedding benchmarks such as MTEB \cite{muennighoff2023mteb} are still missing for tabular embeddings. 
\benchmark~ addresses this gap with a unified benchmark spanning multiple embedding levels and tasks, giving practitioners concrete guidance for selecting embedding models in real-world applications.

%% file: sections_03_benchmark_overview.tex
\begin{table*}[t]
\centering
\caption{Overview of tasks in the tabular embedding benchmark. * indicates that we adapted the datasets for our use cases.}
\small
\begin{tabular}{l|l|p{8cm}|r|p{1cm}} 
\toprule
Embedding Level & Task Name & Task Description & \# Datasets & Source \\ 
\midrule
\multirow{3}{*}{Row} 
& (1) Row Similarity Search & Given a query row, retrieve the top-k most similar rows. & 9  & \cite{koepcke2010entityresolution, mudgal2018deepmatcher}*\\ 
& (2) Triplet-Based Evaluation & Given a triplet of rows (anchor, positive, negative), evaluate whether the anchor is more similar to the positive than the negative. & 2 & ours \\ 
& (3) Tabular Prediction & Use row embeddings as features for supervised classification or regression tasks. & 51  & \cite{erickson2025tabarena} \\ 
\midrule
\multirow{3}{*}{Column}  & (4) Column Similarity Search & Given a query column, retrieve the top-k most similar columns from a data lake.  & 4 & \cite{flores2021nextia, srinivas2023lakebench} \\ 
& (5) Column Type Annotation & Use column embeddings as features to train a classifier that predicts each column's semantic type. & 2 & \cite{korini2022sotab, hulsebos2023gittables}\\
& (6) Schema Linking & Given a natural language query, retrieve the top-k database columns relevant to answering it. & 1 & \cite{li2023bird}*\\
\midrule
Table
& (7) Table Similarity Search & Given a query table, retrieve the top-k most similar tables from a collection. & 1 & \cite{hulsebos2023gittables}*\\ 
&  (8) Table Retrieval & Given a natural language query, retrieve the top-k most relevant tables. & 7  & \cite{ji2025target_benchmark} \\ 
& (9) Table Shuffling Evaluation & Given a triplet of tables (anchor, permuted positive, value-shuffled negative), evaluate whether the anchor is closer to the positive. & 6 & \cite{ji2025target_benchmark, srinivas2023lakebench}* \\
\midrule
Cell & (10) Cell Similarity Search & Given a query cell, retrieve the top-k most semantically similar cells across a set of tables. & 2 & \cite{lou2023s2abel}* \\ 
& (11) Value Linking & Given a natural language query mentioning a value, retrieve the top-k database columns whose cell values match it, either verbatim or via a fuzzy/semantic variant. & 2 & \cite{li2023bird}* \\
\bottomrule
\end{tabular}

\label{tab:task_overview}
\end{table*}

\section{Benchmark Overview}
\label{sec:overview}
With \benchmark~ we introduce a framework to systematically evaluate tabular embedding approaches across a diverse range of tasks and datasets. 
Recognizing that no benchmark can ever be fully comprehensive, we design the framework to be easily extensible, allowing new approaches, tasks and datasets to be integrated with minimal effort, as described in the technical report.
A key design dimension of \benchmark~ is to cover all levels of granularity at which embeddings are computed: row-, column-, cell-, and table-level. 
Not all approaches are able to compute embeddings for every level, as summarized in Table~\ref{tab:approaches_overview}.

\newparagraphwithindent \textbf{Tasks and Datasets.}
\benchmark~ currently includes eleven tasks across a total of 87 datasets, as listed in Table \ref{tab:task_overview}. 
All datasets are publicly available: either constructed by us, with code released alongside the benchmark, or downloaded from their respective repositories.
Evaluation is performed using deterministic test cases, reporting common task-specific metrics.
The extensible design of our framework makes it simple to add further datasets and tasks.
We describe the toolkit's extension interface in the technical report.

\newparagraphwithindent \textbf{Configuration and Reproducibility.}
New approaches can be added by implementing our provided interfaces, enabling consistent evaluation across representation levels. 
Hyperparameters can be set via configuration files.
For our experiments, we use each approach's default configuration as stated in its paper and release all configuration files for reproducibility. 

\newparagraphwithindent \textbf{Performance and Resource Metrics.}
In addition to task performance, our framework tracks runtime of the approaches and resource consumption (CPU, RAM, GPU memory), providing insights into the efficiency and scalability of different approaches, particularly on large datasets. 
All experiments were run on a machine with a single NVIDIA A100 GPU (80~GB VRAM). We additionally report measurements on more constrained hardware (a 16~GB V100 and CPU-only) in the technical report.

\subsection{Included Tasks}
\label{sec:overview_tasks}

\benchmark~ currently includes the following eleven tasks, grouped by representation level: 

\newparagraphwithindent \textbf{Row Level Evaluations.}
Retrieving the most similar rows in a table is a critical task in many real-world applications, such as recommending products in e-commerce or detecting duplicate data entries.
We evaluate this in \textbf{Row Similarity Search}, built from nine entity-matching and clustering datasets. 
To complement the row level evaluations with a more systematic, controlled measure of embedding quality, we include \textbf{Triplet-Based Row Evaluation}, which tests whether row embeddings capture meaningful semantic relationships. 
For example, a smartphone and a laptop (both electronics) should be judged more similar than a smartphone and a jacket.
We construct two new datasets leveraging Wikidata~\cite{wikidata}, which allows us to automatically generate triplets covering a wide range of semantic distances.
Finally, since row embeddings may also be used as features for downstream models, we evaluate them with  \textbf{Tabular Prediction}, using them as input features to classifiers and regressors for decision-making tasks such as risk assessment and credit scoring, using the TabArena \cite{erickson2025tabarena} benchmark. 

\newparagraphwithindent \textbf{Column Level Evaluations.}
Identifying semantically similar columns is the basis for determining joinability and unionability of tables and for performing schema matching between databases, for example, matching a \emph{city} column to a \emph{region} column from an external source.
We evaluate this with \textbf{Column Similarity Search}, using four existing join-discovery datasets. 
Beyond finding related columns, data cataloging and data integration pipelines often need to infer a column's semantic type (e.g. \emph{currency}) when it arrives without reliable metadata. 
We address this with \textbf{Column Type Annotation}, evaluating column embeddings as classifier features for this task. 
We target a third, more query-driven use case with \textbf{Schema Linking}: mapping a natural-language question to the relevant database columns, a core subtask of text-to-SQL, evaluated on a dataset built from BIRD \cite{li2023bird}.

\newparagraphwithindent \textbf{Table Level Evaluations.}
Table embeddings support semantic search over data repositories, such as retrieving all customer transaction tables belonging to the same source. 
We evaluate this with \textbf{Table Similarity Search}, using a new dataset built from GitTables \cite{hulsebos2023gittables}, testing whether embeddings identify tables from the same GitHub repository.
In addition, we evaluate \textbf{Table Retrieval}, locating tables relevant to a natural-language query, a major component in many RAG pipelines. 
Finally, a robust table embedding should be invariant to row/column reordering while remaining sensitive to changes that break the table's meaning \cite{cong2023observatory}. 
We test this with \textbf{Table Shuffling Evaluation} using triplets of an anchor table, a row/column-permuted positive, and a value-shuffled negative.

\newparagraphwithindent \textbf{Cell Level Evaluations.}
Cell embeddings are useful whenever fine-grained semantic information must be captured at the level of individual entries, e.g., identifying all mentions of the same real-world entity, such as \textit{NYC} and \textit{New York City}. 
We test this with \textbf{Cell Similarity Search} on a new dataset built from S2abEL \cite{lou2023s2abel}.
A related challenge arises when translating natural-language questions into database queries: values mentioned in the question must be linked to the actual cell values they refer to, despite synonyms or formatting differences. 
We cover this with \textbf{Value Linking}, in an exact variant, where the mentioned value appears verbatim in the database, and a fuzzy variant, where it does not, both built from the BIRD benchmark \cite{li2023bird}.

\smallskip\noindent Together, these eleven tasks give \benchmark~ coverage of realistic, application-grounded use cases at every representation level.
Our experiments demonstrate that current embedding approaches are not universally applicable, and that model choice depends on the specific task and dataset.
This underscores the importance of carefully selecting a suitable model, and highlights the ongoing need for research into more robust and adaptable embedding models.


\subsection{Included approaches}
\label{sec:overview_approaches}
Although many tabular embedding approaches would be relevant candidates to include in our evaluation, our selection is necessarily limited. 
Our main model inclusion criterion is that approaches must produce tabular embeddings out of the box, without requiring task-specific fine-tuning, and must support multiple of the benchmarked tasks.
This keeps the comparison focused on general-purpose, reusable embeddings rather than task-specialized pipelines.
We have selected a diverse set of models, including specialized approaches for tabular data understanding, models designed for tabular prediction, and general-purpose text embedding models.
The design of our framework makes it simple to add further approaches in the future, ensuring extensibility as new models emerge.
Table \ref{tab:approaches_overview} summarizes which of the models provide embeddings at which level.
We give an overview of each category below and provide detailed per-model descriptions, including hyperparameters and embedding-extraction details, in the technical report.

\begin{table}
     \centering
     \caption{Overview of the embedding approaches in our evaluation with the embedding levels they provide, grouped into table-specific representation models, tabular in-context learners, and general-purpose text embedding models. We fit the in-context learning models (apart from TabuLa-8B) to placeholder labels before extracting embeddings.}
     \footnotesize
     \begin{tabularx}{\columnwidth}{lrrcccc}
         \toprule
         Approach & Dim & Params & Cell & Row & Column & Table   \\
         \midrule
         HyTrel  \cite{chen2023hytrel} & 768 & 179M & \cmark  & \cmark & \cmark & \cmark \\
         TABBIE \cite{iida2021tabbie} & 768 & 340.4M & \cmark  & \cmark & \cmark & \cmark \\
         TaBERT \cite{yin2020tabert} & 768 & 133.5M & \cmark  & \cmark & \cmark & \cmark \\
         TUTA \cite{wang2021tuta} & 768 & 133.9M &  \cmark  &  \cmark  & \cmark &  \cmark  \\
         \midrule
         TabuLa-8B  \cite{gardner2024tabula} & 4096 & 8B & \xmark  & \cmark & \xmark &  \xmark \\
         SAP-RPT-1  \cite{spinaci2025contexttab} & 768 & 172M & \cmark & \cmark & \cmark & \xmark \\ 
         TabPFN v2.5  \cite{hollmann2025tabpfn} & 192 & 7M  & \xmark & \cmark & \xmark & \xmark \\
         TabICL v2 \cite{qu2025tabicl} & 512 & 28M & \cmark & \cmark & \cmark & \xmark \\
         TabDPT \cite{ma2025tabdpt} & 512 & 63.5M & \xmark  & \cmark & \xmark  & \cmark \\
         TARTE \cite{kim2025tarte} & 768 & 20.8M & \cmark  & \cmark & \cmark  & \cmark \\
         \midrule
         All-MiniLM-L6-v2 \cite{wang2020minilm} & 384 & 23M & \cmark  & \cmark & \cmark & \cmark \\
         IBM Granite R2 \cite{awasthy2025granite-r2}  & 768 & 149M & \cmark & \cmark  & \cmark & \cmark \\
         GritLM \cite{muennighoff2024gritlm} & 4096 & 7B & \cmark & \cmark & \cmark & \cmark \\  
         \bottomrule
     \end{tabularx}

     \label{tab:approaches_overview}
\end{table}

\newparagraphwithindent \textbf{Tabular representation learning models:}
These approaches are pretrained specifically to learn general-purpose representation of tables.
We include four models: \textbf{HyTrel} \cite{chen2023hytrel}, which models tables as hypergraphs, \textbf{TaBERT} \cite{yin2020tabert} and \textbf{TABBIE} \cite{iida2021tabbie} which pretrain transformers jointly over cell, row and column context, and \textbf{TUTA}~\cite{wang2021tuta}, which adds tree-based positional encodings to capture cell layout. 
Together, these models represent the main architectural families used in table-specific representation learning (Section \ref{sec:background_tabular_data_representations}).

\newparagraphwithindent \textbf{Tabular in-context learners:}
We include six models primarily designed for tabular classification and regression, and extract their intermediate representations as task-agnostic embeddings. 
\textbf{TabPFN v2.5} \cite{hollmann2025tabpfn} and \textbf{TabICL v2} \cite{qu2025tabicl} are both pretrained on synthetic tabular data to perform inference without gradient updates. 
\textbf{SAP-RPT-1} \cite{spinaci2025contexttab} additionally incorporates semantics aware encodings for different data types during pretraining. 
\textbf{TabuLa-8B}~\cite{gardner2024tabula} fine-tunes an LLM on serialized rows and operates in a few-shot mode at inference.
\textbf{TabDPT} \cite{ma2025tabdpt} combines retrieval based in-context learning with self supervised pretraining on real world data.
\textbf{TARTE} \cite{kim2025tarte} pretrains on knowledge graphs to produce knowledge enhanced representations of table entries. 
Including this category lets us test whether representations optimized for predictive performance also transfer to the similarity and retrieval based tasks in our benchmark. 

\newparagraphwithindent \textbf{General Text Embedding Models:}
We additionally include general purpose text embedding models in our experiments to assess how well they handle tabular data once it is serialized into text.
\textbf{MiniLM-L6-v2} \cite{wang2020minilm} is a compact and widely used sentence embedding model. 
\textbf{IBM Granite R2} \cite{awasthy2025granite-r2} is a retrieval oriented embedding model trained on a mixture of text corpora and tables, giving it substantial exposure to structured data.
\textbf{GritLM} \cite{muennighoff2024gritlm} jointly optimizes generative and embedding objectives.

Comparing these models against the two tabular categories lets us assess whether general natural language understanding transfers to tabular data, and when tabular specific pretraining is required.

%% file: sections_04_row_embeddings.tex
\section{Benchmarking Row Embeddings}
\label{sec:row_embeddings}
\newparagraphwithindent

We benchmark the performance of row embedding models on three critical tasks, that underpin a wide range of applications: row similarity search for recommendation and duplicate detection, triplet tests to systematically assess embedding quality and tabular prediction for assisting decision-making by predicting outcomes.

\subsection{Row Similarity Search}
\label{sec:row_embeddings_row_similarity_search}
\newparagraphwithindent \textbf{Task Description.}
The goal of row similarity search is to find the most similar rows in a table given a query row. 
For example, in an e-commerce table containing sold products, the task could be to identify products similar to a given item that will then be displayed to a user.
Another use case is detecting duplicate rows that refer to the same product but appear multiple times due to data integration errors.
From an embedding perspective, a high-quality row embedding model should map semantically similar rows to nearby vectors in the embedding space, while mapping dissimilar rows farther apart. 
Row similarity search therefore provides a suitable task for assessing the quality of row embeddings. 

\newparagraphwithindent \textbf{Dataset Construction.}
Existing benchmarks do not directly target row similarity search. 
To address this, we construct evaluation data based on datasets for entity matching and entity clustering, two closely related entity-centric tasks.
Entity matching identifies pairs of rows across tables that refer to the same real-world entity, while entity clustering groups multiple rows corresponding to the same entity. 
Both tasks provide ground-truth annotations that define which rows are semantically equivalent, which we leverage to construct semantic search relevance labels.
For \benchmark, we adapt nine datasets: seven entity matching datasets \cite{koepcke2010entityresolution, mudgal2018deepmatcher} and two entity clustering datasets \cite{saeedi2017entityresolution}. 
In these datasets, the two tables always share the same schema, so that we can merge the data easily into a single table. 
Each positive entity matching pair is used to create a test case: one row serves as the query and the other as the ground-truth relevant row. 
For the entity clustering datasets, we select one row from each cluster randomly as the query, and all other rows in the cluster are treated as relevant. 
This construction ensures that each query row has at least one relevant retrieval target, enabling meaningful evaluation of embedding-based similarity.

\newparagraphwithindent \textbf{Evaluation Methodology.}
For the experiments, we create a vector representation for every row in the table.
Given a query row embedding, we retrieve the most similar other row embeddings ranked by cosine similarity, considering the top 50 candidates.
We measure retrieval performance using standard information retrieval metrics.
Mean Reciprocal Rank (MRR) captures how highly the best-ranked relevant row is retrieved for each query. 
Mean Average Precision (MAP) additionally credits retrieving further relevant rows from the same cluster near the top of the ranking, rather than just the first one, because for clustering-based datasets, a query row can have multiple relevant ground-truth rows.
We also measure Recall@k for \(k\in\{1,3,5,10\}\), defined as the fraction of a query's relevant rows that are retrieved among the top-$k$ candidates.

\begin{figure}[t]
    \centering
    
    \begin{subfigure}{\linewidth}
        \centering
        \includegraphics[width=\linewidth]{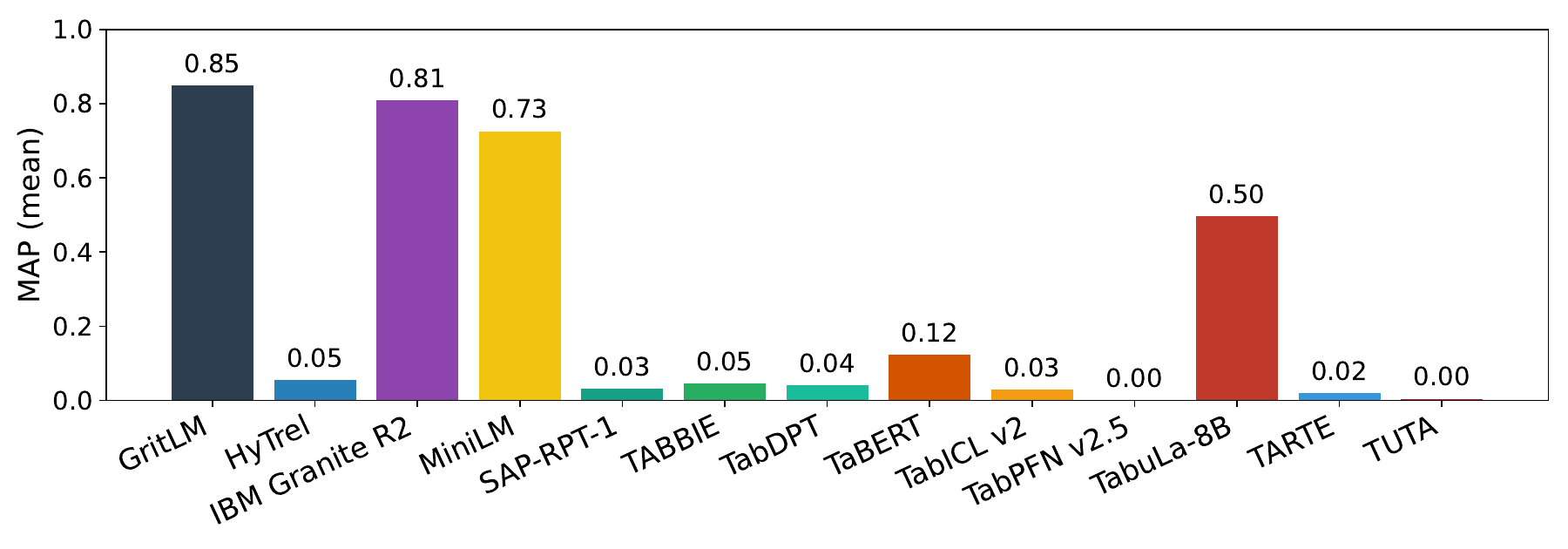}
        \caption{Row Similarity Search: Mean Average Precision (MAP) aggregated across seven datasets using the top-50 retrieved rows. \emph{TabPFN v2.5} and \emph{SAP-RPT-1} were unable to process the two largest datasets.}
        \label{fig:row_sim_mrr}
    \end{subfigure}
    
    \vspace{0.8em} 
    
    \begin{subfigure}{\linewidth}
        \centering
        \includegraphics[width=\linewidth]{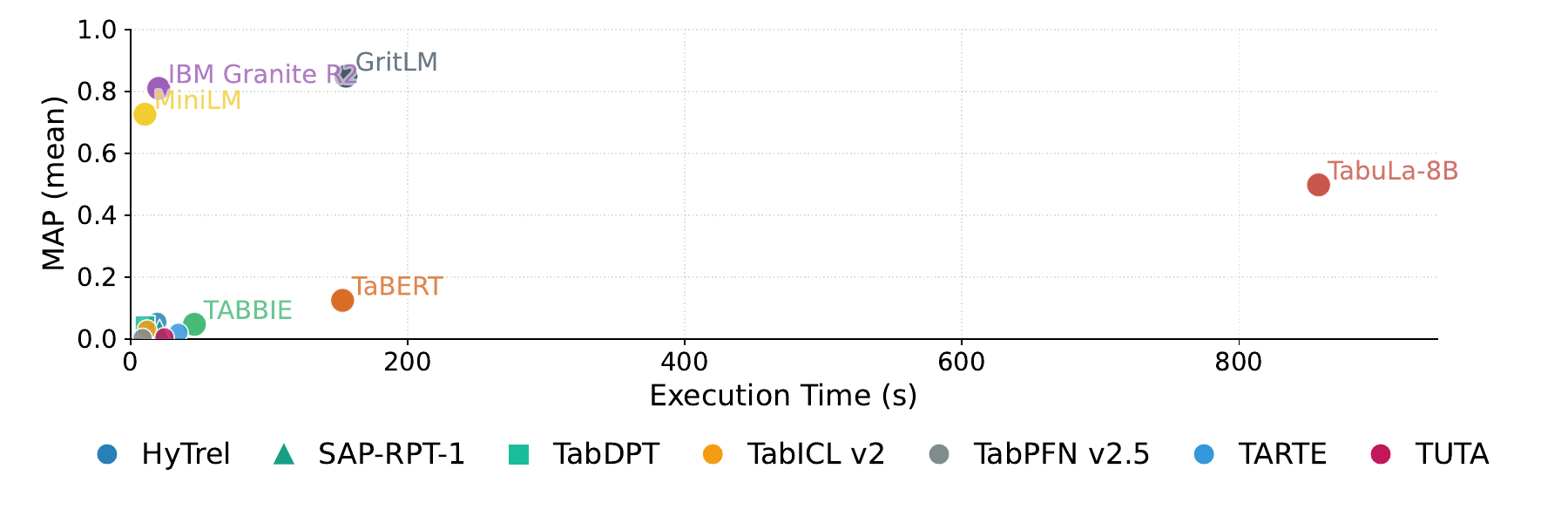}
        \caption{Row Similarity Search: Average execution time in comparison to the mean MAP score averaged over 7 datasets.}
        \label{fig:quadrant_row_sim_time}
    \end{subfigure}
    
    \vspace{0.8em} 
    
    \begin{subfigure}{\linewidth}
        \centering
        \includegraphics[width=\linewidth]{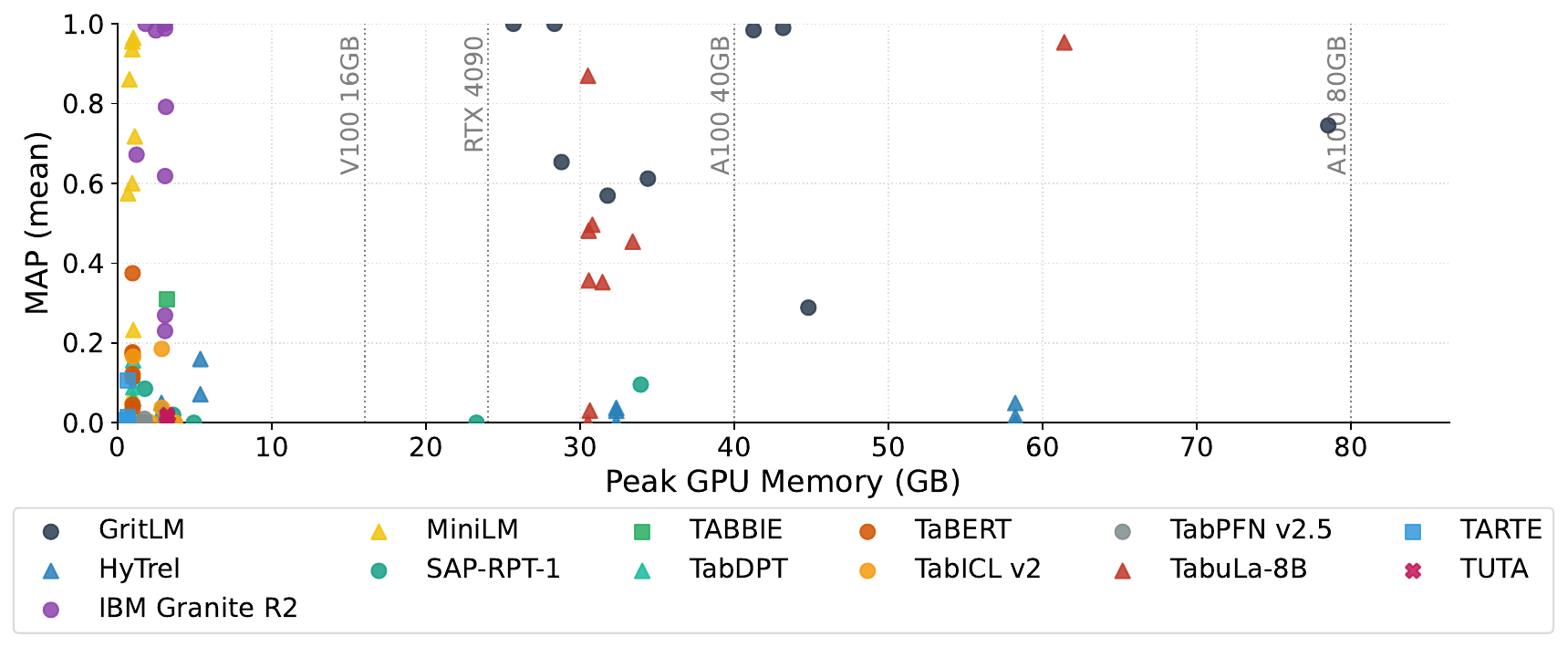}
        \caption{Row Similarity Search: Peak GPU memory in comparison to the mean MAP score for each approach per dataset. While some approaches also run on CPUs or small GPUs, larger models e.g. \emph{GritLM} need close to 80~GB of VRAM.}
        \label{fig:quadrant_row_sim_vram}
    \end{subfigure}
    \caption{Overview of row similarity search results aggregated over seven datasets with required execution time and GPU memory consumption.}
    \label{fig:exp_row_sim_overall}
\end{figure}

\newparagraphwithindent \textbf{Experiments.}
The results of benchmarking the different embedding approaches for row similarity search are shown in Figure~\ref{fig:exp_row_sim_overall}.
Surprisingly, the universal text embedding models GritLM, IBM Granite R2 and MiniLM achieve the strongest performance in this task, while embeddings extracted from tabular prediction models perform substantially worse.
This indicates that representations optimized for row-level prediction encode different aspects of the data than those needed for effective similarity-based retrieval.

\newparagraphwithindent \textbf{Resource Consumption.}
Beyond task performance, we measure model setup time, task inference time, and peak GPU memory (VRAM) for this task, as shown in Figures \ref{fig:quadrant_row_sim_time} and \ref{fig:quadrant_row_sim_vram}.
Execution time varies substantially across approaches: averaged over the datasets, TabuLa-8B needs around 15 minutes to compute embeddings and GritLM around 3 minutes, while most others need less than one minute. 
VRAM usage follows a different pattern. 
While TabuLa-8B is the slowest model, GritLM almost always requires more memory, and on the largest dataset \emph{DBLP-GoogleScholar} needs nearly the full 80GB provided. 
SAP-RPT-1's memory consumption varies considerably, running out-of-memory on two of the largest datasets.

\subsection{Triplet-Based Row Embedding Evaluation}
\label{sec:triplet_row_evaluation}
\begin{figure}
    \centering
    
    \begin{subfigure}{\linewidth}
        \centering
        \includegraphics[width=0.8\linewidth]{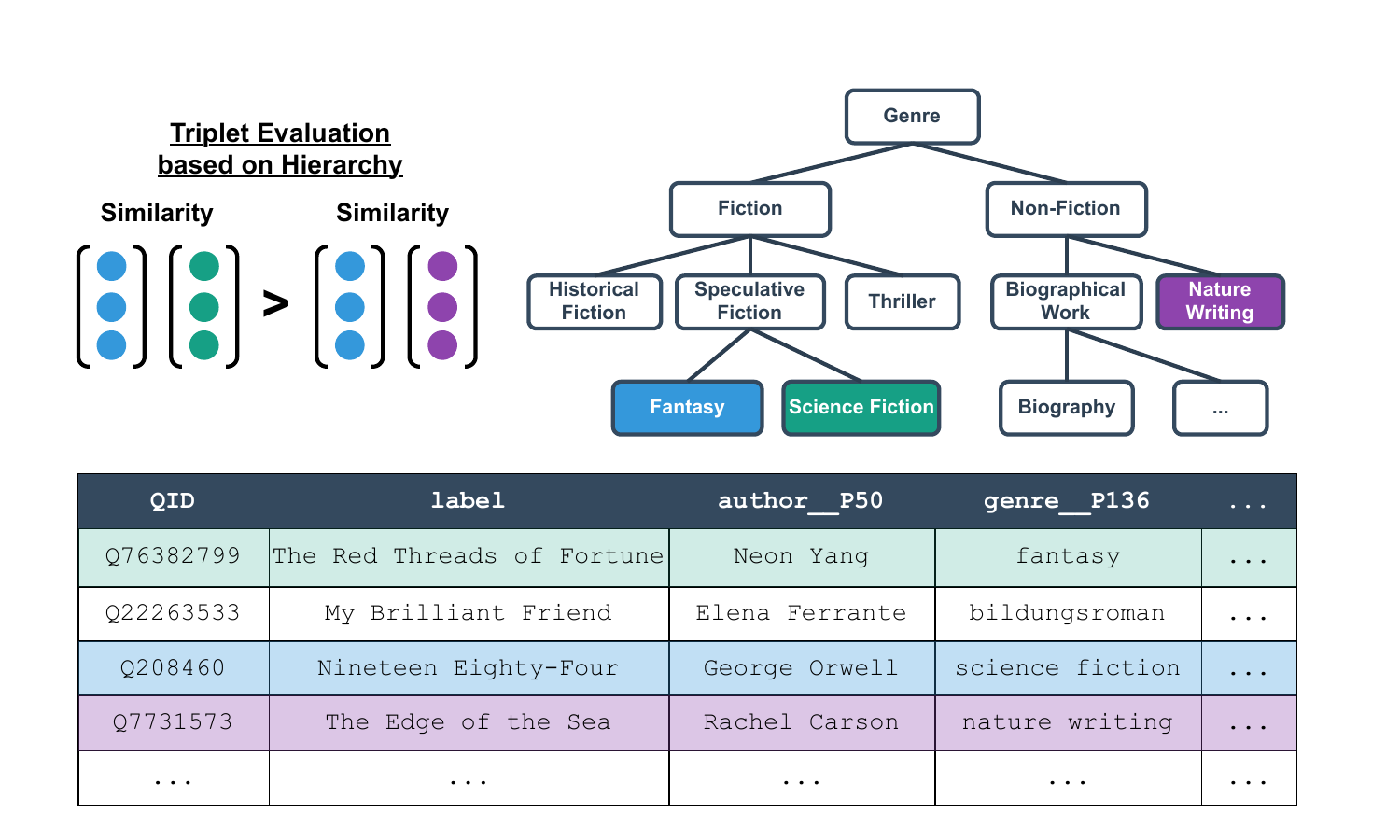}
        \caption{Example data from the Wikidata Books Hierarchy we built and the corresponding table. Triplet testcases consist of two books from the same or sibling genres, and a third book from the opposite branch (fiction vs. non-fiction books).} 
        \label{fig:hierarchy_explained}
    \end{subfigure}
    
    \vspace{0.8em} 
    
    \begin{subfigure}{\linewidth}
    \centering
    \includegraphics[width=\linewidth]{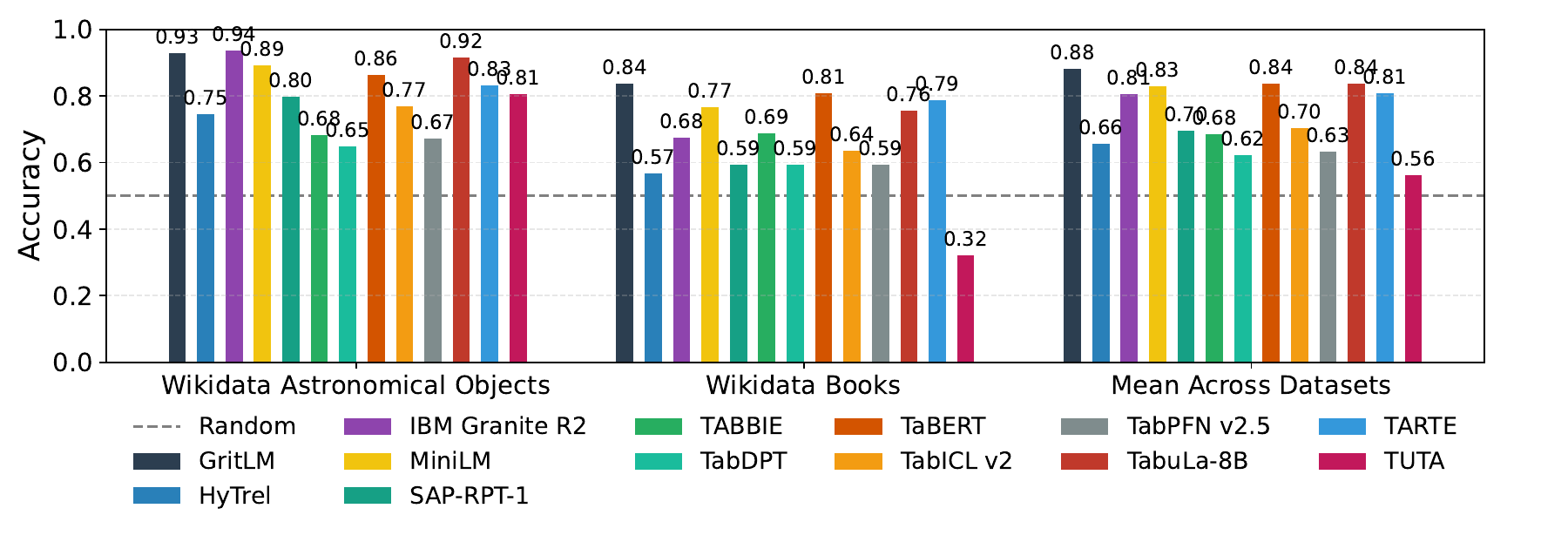}
    \caption{Triplet-based row embedding evaluation results per dataset. Accuracy measures the percentage of correctly passed testcases.}
    \label{fig:exp_row_triplet_per_dataset}
    \end{subfigure}

    \begin{subfigure}{\linewidth}
    \centering
    \includegraphics[width=\linewidth]{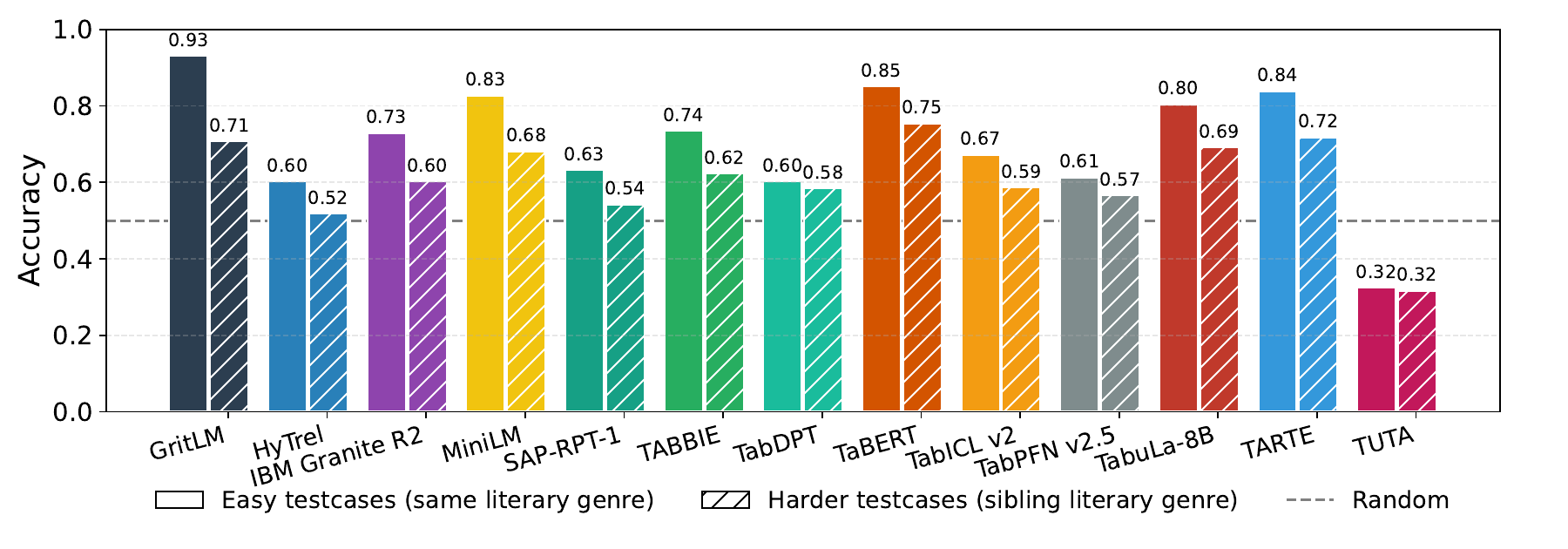}
    \caption{Fine-grained analysis of the book genre data: test cases with positives from the same literary genres vs. sibling literary genres.}
    \label{fig:exp_row_triplet_difficulty}
    \end{subfigure}

    \begin{subfigure}{\linewidth}
    \centering
    \includegraphics[width=\linewidth]{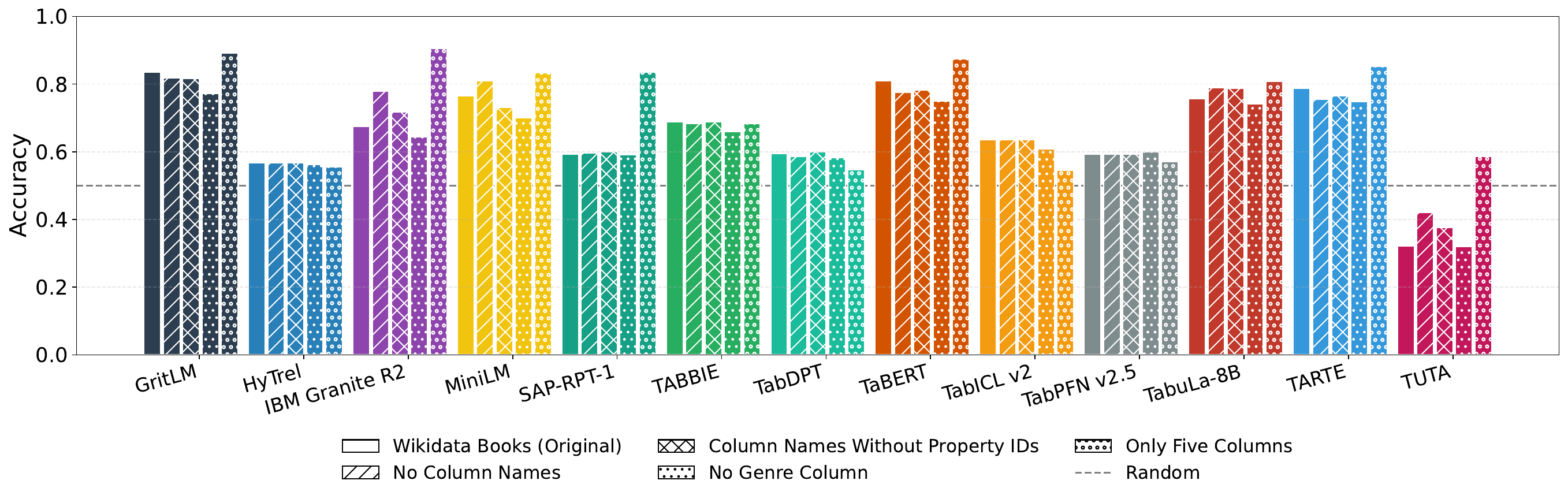}
    \caption{Testing variations of the books table, varying e.g. column names.}
    \label{fig:exp_row_triplet_books_variations}
    \end{subfigure}

    \begin{subfigure}{\linewidth}
    \centering
    \includegraphics[width=\linewidth]{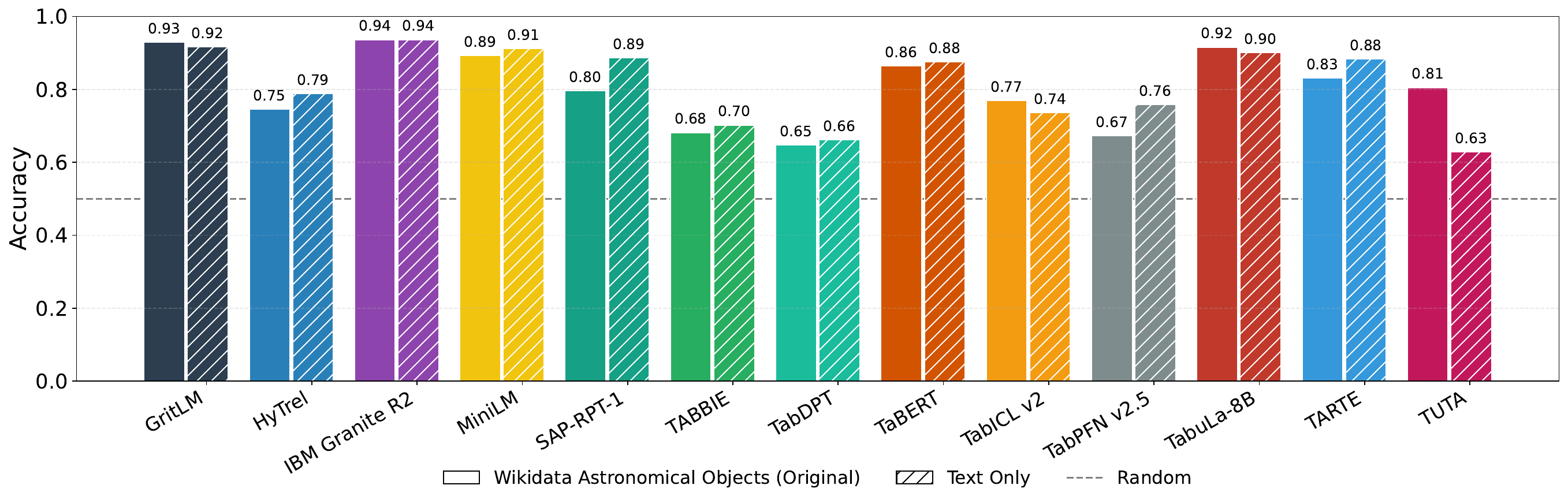}
    \caption{Original astronomical objects table vs. a version where numerical columns are removed.}
    \label{fig:exp_row_triplet_astro_variations}
    \end{subfigure}

    \caption{Triplet-Based Row Evaluations based on Tables and Hierarchies created using Wikidata.}
    \label{fig:exp_triplet}
    \vspace{-2em}
\end{figure}

The relevance labels in row similarity search are binary: derived from entity matching and clustering data, they indicate whether two rows refer to the same entity, but not how similar two different entities are.
To test whether embeddings capture graded similarity, e.g. a smartphone should be closer to a laptop than to a jacket, we add a triplet-based evaluation rooted in real-world class hierarchies. 
Rather than testing whether embeddings reconstruct a full hierarchy, this evaluation checks a more targeted property: for pairs sampled at different hierarchical distances, whether embeddings rank the closer pair as more similar than the more distant one.
We construct two such hierarchies over real entities and instance data from Wikidata \cite{wikidata} (Figure \ref{fig:hierarchy_explained}): \emph{literary genres}, chosen for their natural, interpretable structure, and \emph{astronomical objects} (e.g. star systems, galaxies, quasars), whose many numerical properties (e.g. distance from Earth, radial velocity) let us test models' ability to reason over numeric as well as textual information. 

\newparagraphwithindent \textbf{Task Description.}
We formalize this as a triplet task: given triplets (x,y,z) where y is hierarchically closer to query row x than z (e.g. x and y are both \emph{Fantasy} or \emph{Science Fiction} books), a triplet is correct if sim(x,y)>sim(x,z) using cosine similarity, and incorrect otherwise.
Wikidata's class hierarchy (Figure \ref{fig:hierarchy_explained}) lets us automatically generate such triplets across different semantic distances, enabling fine-grained evaluation of row embedding quality.

\newparagraphwithindent \textbf{Dataset Construction.}
We collect class items for the hierarchies from Wikidata by retrieving all \emph{literary work (Q7725634)} \textit{genre (Q483394)} classes and recursively all \emph{subclassOf (P279)} property classes of the \emph{astronomical object (Q6999)} item, then construct each hierarchy incrementally by inserting classes based on their superclass information (full procedure in our technical report).
As final outputs, we obtain a genre hierarchy with 111 classes and an astronomical object hierarchy with 113 classes.
For each class, we collect real instances (concrete books or astronomical objects) from Wikidata, creating tables as outlined in \cite{vogel2024wikidbs}: a books table with 1203 rows and 281 columns, and an astronomical object table with 8034 rows and 41 columns. 
We then generate triplets based on hierarchical relationships between entities, pairing books with similar genres, and astronomical objects within the same semantic branch, resulting in 903 and 2000 test cases, respectively.

\newparagraphwithindent \textbf{Evaluation Methodology.}
We report the fraction of correctly satisfied triplets as triplet accuracy, providing a direct measure of how well the embeddings capture the relative similarity ordering defined by the hierarchy. 

\newparagraphwithindent \textbf{Experiments.}
Figure \ref{fig:exp_row_triplet_per_dataset} visualizes results for both datasets.
GritLM reaches the highest performance, followed by TaBERT and TabuLa-8B, both strong despite not being optimized for this kind of task. 
TARTE also performs strongly, roughly on par with IBM Granite R2, which leads on the astronomical data but falls to the middle field on the book data. 
TUTA is also mid-field on astronomical data, but falls below chance on book genre, as it cannot process such wide tables directly and falls back to mean-pooling cell embeddings across all columns.
Figures \ref{fig:exp_row_triplet_difficulty}-\ref{fig:exp_row_triplet_astro_variations} report variation studies: e.g., splitting books test cases by difficulty, or removing column names or numeric columns.  
Most models' accuracy is sensitive to such changes.

\subsection{Tabular Prediction}
\label{sec:tabular_prediction}
Beyond capturing semantic or hierarchical relationships, embeddings should preserve information useful for downstream prediction tasks, e.g., predicting whether a product will sell well or its likely price range. 
Higher-quality embeddings should retain more of the information contained in the original row, enabling downstream models to learn more accurate predictors on top of them.
To complement the similarity-focused benchmarks, we therefore evaluate embeddings on standard tabular prediction tasks, including classification and regression, and use prediction performance as an additional indicator of  embedding quality.

\begin{figure}
    \centering
    
    \begin{subfigure}{\linewidth}
        \centering
        \includegraphics[width=\linewidth]{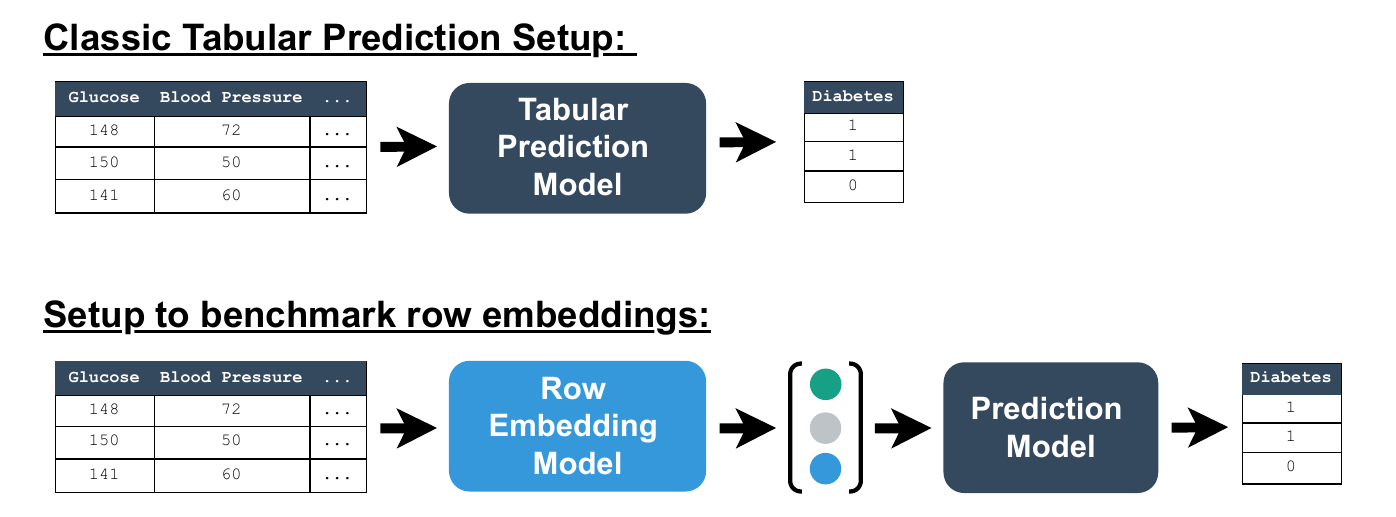}
        \caption{Tabular Prediction: Setup to benchmark prediction tasks on top of row embeddings.}
        \label{fig:tabular_prediction}
    \end{subfigure}

    \vspace{0.5em}
    
    \begin{subfigure}{\linewidth}
        \centering
        \includegraphics[width=\linewidth]{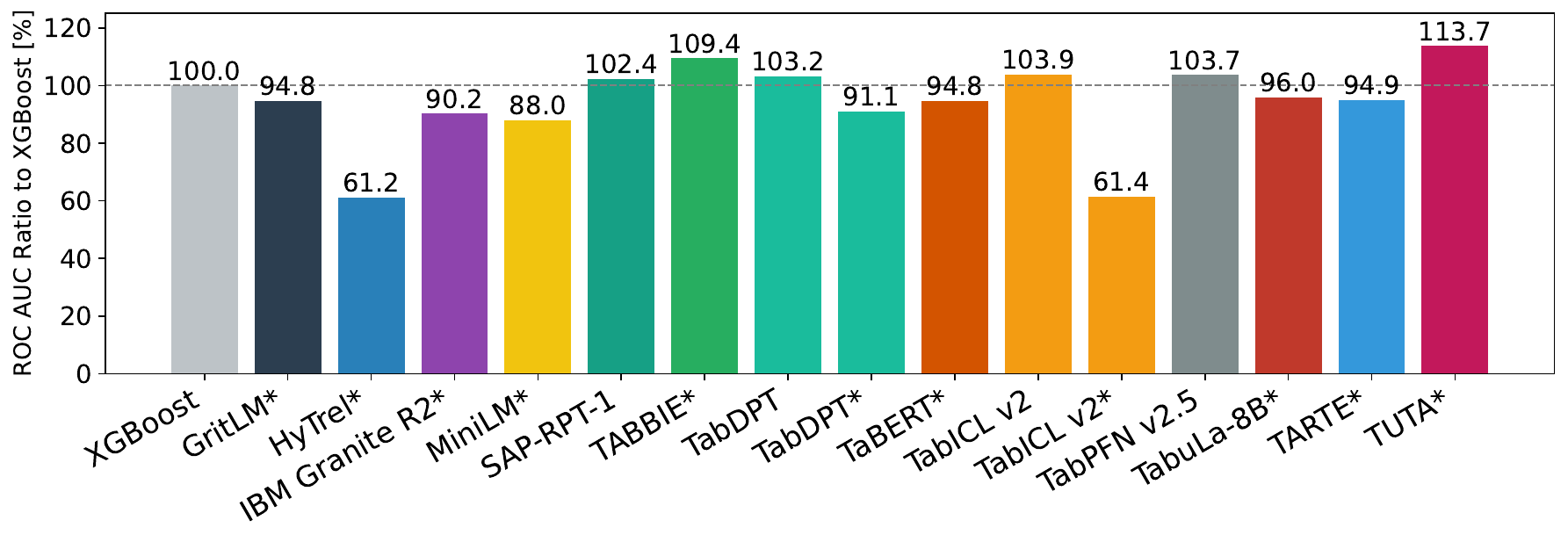}
        \vspace{-2em}
        \caption{Binary Classification: Percentage of ROC AUC Score reached compared to XGBoost trained directly on the tabular features. The results are averaged over the 23 datasets that all approaches were able to produce results for.}
        \label{fig:exp_tabular_pred_binary}
    \end{subfigure}
    
    \begin{subfigure}{\linewidth}
        \centering
        \includegraphics[width=\linewidth]{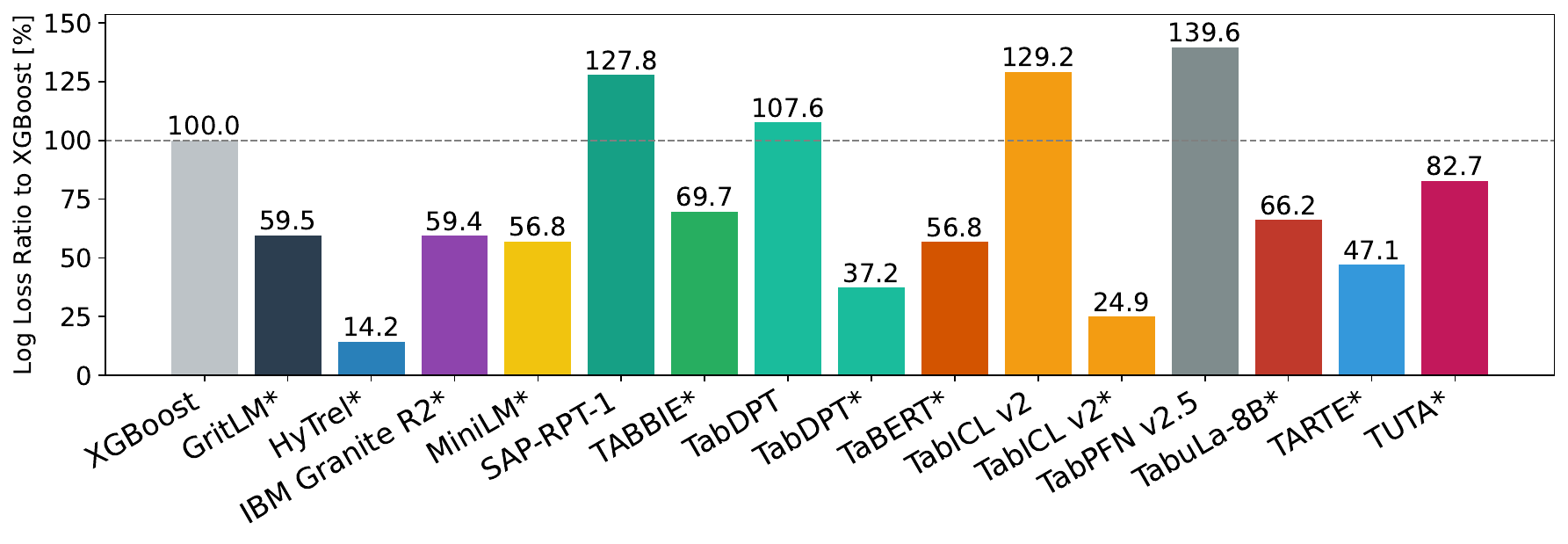}
        \vspace{-2em}
        \caption{Multiclass Classification: Percentage of Log Loss Scores reached, results averaged over 6 datasets.}
        \label{fig:exp_tabular_pred_multiclass}
    \end{subfigure}

    \begin{subfigure}{\linewidth}
        \centering
        \includegraphics[width=\linewidth]{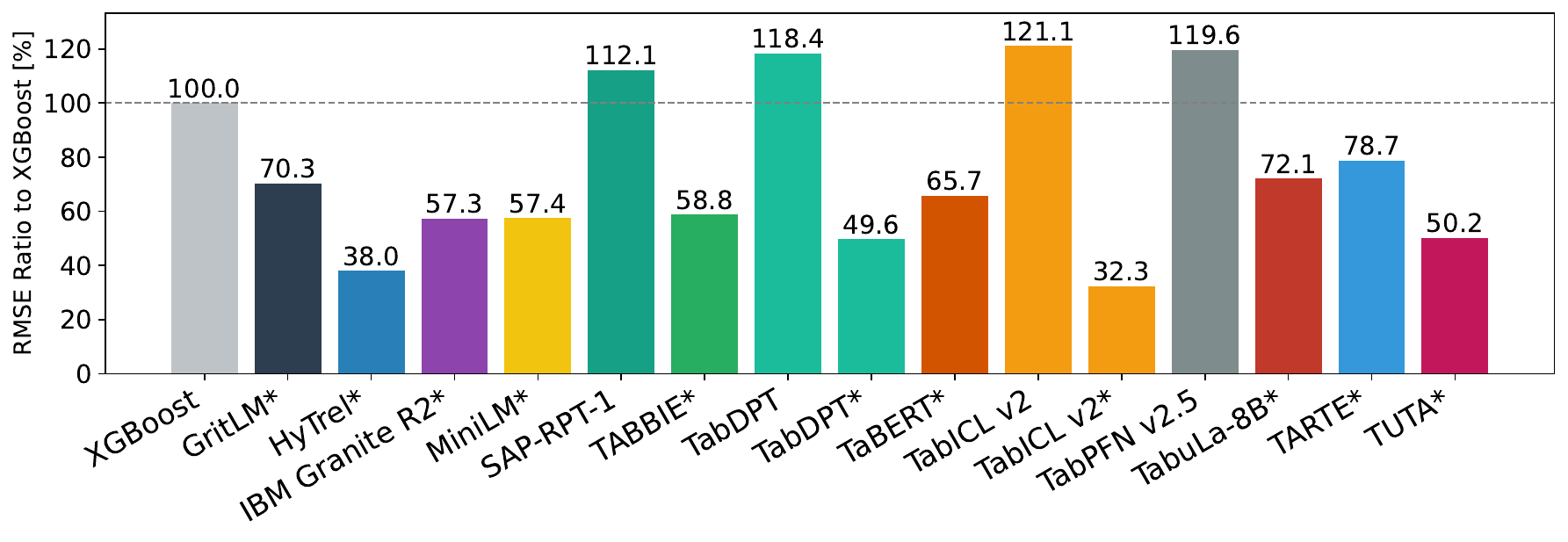}
        \vspace{-1.5em}
        \caption{Regression: Percentage of RMSE Scores reached, results averaged over 11 datasets.}
        \label{fig:exp_tabular_pred_regression}
    \end{subfigure}
    \vspace{-2em}
    \caption{Tabular Prediction. Results are shown as percentage reached in comparison of training XGBoost directly on the tabular features. * indicates that XGBoost was used on top of row embeddings. }
    \label{fig:exp_tabular_pred_overall}
\end{figure}

\newparagraphwithindent \textbf{Task Description.}
We evaluate row embeddings on standard tabular prediction tasks, to assess how well they preserve information useful for supervised learning. 
Since not all embedding models provide a native prediction head (e.g., HyTrel and sentence-transformer models), we use the learned embeddings as input features for a downstream classifier, as visualized in Figure \ref{fig:tabular_prediction}.
This allows us to compare their ability to support predictive modeling.

\newparagraphwithindent \textbf{Dataset.}
We evaluate on all 51 datasets from the TabArena-Lite benchmark \cite{erickson2025tabarena}, a curated collection of real-world tabular classification and regression tasks spanning diverse domains. 
For each dataset, we follow the pre-defined TabArena-Lite splits (fold 0, the “lite” setting).
We report \emph{ROC AUC} for binary classification, \emph{log-loss} for multiclass classification, and \emph{RMSE} for regression tasks.

\newparagraphwithindent \textbf{Evaluation Methodology.}
For each dataset, we embed all rows of the training set using the evaluated model, train a downstream classifier on these embeddings\footnote{Models trained for tabular prediction like \emph{TabPFN}, \emph{TabICL} and \emph{SAP-RPT-1} we also run directly on the data.} and the target labels, and apply the trained model to the test set, following prior work evaluating pretrained embeddings as classifier features \cite{wu2024switchtab, muennighoff2023mteb, li2020sentenceembeddings}.
The prediction performance on the test set reflects how much task-relevant information is preserved in the embeddings. 
We tested both, an MLP and XGBoost \cite{chen2016xgboost} as downstream classifiers, and report XGBoost results exclusively, as it consistently outperformed the MLP.

\newparagraphwithindent \textbf{Experiments.}
Since the multi-class classification and regression scores vary a lot per dataset, we visualize all results relative to XGBoost trained directly on the tabular data (Figure \ref{fig:exp_tabular_pred_binary}-\ref{fig:exp_tabular_pred_regression}). 
We observe that across all tasks only models originally developed for tabular prediction, such as TabPFN, TabICL, TabDPT and SAP-RPT-1 reach higher scores than the XGBoost model.
The only exception is binary classification, where row embeddings extracted from the table-specific representation models TABBIE and TUTA also surpass XGBoost.
This general pattern is expected, as deep learning based approaches before the TabPFN era used to underperform tree based models for tabular prediction \cite{grinsztajn2022tree-outperform, kadra2021castle}.
When extracting row embeddings from these models (indicated with a *), the performance is noticeably lower.
From the other approaches, \emph{TUTA}'s embeddings stand out as the strongest, being the only ones besides TABBIE to beat XGBoost, while \emph{GritLM}'s embeddings seem to contain the best information for the XGBoost classifier to learn from among the general-purpose text embedding models.
Since the three prediction sub-tasks use different metrics, following \cite{erickson2025tabarena} we compute ELO scores.

%% file: sections_05_column_embeddings.tex
\section{Benchmarking Column Embeddings}
\label{sec:col_embeddings}

\subsection{Column Similarity Search}
\label{sec:col_sim_search}
\newparagraphwithindent \textbf{Task Description.}
Given a query column, the task is to rank the most semantically similar columns from a corpus of tables, a central task in data analytics scenarios.
Because similar columns may differ in naming conventions or formatting while representing comparable concepts, this requires capturing both semantic meaning and data characteristics beyond exact string or value matching.

\newparagraphwithindent \textbf{Datasets.}
We evaluate column similarity search on four publicly available joinable column benchmarks drawn from prior work on join discovery and dataset matching, providing ground truth for which pairs of columns across tables are joinable.
\textit{Nextia$_{\boldsymbol{JD}}$} \cite{flores2021nextia} comprises open-data tables (Kaggle and OpenML) labeled with join quality for column pairs. 
Following \citet{cong2023warpgate}, we use its small ($1$ to $100$ MB files) and medium testbeds ($100$ MB to $1$ GB).
\textit{Valentine} \cite{koutras2021valentine} defines schema/attribute matching scenarios for dataset discovery, we use its \emph{semantically-joinable} subset.
\textit{OpenData}~\cite{kokel2025evaluatingjoinablecolumndiscovery} contains heterogeneous open government and web tables with human-annotated syntactic and semantic joinability.
\textit{WikiJoin} from the LakeBench suite \cite{srinivas2023lakebench} provides join discovery testbeds from web table corpora; we subset it to 659 tables (\textit{WikiJoin-Small}).  

\newparagraphwithindent \textbf{Evaluation Methodology.}
Given a query column from one table, we rank all candidate columns from a corpus by cosine similarity to the query embedding, using ground truth joinable column pairs as relevance labels.
Retrieval effectiveness is measured using standard IR metrics such as Mean Average Precision (MAP).

\begin{figure}[t]
    \centering
    \includegraphics[width=\linewidth]{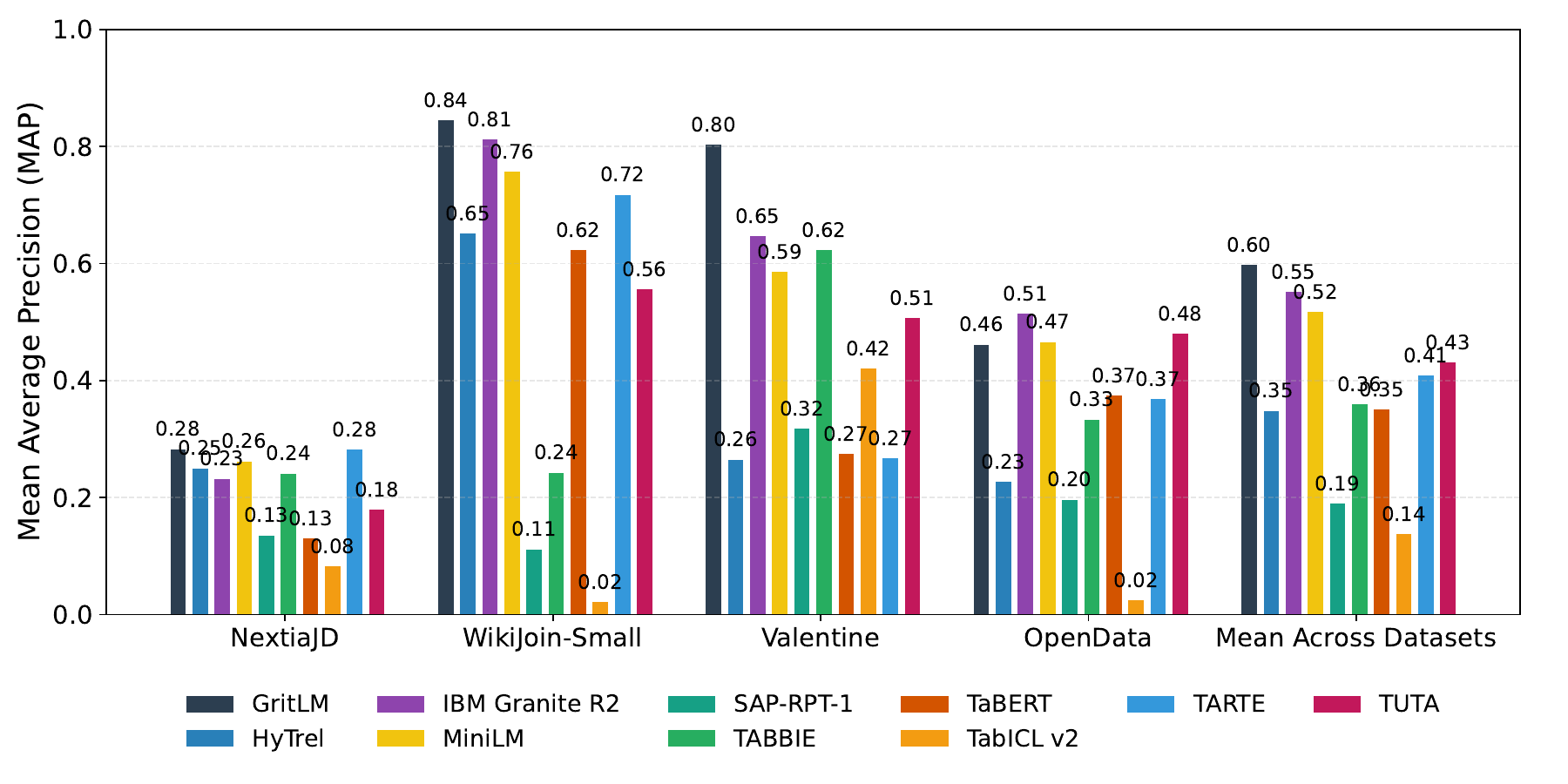}
    \caption{Column Similarity Search: Results per dataset.}
    \label{fig:exp_col_sim_per_dataset}
\end{figure}

\newparagraphwithindent \textbf{Experiments.}
The results of benchmarking different embedding approaches on column similarity search are visualized in Figure \ref{fig:exp_col_sim_per_dataset}.
Overall, \emph{GritLM} again reaches the highest performance, with \emph{IBM Granite R2} and \emph{MiniLM} being close. 
The column embeddings extracted from \emph{TabICL} seem to not include enough signals to solve the task, while those from \emph{SAP-RPT-1} seem to contain a bit more information.

\subsection{Column Type Annotation}
\label{sec:col_type_annotation}

\newparagraphwithindent \textbf{Task Description.}
Column Type Annotation assigns each column of a table a label from a predefined ontology of semantic types, e.g., \textit{date} or \textit{postal address}, a core step in data preparation when tables arrive with unreliable metadata.
We train a classifier on top of column embeddings, testing whether embeddings are separable by type once a small amount of labeled supervision is added.
This complements Column Similarity Search (Section \ref{sec:col_sim_search}), which evaluates embeddings in a purely unsupervised nearest-neighbor setting.

\newparagraphwithindent \textbf{Datasets.}
SOTAB \cite{korini2022sotab} provides schema.org-derived type labels for web table columns.
We use the \emph{CTA-SCH} round of SOTAB V2 (SemTab 2023 release), spanning 80 labels. 
We keep the official test split and subsample training tables to the smallest set covering every label with $\geq$50 columns, giving 2,777 train / 312 test tables. 
GitTables-CTA reuses the DBpedia-typed column annotations released with GitTables \cite{hulsebos2023gittables}, covering 122 labels over 2.5k labeled columns; as no official split exists, we split tables 80/20 (649 train / 165 test tables).

\newparagraphwithindent \textbf{Evaluation Methodology.}
We embed every labeled column of the train and test tables using each approach's column embedding component, train a logistic regression classifier on the training embeddings, and evaluate on the held-out test embeddings. 
We report Macro-F1 as the primary metric (given GitTables-CTA's heavily imbalanced label distribution) and accuracy.

\begin{figure}
    \centering
    \includegraphics[width=\linewidth]{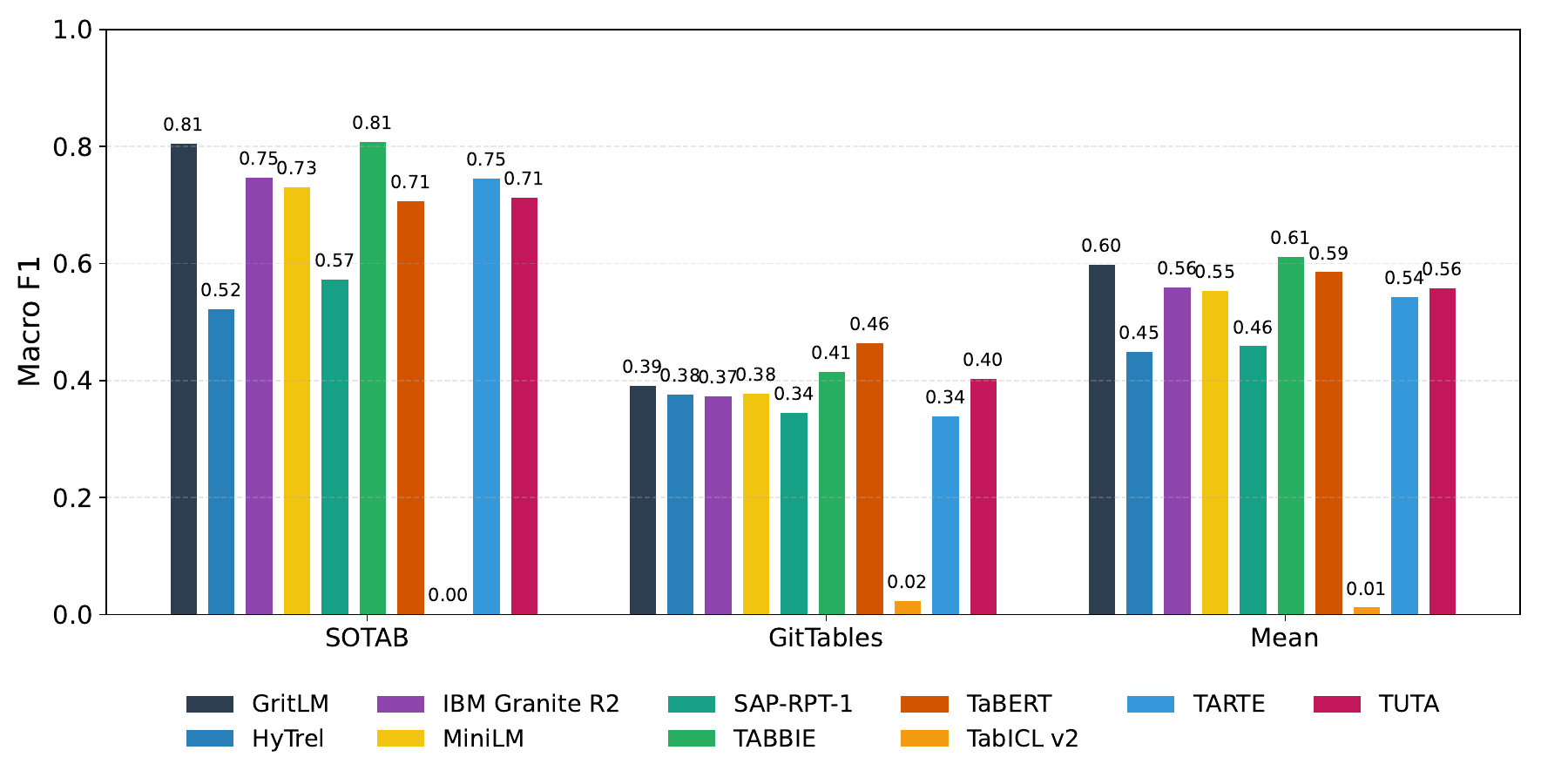}
    \caption{Column Type Annotation: Results per dataset.}
    \label{fig:exp_cta_per_dataset}
\end{figure}

\newparagraphwithindent \textbf{Experiments.}
Results are shown in Figure \ref{fig:exp_cta_per_dataset}. 
TABBIE performs strongly, consistent with a pretraining objective that trains its column representations to capture the distribution of values a column typically contains. 
GritLM follows closely, benefiting from strong general language understanding. 
TabICL v2 performs worse, as it reduces every string column to a numeric rank code before embedding, discarding the textual information CTA depends on. 
Rankings shift between datasets, in general all approaches score lower on GitTables' larger label space (122 vs. 80 types). 

\subsection{Schema Linking}
\label{sec:col_schema_linking}

\newparagraphwithindent \textbf{Task Description.}
Schema linking maps a natural language question to the database columns required to answer it, a prerequisite step in text-to-SQL systems. 
We embed the question itself with each approach's column embedding component and retrieve the nearest database columns by cosine similarity, following the same protocol as Column Similarity Search (Section \ref{sec:col_sim_search}), but with a natural language query in place of an example column.

\newparagraphwithindent \textbf{Datasets.}
We construct an evaluation set from the BIRD \cite{li2023bird} text-to-SQL benchmark: 2,157 questions across 69 databases whose gold columns are identifiable from schema and column names alone, without matching specific cell values. 
Questions that additionally require a cell value match are instead covered by Value Linking (Section \ref{sec:value_linking}). 
On average, each question requires identifying 2.6 gold columns among all columns of its database.

\newparagraphwithindent \textbf{Evaluation Methodology.}
For each database, we index the column embeddings of all its tables using each approach's column embedding component. 
For each query, we embed the question with the same component and retrieve the nearest columns by cosine similarity against the indexed corpus.
We report Mean Average Precision (MAP) as the primary metric.

\begin{figure}
    \centering
    \includegraphics[width=\linewidth]{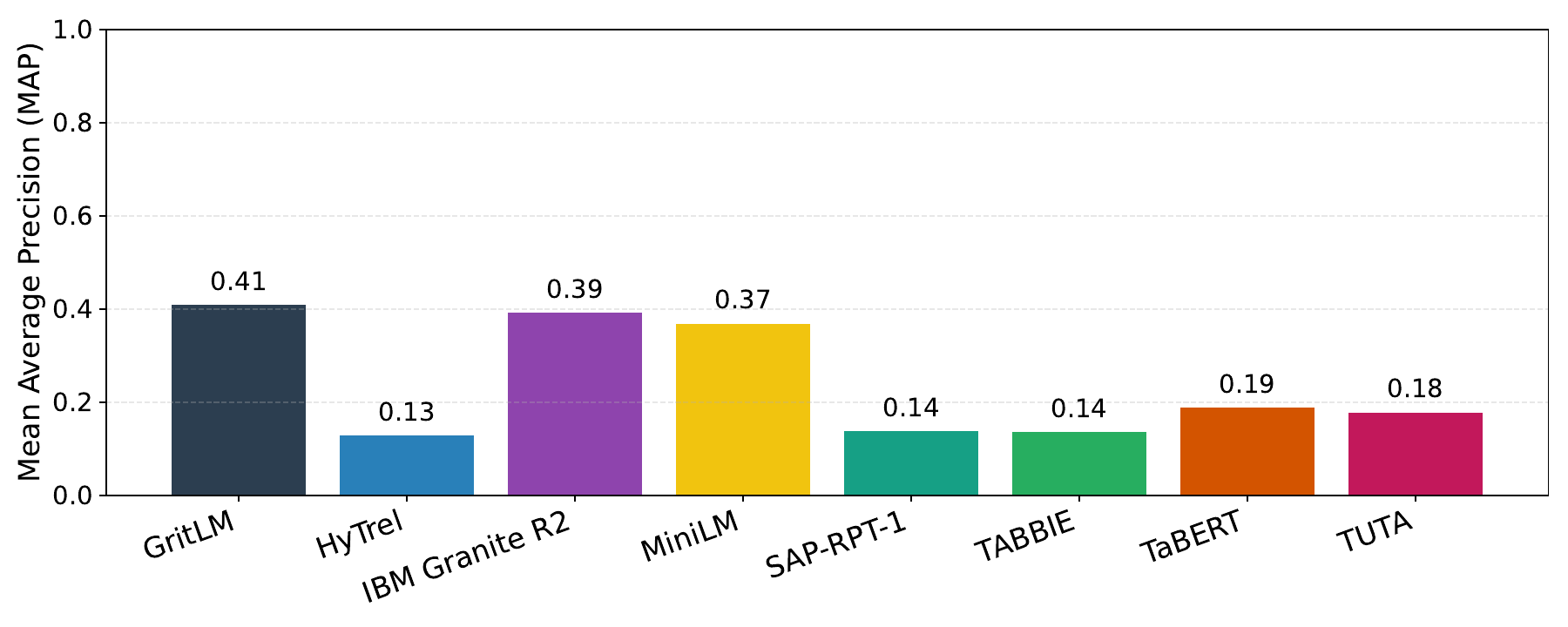}
    \caption{Schema Linking: Mean Average Precision (MAP) per approach on the adapted BIRD dataset.}
    \label{fig:exp_schema_linking}
\end{figure}

\newparagraphwithindent \textbf{Experiments.}
Figure \ref{fig:exp_schema_linking} visualizes the results.
All approaches score notably lower than on Column Similarity Search, reflecting the harder matching problem: retrieval must bridge a natural-language question and a column representation, rather than two columns of comparable structure. 
GritLM, IBM Granite R2, and MiniLM perform strongest, as their natural-language pretraining lets them embed free-form questions into the same space as column representations. 
Table-native encoders such as HyTrel, TABBIE, and SAP-RPT-1 score lower, since their column embeddings capture column content rather than alignment with query text.

%% file: sections_06_table_embeddings.tex
\section{Benchmarking Table Embeddings}
\label{sec:table_embeddings}

We benchmark table embedding models on three tasks that probe complementary properties: Table Similarity Search for identifying related tables within a repository, Table Retrieval for locating tables relevant to a natural-language query, and Table Shuffling Evaluation for testing structural fidelity to a table's row/column layout.

\subsection{Table Similarity Search}
\label{sec:table_similarity_search}

\newparagraphwithindent \textbf{Task Description.}
Given a query table, we retrieve the most similar tables from a repository, testing whether table embeddings distinguish tables from the same source from unrelated ones.

\newparagraphwithindent \textbf{Dataset.}
We construct a new dataset based on GitTables \cite{hulsebos2023gittables}, a large collection of real-world CSV tables extracted from public GitHub repositories. 
From GitTables, we select 20 topic-specific subsets (referred to as \emph{data lakes}), each corresponding to a high-level table topic provided in the dataset metadata (e.g. \textit{orbit period} or \textit{inflation rate}), with a few hundred tables each.
Restricting each data lake to a single topic ensures that retrieval cannot rely solely on coarse domain differences and instead requires identifying semantically related tables within a domain.
To construct test cases, we select query tables from repositories containing at least five tables to ensure a sufficient number of relevant targets.
For each query table, the remaining tables from the same repository are treated as ground-truth relevant tables, yielding 348 test cases in total.

\newparagraphwithindent \textbf{Evaluation Methodology.}
For each data lake, we embed all tables and, given a query table, retrieve the most similar other tables by cosine similarity. 
We report standard information retrieval metrics (\emph{MRR} and \emph{MAP}). 
Recall measures the fraction of relevant tables retrieved among the results, where $k$ equals the number of ground-truth tables. 

\newparagraphwithindent \textbf{Experiments.}
Table \ref{tab:table_similarity_gitTables} shows the results averaged across all test cases and data lakes.
We experiment with two variants of creating table embeddings: one including just the table schema, the other adding 100 example rows \cite{ji2025target_benchmark}.
Most of the models reach higher results when provided example rows as context, and TaBERT reaches performance close to GritLM despite its smaller size. 
MiniLM and TABBIE are exceptions, both scoring higher with schema alone, suggesting the extra row content does not add useful signals for their representations. TabDPT, whose table embedding is a mean over row embeddings rather than a dedicated table-level representation, performs markedly worse than the other approaches.

\begin{table}[t]
\centering
\caption{Table Similarity Search: Results on GitTables, retrieving tables from the same git repository. Comparing schema-only vs schema+rows. TabDPT's and TARTE's table embeddings are a mean over row embeddings, supporting no schema-only representation.}
\resizebox{\columnwidth}{!}{%
\begin{tabular}{lccc|ccc}
\toprule
 & \multicolumn{3}{c}{Variant: Just Schema} & \multicolumn{3}{c}{Variant: Schema + 100 rows} \\
Approach & MRR & MAP & Recall & MRR & MAP & Recall \\
\midrule
GritLM & \textbf{0.8873} & \textbf{0.7773} & \textbf{0.8028} & \textbf{0.8933} & \textbf{0.7952} & \textbf{0.8186} \\
HyTrel & 0.8182 & 0.6673 & 0.6999 & 0.8299 & 0.6687 & 0.6901 \\
IBM Granite R2 & 0.8586 & 0.7465 & 0.7705 & 0.8860 & 0.7811 & 0.8077 \\
MiniLM & 0.8495 & 0.7464 & 0.7695 & 0.8483 & 0.6754 & 0.7085 \\
TABBIE & 0.8342 & 0.6923 & 0.7144 & 0.8020 & 0.6051 & 0.6345 \\
TabDPT & - & - & - & 0.6402 & 0.4002 & 0.4408 \\
TaBERT & 0.8728 & 0.7614 & 0.7876 & 0.8903 & 0.7816 & 0.8049 \\
TARTE & - & - & - & 0.8667 & 0.7515 & 0.7737 \\
TUTA & 0.8523 & 0.7167 & 0.7383 & 0.8506 & 0.7078 & 0.7311 \\
\bottomrule
\end{tabular}%
}

\label{tab:table_similarity_gitTables}

\end{table}

\subsection{Table Retrieval}
\label{sec:table_retrieval}

\newparagraphwithindent \textbf{Task Description.}
Table retrieval ranks tables by relevance to a natural-language query, e.g. finding the right table(s) in a data lake to answer a question. 
This complements Table Similarity Search (Section \ref{sec:table_similarity_search}), which retrieves via an example table rather than free text. 
Adding the task was originally proposed as an extension to \benchmark~ by \cite{poostforoushan2026tembedt}, and we integrate it into the main benchmark.

\newparagraphwithindent \textbf{Datasets.}
We adopt the five corpora from the TARGET \cite{ji2025target_benchmark} benchmark: \emph{FeTaQA} \cite{nan2022fetaqa}, \emph{TabFACT} \cite{chen2020tabfact}, \emph{OTT-QA} \cite{chen2021ottqa}, \emph{Spider} \cite{yu2018spider} and \emph{BIRD} \cite{li2023bird}.
Following \cite{poostforoushan2026tembedt}, we treat Spider's train, validation, and test splits as separate datasets, yielding 7 datasets in total, ranging from 75 to 2,003 tables and 1,034 to 12,779 queries each.

\newparagraphwithindent \textbf{Evaluation Methodology.}
For each dataset, we embed every table of its corpus, linearized to markdown with up to 100 example rows for the text serialization based models. 
The top-k closest tables to the query are retrieved by cosine similarity. 
We report Recall@k for \(k\in\{1,3,5,10\}\) to show how retrieval quality improves as more candidates are considered.

\begin{figure}
    \centering
    \includegraphics[width=\columnwidth]{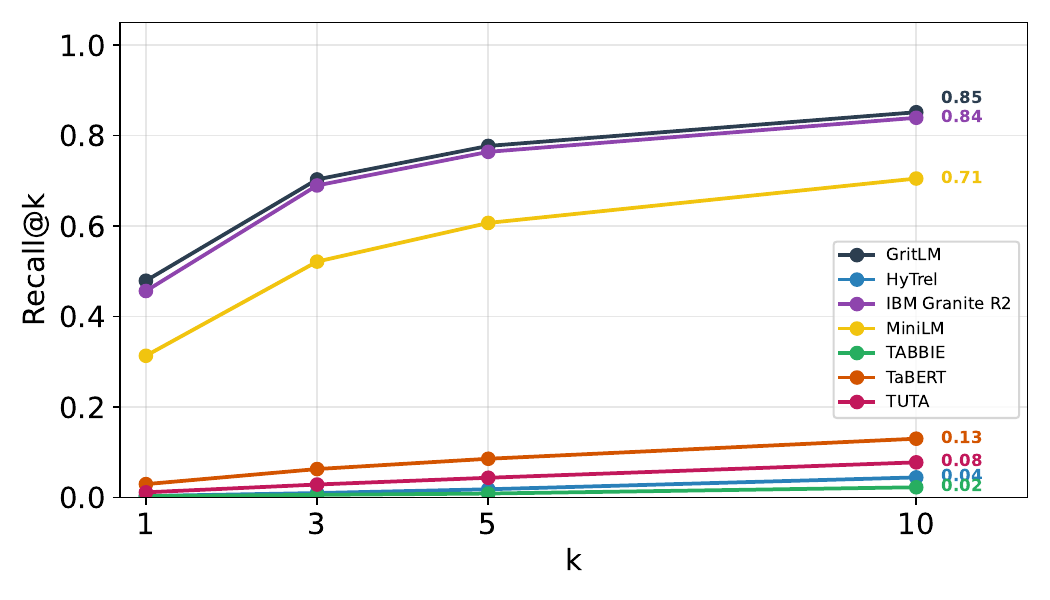}
    \caption{Table Retrieval: Recall@k for \(k\in\{1,3,5,10\}\) averaged over the 7 datasets from the TARGET benchmark \cite{ji2025target_benchmark}}
    \label{fig:exp_table_retrieval}
\end{figure}

\newparagraphwithindent \textbf{Experiments.}
Figure \ref{fig:exp_table_retrieval} shows results averaged across the seven datasets. 
GritLM and IBM Granite R2 perform strongest, again benefiting from natural-language pretraining that lets them embed queries and table content into a shared space.
The table-native encoders perform markedly worse, mirroring the pattern observed in Schema Linking (Section \ref{sec:col_schema_linking}): their table representations are trained to capture table content, not to align with free-form natural-language queries.

\subsection{Table Shuffling Evaluation}
\label{sec:table_shuffling}

\newparagraphwithindent \textbf{Task Description.}
Table Shuffling Evaluations test structural fidelity: whether an embedding captures a table's relational structure rather than only its token content \cite{cong2023observatory}.
For an anchor table, we construct a structurally faithful positive by permuting rows and/or columns, and a value-shuffled negative by permuting values independently within a subset of columns, breaking row-wise associations while leaving the column-wise token multiset unchanged.
As in the row-level triplet evaluation (Section 4.2), a triplet (anchor x, positive y, negative z) is correct if sim(x,y)>sim(x,z) under cosine similarity.
Since the positive and negative are lexically indistinguishable from the anchor, a model can only rank the positive closer by encoding structure rather than just content. 
This task was proposed by \cite{poostforoushan2026tembedt}, and complements Table Retrieval and Table Similarity Search, which probe semantic rather than structural properties of table embeddings.

\newparagraphwithindent \textbf{Datasets.}
We generate triplets from four table retrieval corpora (FeTaQA \cite{nan2022fetaqa}, TabFact \cite{chen2020tabfact}, OTT-QA \cite{chen2021ottqa}, Spider-train \cite{yu2018spider}) and two LakeBench \cite{srinivas2023lakebench} datasets, CKAN and ECB (chosen for their numerical contents).
Each triplet is parameterized along two axes: permutation type (row-only, column-only, or both), and perturbation magnitude, applied independently to the positive and negative, yielding nine variations. 
The positive perturbation magnitude (fraction of rows/columns permuted) is set to 0.75 (high) or 0.25 (low); the negative perturbation (intra-column shuffling) combines a column fraction and a within-column shuffle degree, set to 0.7/0.8 (high) or 0.2/0.2 (low).
We refer to \cite{poostforoushan2026tembedt} for the corresponding table-size ablation, which we do not repeat here.

\newparagraphwithindent \textbf{Evaluation Methodology.}
As in Section \ref{sec:triplet_row_evaluation}, we report the fraction of correctly satisfied triplets as triplet accuracy, as a direct measure of structural fidelity. A score of 0.5 corresponds to chance.

\begin{figure}
    \centering
    \includegraphics[width=\columnwidth]{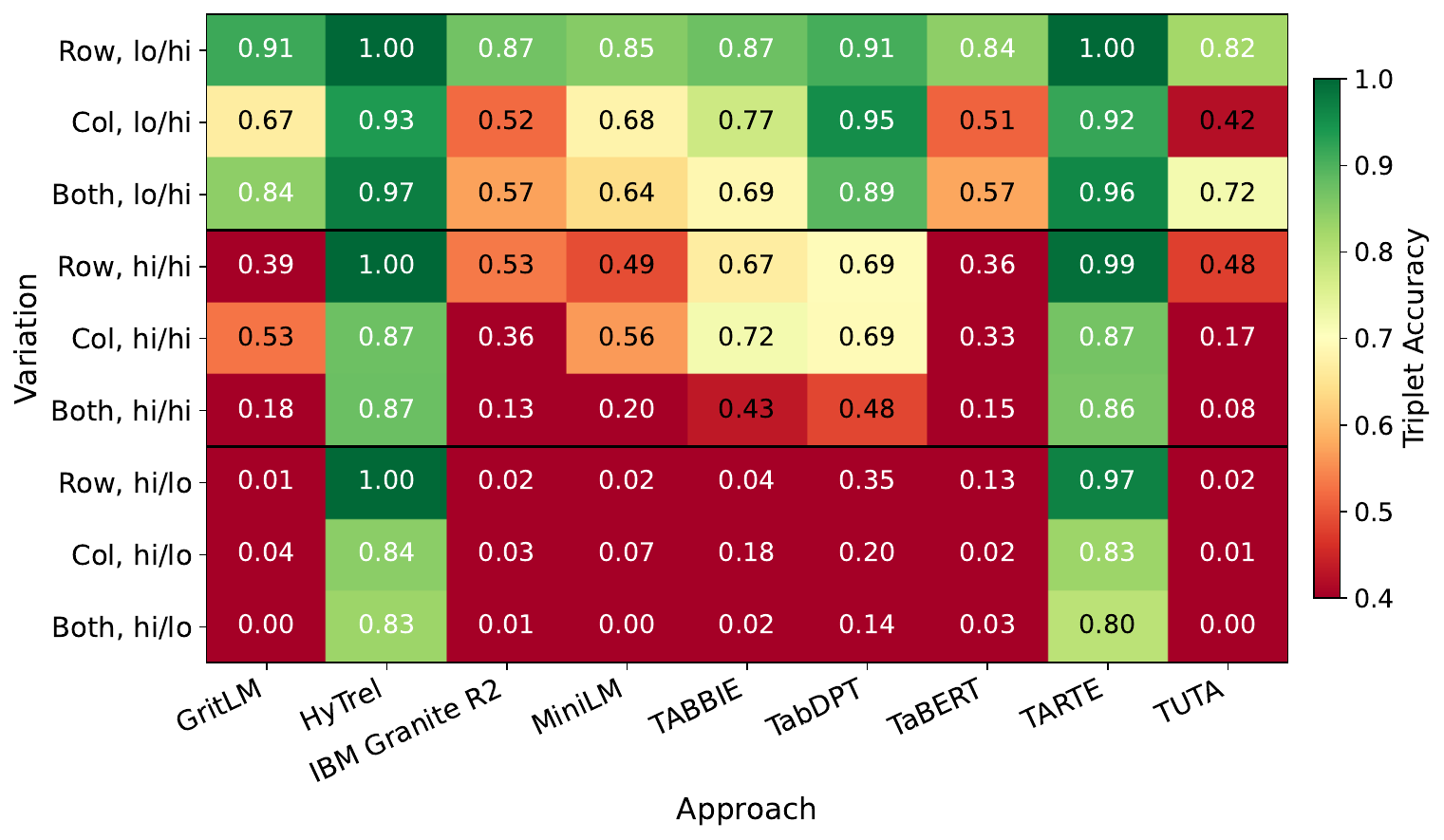}
    \caption{Table Shuffling: Triplet accuracy (mean over datasets) per approach and shuffling variation. Each variation applies a reorder type (Row, Col, or Both) for the positive example, and independently scrambles values in a subset of columns for the negative example. The two-letter label gives the strength of each: hi/hi = strong reordering \& strong scrambling, lo/hi = weak reordering \& strong scrambling, hi/lo = strong reordering \& weak scrambling. Rows are grouped into three magnitude bands, separated by horizontal lines.}
    \label{fig:exp_table_shuffling}
\end{figure}

\newparagraphwithindent \textbf{Experiments.}
Figure \ref{fig:exp_table_shuffling} visualizes the results. 
Only two approaches, HyTrel and TARTE, remain near-ceiling across all nine variations, showing they capture genuine row/column permutation invariance rather than surface token order. 
All other approaches are far more sensitive to perturbation magnitude than to permutation type: accuracy is high when the positive is only weakly reordered, but collapses toward chance or below once the positive is strongly reordered while the negative is left largely intact. 
These embeddings thus treat strong reordering as dissimilarity, capturing content overlap rather than structure.

%% file: sections_07_cell_embeddings.tex
\section{Benchmarking Cell Embeddings}
\label{sec:cell_embeddings}

We benchmark cell embeddings on two tasks that probe fine-grained, entity-level semantics: Cell Similarity Search for identifying mentions of the same real-world entity across tables, and Value Linking for matching a value mentioned in a natural-language question to the database cell it refers to.

\subsection{Cell Similarity Search}
\label{sec:cell_similarity_search}

\newparagraphwithindent \textbf{Task Description.}
Given a query cell from a table, the task is to retrieve the most semantically similar cells from a corpus of tables, despite, e.g., variation in surface form or formatting. 
The benchmark does not restrict how embeddings are generated: some models encode only the cell content, while others may incorporate row, column, or table context. 

\begin{figure}
    \centering
    
    \begin{subfigure}{\linewidth}
        \centering
        \includegraphics[width=\linewidth]{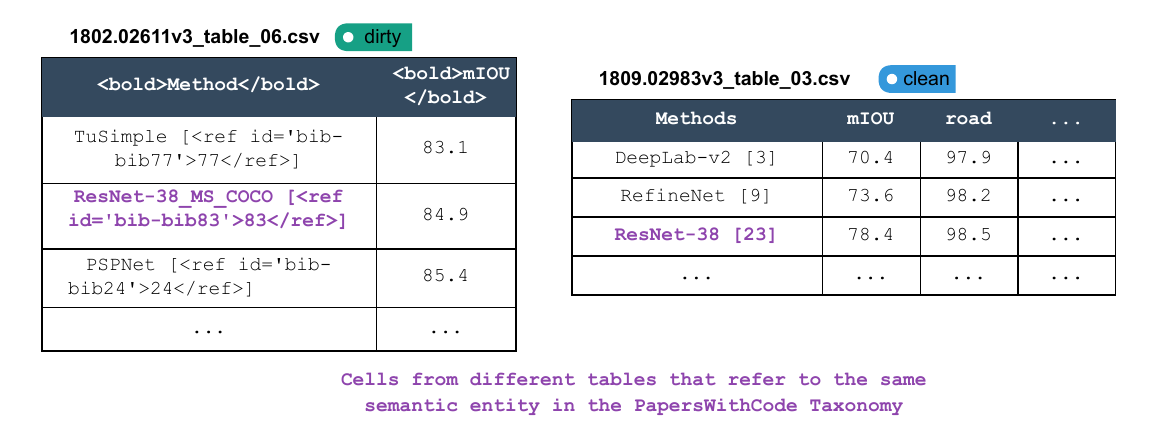}
        \vspace{-1em}
        \caption{Example data from the S2abEL dataset \cite{lou2023s2abel}. Cells containing information on the same semantic entity are considered ground truth for our similarity search.}
        \label{fig:cell_task_explained}
    \end{subfigure}
    
    \vspace{0.8em} 
    
    \begin{subfigure}{\linewidth}
        \centering
        \includegraphics[width=\linewidth]{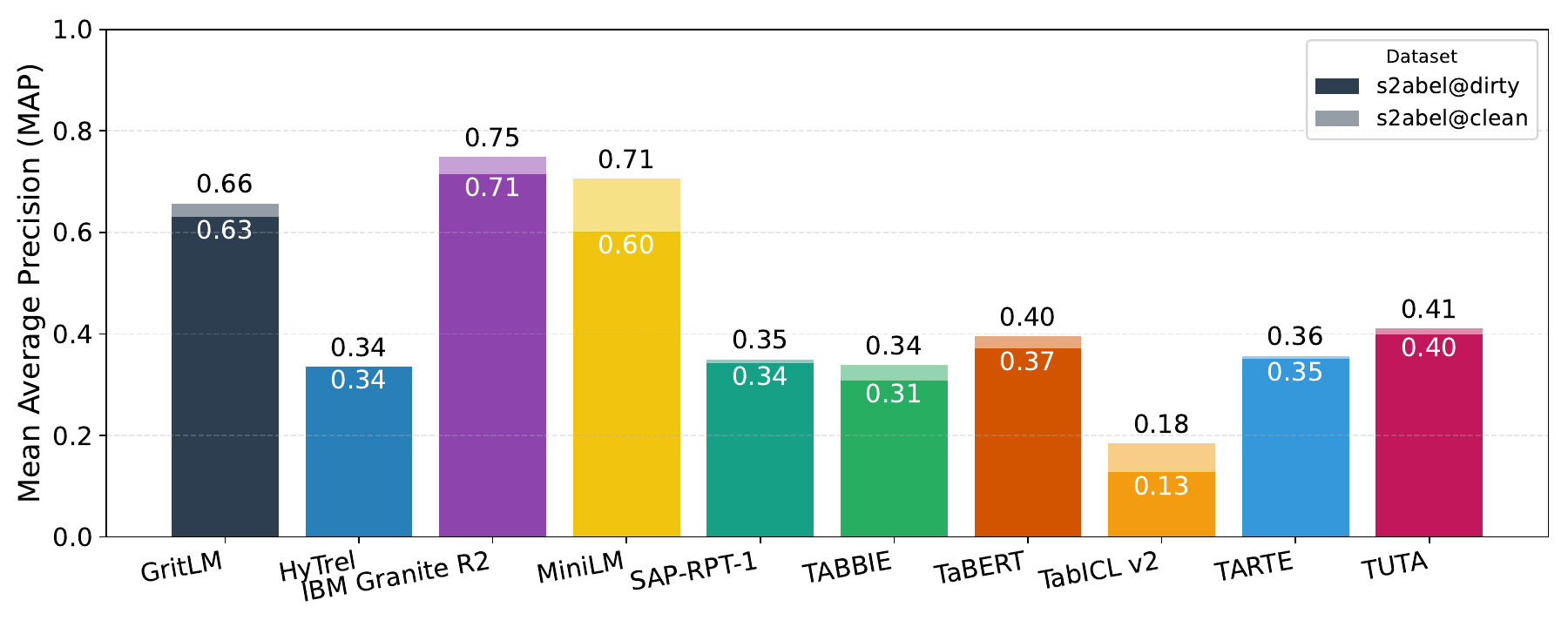}
        \caption{Mean Average Precision (MAP) for methods that provide cell embeddings on the S2abEL clean and dirty dataset versions. Bars indicate how well each method ranks ground-truth cells of the same entity near the top of the retrieval results. The semi-transparent overlay indicates results on the clean-dataset results for comparison.}
        \label{fig:cell_results_bar_chart}
    \end{subfigure}

    \vspace{0.8em} 

    \begin{subfigure}{\linewidth}
        \centering
        \includegraphics[width=\linewidth]{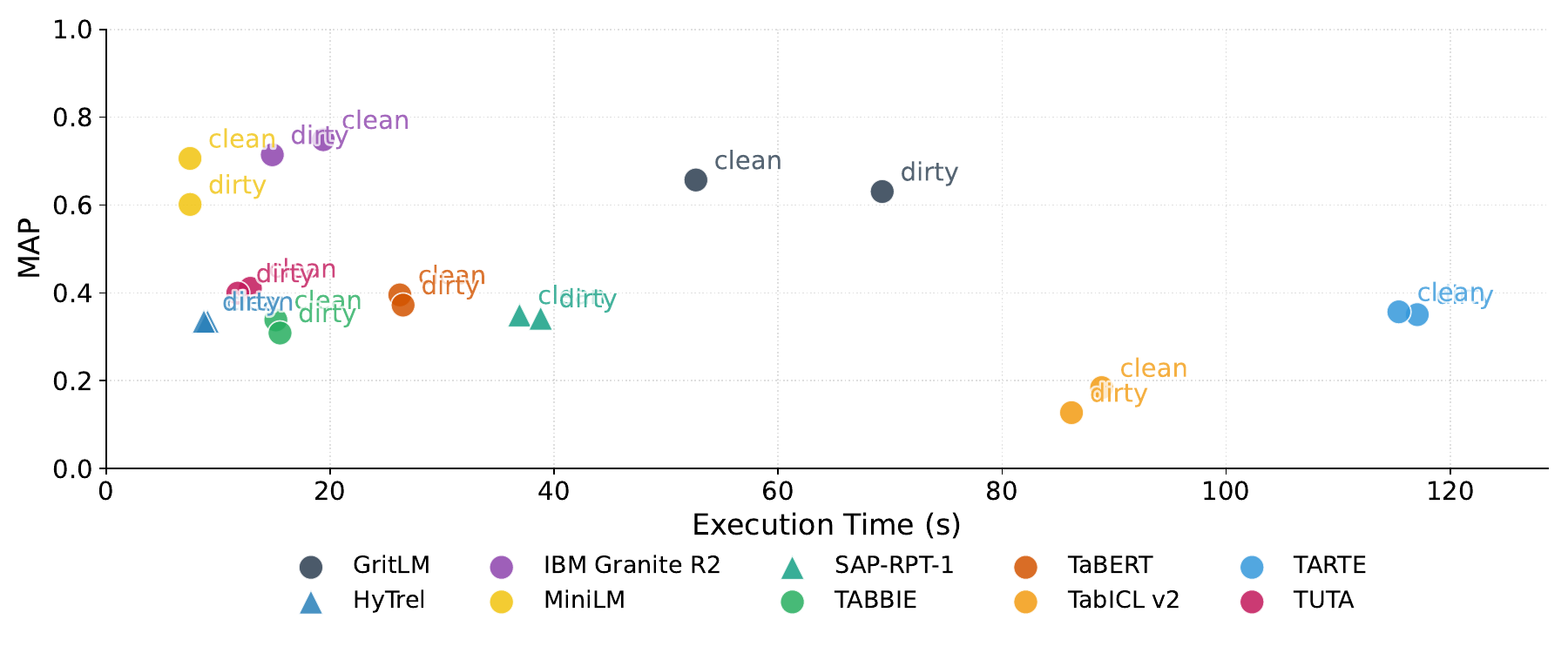}
        \vspace{-1em}
        \caption{Execution time compared to MAP. Different approaches require varying amounts of extra time if the cell texts are longer. The \textit{dirty} version of the s2abEL dataset has on average 10.2 characters per cell, the \textit{clean} version only 7.4 characters.}
        \label{fig:cell_sim_quadrant_time}
    \end{subfigure}

    \caption{Cell Similarity Search.}
    \label{fig:exp_cell_overall}
\end{figure}

\newparagraphwithindent \textbf{Dataset.}
We construct a benchmark from the S2abEL dataset~\cite{lou2023s2abel}, an entity linking dataset of 732 result tables extracted from 327 machine learning papers, whose cells are manually linked to the PaperswithCode taxonomy. 
These annotations provide a semantically grounded relevance signal: cells linked to the same entity are considered equivalent for similarity purposes, as visualized in Figure \ref{fig:cell_task_explained}.
Since the original release retains LaTeX commands and formatting artifacts, e.g.,  \texttt{\textbackslash textbf\{BERT\}}, we construct \emph{`clean'} and \emph{`dirty'} versions with the original content, allowing evaluation of both preprocessed and realistic, noisy inputs.
We sample 1000 test cases, each pairing a query cell with ground-truth cells of the same entity from up to two other tables, excluding entities linked to only a single table.

\newparagraphwithindent \textbf{Evaluation Methodology.}
For each test case, embeddings are generated for all cells of the test case tables, including header cells.
We rank candidate cells by cosine similarity to the query cell, excluding the query cell itself. 
We report Mean Average Precision (MAP), to credit retrieving multiple ground-truth cells near the top of the ranking rather than just the first one.

\newparagraphwithindent \textbf{Experiments.}
As expected, most models score higher on clean data than on the noisy version (Figure \ref{fig:cell_results_bar_chart}), with \emph{IBM Granite R2} reaching the highest performance on both variants.
Interestingly, formatting noise barely affects \emph{GritLM}, \emph{HyTrel}, and \emph{SAP-RPT-1}, whereas \emph{MiniLM} and \emph{TabICL} show a clear gap.
Execution time is largely unaffected by cell content length for most models (Figure \ref{fig:cell_sim_quadrant_time}), but \emph{GritLM} requires around 16 additional seconds to embed the longer data.

\subsection{Value Linking}
\label{sec:value_linking}

\begin{figure}
    \centering
    \includegraphics[width=\linewidth]{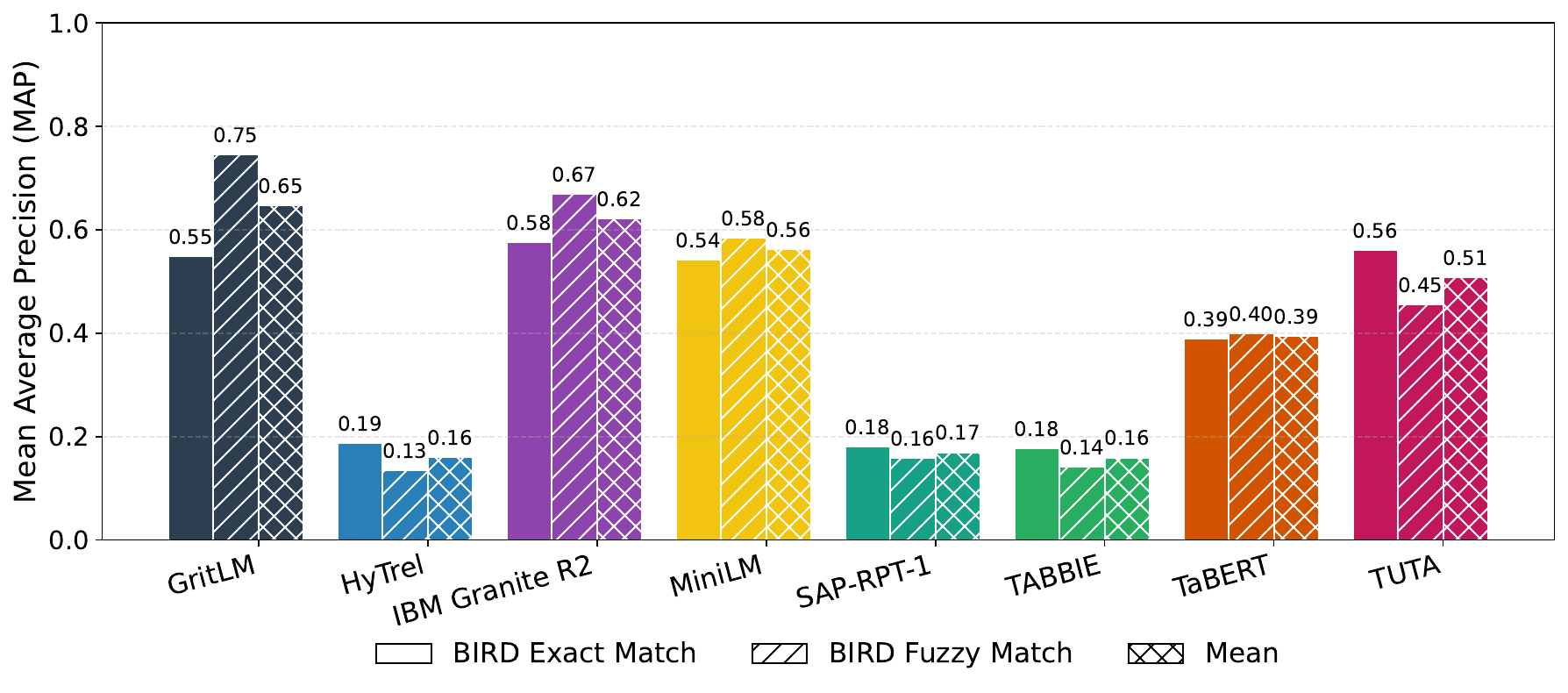}
    \vspace{-2em}
    \caption{Value Linking: MAP Scores per approach on the two variants: \emph{exact} vs. \emph{fuzzy} match data.}
    \label{fig:exp_value_linking}
    \vspace{-1em}
\end{figure}

\newparagraphwithindent \textbf{Task Description.}
Value Linking is the cell-level counterpart to Schema Linking (Section \ref{sec:col_schema_linking}): given a value mentioned in a natural-language question, we retrieve the database columns whose cell values match it, either verbatim or through a synonym, abbreviation, typo, or date-format variant.
This tests whether embeddings bridge the surface gap between how a value is phrased in text and how it is stored in a database, a capability column-level matching cannot exercise, since it never grounds a query in a specific value.

\newparagraphwithindent \textbf{Datasets.}
We construct two evaluation sets from BIRD's \cite{li2023bird} gold-standard SQL and evidence annotations.
The \emph{Exact Match} set (1,354 queries) covers WHERE/HAVING comparisons whose literal value also appears verbatim in the question. 
The \emph{Fuzzy Matching} set (386 queries) is built primarily from BIRD's evidence field, where annotators wrote mappings such as ``X refers to \textit{column=Y}'' for translations not stated verbatim. 
This yields 232 synonyms (e.g. "female" → "F"), 57 abbreviations, 52 date-format mismatches, and 45 hand-curated typos (e.g. "Bsihopville" → "Bishopville").

\newparagraphwithindent \textbf{Evaluation Methodology.}
For each database, we embed the unique cell values of every column\footnote{We sample up to 1k distinct values per column and always make sure to include the queries' ground-truth values regardless of this cap, to guarantee that the correct answer is retrievable for every query}.
For every column, we retrieve its single best-matching indexed value and rank columns by that match's similarity score, rather than by any aggregate over all of a column's values to avoid biasing retrieval toward columns with more distinct values.
We report Mean Average Precision (MAP) as the primary metric, and also measure MRR and Recall@k.

\newparagraphwithindent \textbf{Experiments.}
Figure \ref{fig:exp_value_linking} shows MAP scores for the two dataset versions. 
GritLM achieves the highest score, SAP-RPT-1, HyTrel and TABBIE perform considerably worse.
GritLM, Granite R2, and MiniLM score higher on \emph{fuzzy} than \emph{exact}, while TUTA shows the opposite pattern: exact-match value phrases are literal substrings of the gold cell, whereas fuzzy phrases are paraphrases requiring world knowledge (e.g., "New York" → JFK). 
GritLM's fuzzy gain is larger than Granite's or MiniLM's, consistent with its instruction-tuned LLM backbone versus their smaller encoders.
Table-native encoders, pretrained on structural objectives, show no such gain.

%% file: sections_08_discussion.tex
\begin{table*}[t]
\centering
\caption{Overall Ranking of approaches across tasks. Each task cell shows the approach's rank followed by its raw metric value in parentheses; for Tabular Prediction this is the ELO delta from the initial rating of 1500. \text{---} indicates the approach does not support the task. $^\dagger$ indicates the approach could not complete all datasets for the task; missing datasets were imputed with a worst-case value. Overall Z-Score reports the mean, per-task z-score of the raw metric across approaches, preserving the magnitude of cross-approach differences that the rank-based Overall column collapses; approaches that do not support a task are assigned a z-score 1 standard deviation below the worst supporting approach on that task.}
\resizebox{\textwidth}{!}{%
\begin{tabular}{lccccccccccccc}
\toprule
Approach & \makecell{Row \\ Sim. Search \\ (MAP)} & \makecell{Triplet \\ Evaluation \\ (Accuracy) } & \makecell{Tabular \\ Prediction\\ (ELO)} & \makecell{Column \\ Sim. Search \\ (MAP)} & \makecell{Column Type \\ Annotation \\ (Macro F1)} & \makecell{Schema \\ Linking \\ (MAP)} & \makecell{Table \\ Retrieval \\ (MAP@10)} & \makecell{Table \\ Sim. Search \\ (MAP)} & \makecell{Table \\ Shuffling \\ (Accuracy)} & \makecell{Cell \\ Sim. Search \\ (MAP)} & \makecell{Value \\ Linking \\ (MAP)} & \makecell{Overall \\ Rank} & \makecell{Overall \\ Z-Score} \\
\midrule
GritLM & \phantom{0}\textbf{1}\phantom{$^\dagger$} (0.76) & \phantom{0}\textbf{1}\phantom{$^\dagger$} (0.88) & \phantom{0}9\phantom{$^\dagger$} \phantom{0}(-47) & \phantom{0}\textbf{1}\phantom{$^\dagger$} (0.48) & \phantom{0}\underline{2}\phantom{$^\dagger$} (0.60) & \phantom{0}\textbf{1}\phantom{$^\dagger$} (0.41) & \phantom{0}\textbf{1}\phantom{$^\dagger$} (0.67) & \phantom{0}\textbf{1}\phantom{$^\dagger$} (0.79) & \phantom{0}5\phantom{$^\dagger$} (0.41) & \phantom{0}3\phantom{$^\dagger$} (0.64) & \phantom{0}\textbf{1}\phantom{$^\dagger$} (0.65) & \textbf{2.36}\phantom{$^\dagger$} & \textbf{0.92} \\
IBM Granite R2 & \phantom{0}\underline{2}\phantom{$^\dagger$} (0.73) & \phantom{0}6\phantom{$^\dagger$} (0.81) & 11\phantom{$^\dagger$} (-158) & \phantom{0}\underline{2}\phantom{$^\dagger$} (0.44) & \phantom{0}4\phantom{$^\dagger$} (0.56) & \phantom{0}\underline{2}\phantom{$^\dagger$} (0.39) & \phantom{0}\underline{2}\phantom{$^\dagger$} (0.66) & \phantom{0}3\phantom{$^\dagger$} (0.76) & \phantom{0}7\phantom{$^\dagger$} (0.35) & \phantom{0}\textbf{1}\phantom{$^\dagger$} (0.73) & \phantom{0}\underline{2}\phantom{$^\dagger$} (0.62) & \underline{3.82}\phantom{$^\dagger$} & \underline{0.72} \\
MiniLM & \phantom{0}3\phantom{$^\dagger$} (0.66) & \phantom{0}4\phantom{$^\dagger$} (0.83) & 12\phantom{$^\dagger$} (-186) & \phantom{0}3\phantom{$^\dagger$} (0.41) & \phantom{0}6\phantom{$^\dagger$} (0.55) & \phantom{0}3\phantom{$^\dagger$} (0.37) & \phantom{0}3\phantom{$^\dagger$} (0.49) & \phantom{0}6\phantom{$^\dagger$} (0.71) & \phantom{0}6\phantom{$^\dagger$} (0.40) & \phantom{0}\underline{2}\phantom{$^\dagger$} (0.65) & \phantom{0}3\phantom{$^\dagger$} (0.56) & 4.64\phantom{$^\dagger$} & 0.53 \\
TaBERT & \phantom{0}5\phantom{$^\dagger$} (0.12) & \phantom{0}\underline{2}\phantom{$^\dagger$} (0.84) & 10\phantom{$^\dagger$} \phantom{0}(-86) & \phantom{0}7\phantom{$^\dagger$} (0.28) & \phantom{0}3\phantom{$^\dagger$} (0.59) & \phantom{0}4\phantom{$^\dagger$} (0.19) & \phantom{0}4\phantom{$^\dagger$} (0.06) & \phantom{0}\underline{2}\phantom{$^\dagger$} (0.77) & \phantom{0}8\phantom{$^\dagger$} (0.34) & \phantom{0}5\phantom{$^\dagger$} (0.38) & \phantom{0}5\phantom{$^\dagger$} (0.39) & 5.00\phantom{$^\dagger$} & -0.11 \\
TUTA & 12\phantom{$^\dagger$} (0.00) & 13\phantom{$^\dagger$} (0.56) & \phantom{0}6\phantom{$^\dagger$} (+105) & \phantom{0}4\phantom{$^\dagger$} (0.34) & \phantom{0}5\phantom{$^\dagger$} (0.56) & \phantom{0}5\phantom{$^\dagger$} (0.18) & \phantom{0}5\phantom{$^\dagger$} (0.03) & \phantom{0}5\phantom{$^\dagger$} (0.71) & \phantom{0}9\phantom{$^\dagger$} (0.31) & \phantom{0}4\phantom{$^\dagger$} (0.40) & \phantom{0}4\phantom{$^\dagger$} (0.51) & 6.55\phantom{$^\dagger$} & -0.28 \\
TABBIE & \phantom{0}9\phantom{$^\dagger$} (0.04) & \phantom{0}9\phantom{$^\dagger$} (0.68) & \phantom{0}5\phantom{$^\dagger$} (+118) & \phantom{0}6\phantom{$^\dagger$} (0.29) & \phantom{0}\textbf{1}\phantom{$^\dagger$} (0.61) & \phantom{0}7\phantom{$^\dagger$} (0.14) & \phantom{0}7\phantom{$^\dagger$} (0.01) & \phantom{0}8\phantom{$^\dagger$} (0.65) & \phantom{0}4\phantom{$^\dagger$} (0.51) & \phantom{0}9\phantom{$^\dagger$} (0.32) & \phantom{0}8\phantom{$^\dagger$} (0.16) & 6.64\phantom{$^\dagger$} & -0.38 \\
HyTrel & \phantom{0}7\phantom{$^\dagger$} (0.05) & 10\phantom{$^\dagger$} (0.66) & 13\phantom{$^\dagger$} (-393) & \phantom{0}8\phantom{$^\dagger$} (0.28) & \phantom{0}9\phantom{$^\dagger$} (0.45) & \phantom{0}8\phantom{$^\dagger$} (0.13) & \phantom{0}6\phantom{$^\dagger$} (0.01) & \phantom{0}7\phantom{$^\dagger$} (0.67) & \phantom{0}\textbf{1}\phantom{$^\dagger$} (0.93) & \phantom{0}8\phantom{$^\dagger$} (0.34) & \phantom{0}7\phantom{$^\dagger$} (0.16) & 7.64\phantom{$^\dagger$} & -0.53 \\
TARTE & 11\phantom{$^\dagger$} (0.02) & \phantom{0}5\phantom{$^\dagger$} (0.81) & \phantom{0}8\phantom{$^\dagger$} \phantom{0}(-44) & \phantom{0}5\phantom{$^\dagger$} (0.33) & \phantom{0}7\phantom{$^\dagger$} (0.54) & --- & --- & \phantom{0}4\phantom{$^\dagger$} (0.75) & \phantom{0}\underline{2}\phantom{$^\dagger$} (0.91) & \phantom{0}6\phantom{$^\dagger$} (0.35) & --- & 8.18\phantom{$^\dagger$} & -0.38 \\
SAP-RPT-1 & 10$^\dagger$ (0.02) & \phantom{0}8\phantom{$^\dagger$} (0.70) & \phantom{0}4\phantom{$^\dagger$} (+194) & \phantom{0}9\phantom{$^\dagger$} (0.15) & \phantom{0}8\phantom{$^\dagger$} (0.46) & \phantom{0}6\phantom{$^\dagger$} (0.14) & --- & --- & --- & \phantom{0}7\phantom{$^\dagger$} (0.35) & \phantom{0}6\phantom{$^\dagger$} (0.17) & 9.09\phantom{$^\dagger$} & -1.04 \\
TabICL v2 & \phantom{0}8\phantom{$^\dagger$} (0.04) & \phantom{0}7\phantom{$^\dagger$} (0.70) & \phantom{0}\textbf{1}\phantom{$^\dagger$} (+318) & 10\phantom{$^\dagger$} (0.11) & 10\phantom{$^\dagger$} (0.01) & --- & --- & --- & --- & 10\phantom{$^\dagger$} (0.16) & --- & 10.55\phantom{$^\dagger$} & -1.52 \\
TabDPT & \phantom{0}6\phantom{$^\dagger$} (0.05) & 12\phantom{$^\dagger$} (0.62) & \phantom{0}3\phantom{$^\dagger$} (+258) & --- & --- & --- & --- & \phantom{0}9\phantom{$^\dagger$} (0.40) & \phantom{0}3\phantom{$^\dagger$} (0.60) & --- & --- & 10.64\phantom{$^\dagger$} & -1.60 \\
TabuLa-8B & \phantom{0}4\phantom{$^\dagger$} (0.44) & \phantom{0}3\phantom{$^\dagger$} (0.84) & \phantom{0}7\phantom{$^\dagger$} \phantom{0}\phantom{0}(-5) & --- & --- & --- & --- & --- & --- & --- & --- & 11.45\phantom{$^\dagger$} & -1.70 \\
TabPFN v2.5 & 13$^\dagger$ (0.00) & 11\phantom{$^\dagger$} (0.63) & \phantom{0}\underline{2}\phantom{$^\dagger$} (+297) & --- & --- & --- & --- & --- & --- & --- & --- & 12.55\phantom{$^\dagger$} & -1.88 \\
\bottomrule
\end{tabular}
}
\label{tab:overall_ranking}
\end{table*}

\section{Discussion}
\label{sec:discussion_chapter}

\subsection{Overall Rankings}
\label{sec:ranking}
Other multi-task benchmarks, such as MTEB \cite{muennighoff2023mteb} or MMTU \cite{xing2025mmtu} aggregate task scores directly, as their tasks share comparable metrics (e.g., accuracy or F1).
\benchmark's eleven tasks instead span heterogeneous metrics, including ranking-based measures, classification accuracy, and regression errors (where lower values indicate better performance). 
Directly averaging these would require metric normalization and introduce assumptions about scale.
We therefore adopt rank-based aggregation: the Overall Rank in Table \ref{tab:overall_ranking} is the mean rank across tasks, weighted equally regardless of dataset count to avoid structural bias in the benchmark.
Because rank alone can obscure the size of gaps between approaches, we additionally report an Overall Z-Score.
Both penalize approaches that cannot perform a task at all, favoring models with broad task coverage, reflecting our goal of evaluating general-purpose embedding models.

\subsection{Discussion}
\label{sec:discussion}

\newparagraphwithindent \textbf{No single embedding generalizes across all tasks.}
Our experiments indicate that strong performance across \benchmark's tasks draws on at least three competencies: (1) semantic similarity, (2) predictive performance, and (3) structural fidelity to a table's row/column layout.
This is somewhat surprising, as it suggests that predictive models encode information different from what is required for structural or semantic similarity. 
Thus, specialized models for tabular prediction do not generalize to similarity-based tasks, and vice versa. \\
\textbf{Text embeddings excel at content, not structure.}
General-purpose text embeddings often solve similarity-based tasks effectively, outperforming specialized tabular models such as TabuLa-8B.
This is in line with \citet{boutaleb2025fishy}, who attribute a similar finding for table union search to benchmarks containing mostly vocabulary well covered by language models.
Yet this advantage is content-specific, not structural: our Table Shuffling Evaluation (Section~\ref{sec:table_shuffling}) shows that the same text embedders which dominate content-based similarity tasks are among the weakest at distinguishing a row/column-permuted table from one with scrambled cell values, while table-specialized architectures such as HyTrel and TARTE, otherwise weak on similarity tasks, excel here.
Variation studies (Sections \ref{sec:triplet_row_evaluation} and \ref{sec:cell_similarity_search}) further show that most models' accuracy shifts substantially under column-name removal or noise, while a few remain comparatively robust.
LLMs also struggle with enterprise data with domain-specific schemas and terminology \cite{bodensohn2025enterprise}, suggesting that general-purpose embeddings may face limitations in these settings.
Taken together, these results suggest model selection should be guided by whether the target task depends on content, structure, or predictive signal.

\newparagraphwithindent \textbf{Outlook.}
A universal table embedding thus remains an open challenge. 
Future work should explore hybrid approaches, combining embeddings optimized for similarity and prediction, as well as light-weight task-adaptive fine-tuning strategies to bridge the gap between general-purpose language understanding and tabular semantics and structure.

%% file: sections_09_conclusion.tex
\section{Conclusion}
\label{sec:conclusion}

We introduce \benchmark, the Tabular Embedding Test Bed, to evaluate tabular embeddings across a diverse set of application oriented tasks on different embedding levels.
Our evaluation shows that currently no single model consistently performs best across all tasks and levels.
Instead, performance depends strongly on the target application, with a clear distinction between models that excel at semantic similarity tasks and those optimized for predictive settings.
With our work, we guide users in selecting suitable models and we hope to provide the groundwork for the development of more general-purpose tabular representation models.

%% file: sections_10_appendix.tex



\subsection{Hardware}
\label{sec:appendix_hardware}
Our experiments were conducted on a server using the following setup: 
\begin{itemize}
    \item \textit{CPU:} AMD EPYC 9554P 64-Core Processor (128 threads) @ 3.1 GHz
    \item \textit{RAM:} 792 GB 
    \item \textit{GPU:} 1x NVIDIA A100 80GB PCIe
\end{itemize}

We also ran the Row Similarity Search (Section \ref{sec:row_embeddings_row_similarity_search}) experiments on cheaper hardware with the following setup:
\begin{itemize}
    \item \textit{CPU:} 2x Intel(R) Xeon(R) Gold 5120 CPU (28 cores / 56 threads total) @ 2.20GHz
    \item \textit{RAM:} 503 GB
    \item \textit{GPU:} 1x NVIDIA V100 16GB
\end{itemize}
\subsection{Toolkit Description}
\label{sec:toolkit}

Beyond the tasks and datasets described in Section \ref{sec:overview}, \benchmark~ is implemented as a runnable evaluation harness. Its central design challenge is that many approaches in Table \ref{tab:approaches_overview} are mutually incompatible as software, due to conflicting Python and torch/transformers dependencies, while the evaluation protocol applied to each of them must remain identical. 
This section describes the resulting architecture, the interface approaches implement to participate in the benchmark, and the toolkit's resource-measurement methodology.

\subsubsection{\textbf{Architecture.}}
\label{sec:toolkit_architecture}
TEmBed is structured around three concerns: configuration, orchestration, and execution. A configuration component holds the dataset registry, task defaults, approach declarations, and run specifications, all expressed as YAML files that determine which approaches, tasks, and datasets a given run covers. An orchestrator reads a run specification and expands it into individual jobs, one per approach-task-dataset combination, and dispatches them for execution. Task-specific evaluation drivers then carry out the protocol for a given task (Section \ref{sec:overview_tasks}), calling into each approach through a shared adapter interface (Section \ref{sec:toolkit_interface}). 
Adding a new approach to \benchmark~ requires only implementing this adapter interface for the representation levels the approach supports, and registering it via a configuration entry; no changes to the orchestrator, task drivers, or other approaches are needed.

The orchestrator is also responsible for resolving a key constraint: many approaches are mutually incompatible as software, due to conflicting Python and torch/transformers dependencies, so they cannot run within a single environment. Each approach declares the conda environment it requires; the orchestrator groups jobs by environment and executes each group as a separate subprocess, allowing a single benchmark run specification to combine approaches with incompatible dependencies without manual environment management.

\subsubsection{\textbf{Approach Interface.}}
\label{sec:toolkit_interface}

Each representation level is defined as a small abstract interface.\footnote{The interfaces and adapter examples for each representation level are documented in the repository README: https://github.com/IBM/table-representation-evals}
A task driver never imports an approach's implementation directly; instead, it resolves the approach's declared module path at run time and loads only the components that are present. 
An approach therefore only implements the adapters for the representation levels it supports, delegating preprocessing and encoding to its own model implementation rather than the interface itself.

Configuration follows the same registry structure introduced in Section \ref{sec:toolkit_architecture}: a global dataset registry, per-task protocol defaults, per-approach declarations, and a run specification. 
New datasets are added the same way, independent of any specific approach; the row-triplet perturbation variants described in Section \ref{sec:triplet_row_evaluation}, for instance, are implemented entirely as dataset-registry entries, and are evaluated against every approach without requiring approach-specific code.

\subsubsection{\textbf{Resource Measurement.}}
\label{sec:toolkit_resource}

Beyond task performance, the computational cost of an embedding approach is often decisive for whether it is practical to deploy, particularly for large datasets or resource-constrained environments. 
Every task driver therefore wraps both model setup and task inference in a shared monitoring component, so resource cost is measured identically across all approaches and tasks rather than through per-approach instrumentation. 
A monitoring thread within the job's own process samples the process tree at fixed intervals (every 100~ms), recording CPU utilization and resident memory summed over the job's process and its children, and per-process GPU memory restricted to the process IDs belonging to that job.
Since jobs are executed sequentially by the orchestrator (Section \ref{sec:toolkit_architecture}), this attribution is exact.

For each phase, the framework measures wall-clock time together with peak and mean CPU, RAM, and GPU memory usage; setup and inference are reported separately so that one-time model-loading cost is not conflated with per-query cost. 
Results are recorded at the same per-approach, per-task, per-dataset granularity as the primary performance metrics.


\begin{table}
    \centering
    \footnotesize 

    \begin{subtable}{\linewidth} 
    \centering
        \begin{tabular}{lcrcr}
            \toprule
            Dataset name & Type & \# Rows & \# Columns & \# Testcases  \\
            \midrule
            Amazon-Google & EM & 4,589 & 4 & 234 \\
            Beer & EM & 7,345 & 5 & 14 \\
            DBLP-ACM & EM & 4,910 & 5 & 444 \\
            DBLP-GoogleScholar & EM & 66,879 & 5 & 1,070 \\
            Fodors-Zagats & EM & 864 & 7 & 22 \\
            geological-settlements & C & 3,054  & 8 & 786 \\
            iTunes-Amazon & EM & 62,830 & 9 & 27\\
            MusicBrainz & C & 19,375 & 10 & 5,000\\
            Walmart-Amazon & EM & 24,628 & 6 & 193\\
            \bottomrule
        \end{tabular}
        \vspace{5px}
        \caption{Row Similarity Search Datasets (Section \ref{sec:row_embeddings_row_similarity_search}). Type \emph{EM} stands for datasets originally proposed for Entity Matching and \emph{C} for Clustering. The number of testcases results from the number of positive matching pairs in each dataset.}
        \label{tab:row_sim_dataset_statistics}
    \end{subtable}

    \begin{subtable}{\linewidth} 
        \centering
        \begin{tabular}{lrrrrr}
            \toprule
            Dataset name & \# Rows & \# Cols & \# Numerical Cols & \# Testcases  \\
            \midrule
            Wikidata Books & 1203 & 281 & 88* & 903\\
            Wikidata Astronomical Objects & 8034 & 41 & 27 & 2000 \\
            \bottomrule
        \end{tabular}
        \vspace{5px}
        \caption{Row Triplet Evaluation Datasets (Section \ref{sec:triplet_row_evaluation}). *The numerical columns of the Wikidata Books dataset are mostly identifiers, e.g. \emph{Goodreads work ID}. }
        \label{tab:row_triplet_statistics}
    \end{subtable}
    
    \begin{subtable}{\linewidth} 
        \centering
        \begin{tabular}{lrrrr}
            \toprule
            Dataset name & \# Tables & \# Rows & \# Columns & \# Testcases  \\
            \midrule
            NextiaJD  & 45+47 & 213,853 & 56.00 & 171+167 \\
            Valentine  & 576 & 14,418 & 21.96 & 288\\
            OpenData  & 3102 & 1,742 & 8.58 & 42\\
            Wikijoin-Small  & 659 & 44 & 2.54 & 100 \\
            \bottomrule
        \end{tabular}
        \vspace{5px}
        \caption{Column Similarity Search Datasets (Section \ref{sec:col_sim_search}). We report the average number of rows and columns per dataset.}
        \label{tab:col_sim_dataset_statistics}
    \end{subtable}

    \begin{subtable}{\linewidth}
        \centering
        \begin{tabular}{lrrrrr}
            \toprule
            Dataset name & \# Tables & \# Rows & \# Columns & \# Classes & \# Testcases \\
            \midrule
            SOTAB & 3,089 & 324.81 & 9.10 & 80 & 1,112 \\
            GitTables CTA & 814 & 47.28 & 18.00 & 122 & 532 \\
            \bottomrule
        \end{tabular}
        \vspace{5px}
        \caption{Column Type Annotation Datasets (Section \ref{sec:col_type_annotation}). \# Testcases is the number of labeled columns in the test split, which is what the classifier is evaluated on (trained on the labeled columns of the train split).}
        \label{tab:cta_dataset_statistics}
    \end{subtable}
    
    \begin{subtable}{\linewidth}
    \centering
        \begin{tabular}{lrrrr}
            \toprule
            Dataset name & \# Databases & \# Tables & \# Columns & \# Testcases \\
            \midrule
            BIRD Column Schema & 69 & 522 & 3,539 & 2,157 \\
            \bottomrule
        \end{tabular}
        \vspace{5px}
        \caption{Schema Linking Datasets (Section \ref{sec:col_schema_linking}). Derived from BIRD.}
        \label{tab:schema_linking_dataset_statistics}
    \end{subtable}

    \begin{subtable}{\linewidth} 
        \centering
        \begin{tabular}{lrrrr}
            \toprule
            Dataset name & \# Tables & \# Rows & \# Columns & \# Testcases  \\
            \midrule
            GitTables data lakes & 12687 & 283.62 & 11.43 & 348 \\
            \bottomrule
        \end{tabular}
        \vspace{5px}
        \caption{Table Similarity Search (Section \ref{sec:table_similarity_search}). }
        \label{tab:table_sim_dataset_statistics}
    \end{subtable}

    \begin{subtable}{\linewidth}
        \centering
        \begin{tabular}{lrrrr}
            \toprule
            Dataset name & \# Tables & \# Rows & \# Columns & \# Testcases \\
            \midrule
            FeTaQA & 2,003 & 14.16 & 5.70 & 2,003 \\
            TabFact & 1,695 & 13.43 & 6.25 & 12,779 \\
            OTT-QA & 789 & 12.40 & 4.72 & 2,214 \\
            Spider-Train & 795 & 20.42 & 5.11 & 6,997 \\
            Spider-Validation & 81 & 23.02 & 5.44 & 1,034 \\
            Spider-Test & 180 & 14.39 & 4.36 & 2,147 \\
            BIRD-Validation & 75 & 52,436.47 & 10.64 & 1,534 \\
            \bottomrule
        \end{tabular}
        \vspace{5px}
        \caption{Table Retrieval Datasets (Section \ref{sec:table_retrieval}) from the TARGET benchmark \cite{ji2025target_benchmark}.}
        \label{tab:table_retrieval_dataset_statistics}
    \end{subtable}
    
    \caption{Statistics of the included datasets, continued in Table \ref{tab:appendix_dataset_characteristics_continued}.}
    \label{tab:appendix_dataset_characteristics}
\end{table}

\begin{table}
    \centering
    \footnotesize 

    \begin{subtable}[t]{\linewidth}
        \centering
        \begin{tabular}{llrrrr}
            \toprule
            Dataset name & Source & \# Tables & \# Rows & \# Columns & \# Testcases \\
            \midrule
            FeTaQA & TARGET \cite{ji2025target_benchmark} & 500 & 17.33 & 6.03 & 3,278 \\
            TabFact & TARGET \cite{ji2025target_benchmark} & 1,692 & 14.90 & 6.35 & 10,565 \\
            OTT-QA & TARGET \cite{ji2025target_benchmark} & 500 & 12.47 & 4.72 & 3,090 \\
            Spider-Train & TARGET \cite{ji2025target_benchmark} & 500 & 31.03 & 6.46 & 3,114 \\
            CKAN Subset & LakeBench \cite{srinivas2023lakebench}& 5,000 & 53.71 & 11.83 & 41,436 \\
            ECB & LakeBench \cite{srinivas2023lakebench} & 1 & 22.00 & 23.00 & 7$^*$ \\
            \bottomrule
        \end{tabular}
        \vspace{5px}
        \caption{Table Shuffling Datasets (Section \ref{sec:table_shuffling}). Each base dataset is evaluated under a grid of 9 row/column-reorder perturbation variations (differing in perturbation type and magnitude), all at the same table-size window (5--100 rows, 3--30 columns). $^*$ECB's raw pool contains only 74 tables, almost all larger than this size window (median 500 rows, 34 columns), so only 1 table qualifies as an anchor and each variation yields at most 1 triplet.}
        \label{tab:table_shuffling_dataset_statistics}
    \end{subtable}  

    \begin{subtable}{\linewidth} 
        \centering
        \begin{tabular}{lrrrr}
            \toprule
            Dataset name & \# Tables & \# Rows & \# Columns & \# Testcases  \\
            \midrule
            s2abLE & 732 & 7.40 & 5.44 & 1000 \\
            \bottomrule
        \end{tabular}
        \vspace{5px}
        \caption{Cell Similarity Search Dataset (Section \ref{sec:cell_similarity_search}), adapted from \cite{lou2023s2abel}.}
        \label{tab:cell_dataset_statistics}
    \end{subtable}

    \begin{subtable}{\linewidth}
    \centering
    \begin{tabular}{lrrr}
        \toprule
        Dataset name & \# Databases & \# Indexed Cell Values & \# Testcases \\
        \midrule
        BIRD Cell Exact & 69 & 1,129,118 & 1,354 \\
        BIRD Cell Fuzzy & 51 & 960,582 & 386 \\
        \bottomrule
    \end{tabular}
    \vspace{5px}
    \caption{Value Linking Datasets (Section \ref{sec:value_linking}). Derived from BIRD.}
    \label{tab:value_linking_dataset_statistics}
\end{subtable}
    
    \caption{Statistics of the included datasets - continued. For the Tabular Prediction datasets from the TabArena Benchmark we refer to \cite{erickson2025tabarena}.}
    \label{tab:appendix_dataset_characteristics_continued}
    \vspace{-2em}
\end{table}

\subsection{Wikidata Hierarchies - Details on Dataset Construction}
\label{sec:appendix_dataset_wikidata}
In this section, we explain in detail how we construct the hierarchies and associated tables for the dataset proposed for the triplet-based row embedding evaluation in Section \ref{sec:row_embeddings_row_similarity_search}.

For the dataset, we leverage hierarchical relationships between classes in Wikidata \cite{wikidata}, as visualized in Figure \ref{fig:hierarchy_explained}.
We first build each hierarchy, one for literary genre data and one about astronomical object classes, and then the associated tables. 

\textbf{Collecting classes for the hierarchies}
Each hierarchy consists of \textit{class items}, that are structured into a tree-structure to define parent-child relationships between the classes, as visualized in Figure \ref{fig:hierarchy_explained}.  
To prepare the construction of a hierarchy from Wikidata, a list of classes is needed, as well as the relationships between these classes. 
For the genre hierarchy, therefore first all possible literary genres are retrieved from Wikidata.
Unfortunately, it's not straightforward to identify literary genres, as e.g. subclasses of \textit{genre (Q483394)}, \textit{fiction (Q8253)} or \textit{non-fiction (Q213051)} not only include literary genres but also genres used in other art forms. 
We therefore go a different route and collect books first and extract genres from concrete book items:
To identify all literary genres, all items that have the property \textit{instance of (P31)} \textit{literary work (Q7725634)} are retrieved using the Wikidata SPARQL endpoint. 
From each of the retrieved items, we extract the value for the property \textit{genre (P136)} to get a list of all literary genres. 
Wikidata is stored in a document-oriented database; its triplets do not form one connected graph. 
Therefore we experienced, that not all of the genre classes we retrieved in the previous step actually are connected to the the \textit{genre (Q483394)} class, which we will handle in the next section. 
We filter the list to keep only genre items that have an English label, and for genres that are connected to at least three books in Wikidata.
Next, we recursively retrieve the superclasses of each genre class by querying for its \textit{subclass of (P279)} values. 
We end up with 608 possible genres that could be included in the hierarchy. 
For astronomical objects the process is easier, as we can directly perform a breadth-first-search over all subclasses of the \emph{astronomical object (Q6999)} item in Wikidata, resulting in 143 possible astronomical object classes for the hierarchy.

\textbf{Building the hierarchies}
From the previous step, we obtain a set of candidate classes for the hierarchy together with their associated superclass relations, which form the basis for constructing the hierarchy.
For the genre hierarchy, we explicitly enforce a predefined root structure where the root node \textit{genre (Q483394)} has two direct children, \textit{fiction (Q8253)} and \textit{non-fiction (Q213051)}, to ensure a clear and consistent top-level split that facilitates the creation of reliable test cases.

To construct the hierarchy, we first identify all classes that appear as superclasses of at least one other class. 
These classes are treated as internal nodes, whereas classes that never occur as superclasses are considered leaf nodes.
The hierarchy is then constructed incrementally. Starting from the predefined root structure, classes are inserted one by one based on their superclass information. 
For each class, we determine the deepest valid insertion path in the current hierarchy by matching its superclasses, preferring the longest compatible path when multiple options exist. 
Classes that cannot yet be placed are revisited in subsequent iterations until no further insertions are possible.
Because insertion decisions are made greedily, some classes may initially be placed higher in the hierarchy than appropriate. To address this, we perform a reordering step in which nodes are iteratively relocated to deeper positions whenever a more specific valid superclass path becomes available. 
This process continues until no further improvements can be made.
As final outputs, we obtain a genre hierarchy with 111 genre classes and an astronomical object classes hierarchy with 113 included classes. 

\textbf{Building the associated tables}
We leverage Wikidata to build a table with real-world data associated to each of the two hierarchies. 
Corresponding to the literary genre hierarchy, we construct a table with information about books.
Analogous, we also build a table containing concrete astronomical objects.
We follow the process to create tables outlined in \cite{vogel2024wikidbs}.
The first step of the dataset creation already gave us a list of books from Wikidata, we discard books that are annotated with multiple genres. 
For the astronomical objects data, we retrieve all items from Wikidata where the \textit{instance of (P31)} property is the respective astronomical object, again keeping only items with an English label. 
These items are the rows for our tables. 
For each of the items, we collect all properties and values, which we use to fill the columns of the tables.
As the approach results in very broad and sparse tables, we prune columns where more than 95\% of the column values are missing. 
As final outputs, we obtain a table containing books with 1203 rows and 281 columns, and a table with astronomical objects, with 8034 rows and 41 columns. 

\textbf{Constructing Test cases}
Based on the constructed hierarchies and tables, we generate evaluation test cases for the two domains. 
Each test case follows a triplet format, as described in Section \ref{sec:triplet_row_evaluation}.

For the books domain, we explicitly split the hierarchy during construction into \textit{fiction} and \textit{non-fiction}.
Each book is assigned to one of these high-level categories based on its genre annotation and the hierarchy structure. 
To ensure consistency, we remove ambiguous cases where an author has written both fiction and non-fiction works. 
In such cases, only the majority category per author is retained and the other books are discarded from the table. 
This step avoids conflicting signals within the dataset.
After cleaning, the dataset is shuffled deterministically, and test cases are generated using the hierarchy. Specifically, pairs of books are considered similar if they share closely related genres within the hierarchy, while dissimilar pairs are sampled from different branches (e.g., fiction vs. non-fiction or distant subgenres). The hierarchy thus directly defines the notion of semantic similarity.

For the astronomy domain, we construct test cases based on high-level semantic branches derived from object types (e.g., galaxies, nebulae, star clusters, and variable stars). 
Test cases are then generated by pairing objects within the same branch as similar examples, while dissimilar examples are drawn from different branches. To ensure diversity and reproducibility, pairs are generated deterministically and distributed across branches in a balanced manner.

In total, we create 903 test cases for the books table and 2000 for the astronomical objects table.


\subsection{Additional Results.}
In this section, we include additional results and analysis that go
beyond the main paper.

\begin{table*}[t]
\centering
\begin{tabular*}{\textwidth}{@{\extracolsep{\fill}} lcccccc @{}}
\toprule
Approach & MAP & VRAM A100 (GB) & VRAM V100 (GB) & Speedup (A100 vs V100) & Speedup (A100 vs CPU) & Peak RAM CPU (GB) \\
\midrule
GritLM & 0.760 & 78.5 & OOM & --- & --- & impractical \\
HyTrel & 0.046 & 58.2 & OOM & --- & 5.38$\times$ & 38.8 \\
IBM Granite R2 & 0.728 & 3.1 & 1.4 & 1.58$\times$ & 15.20$\times$ & 11.1 \\
MiniLM & 0.657 & 1.1 & 0.7 & 1.63$\times$ & 5.60$\times$ & 1.3 \\
SAP-RPT-1 & 0.031 & 33.9 & OOM & --- & 20.51$\times$ & 41.3 \\
TABBIE & 0.036 & 3.3 & 3.2 & 1.07$\times$ & 8.24$\times$ & 5.7 \\
TabDPT & 0.051 & 1.0 & 0.9 & 1.75$\times$ & 2.68$\times$ & 5.0 \\
TaBERT & 0.116 & 1.0 & 0.8 & 1.16$\times$ & 3.70$\times$ & 2.3 \\
TabICL v2 & 0.043 & 3.7 & 3.2 & 1.53$\times$ & 1.65$\times$ & 5.0 \\
TabPFN v2.5 & 0.002 & 3.1 & 2.0 & 2.62$\times$ & --- & impractical \\
TabuLa-8B & 0.444 & 61.4 & OOM & --- & --- & impractical \\
TARTE & 0.016 & 0.7 & 0.5 & 1.78$\times$ & 1.76$\times$ & 11.8 \\
TUTA & 0.004 & 3.2 & 3.1 & 3.28$\times$ & 23.95$\times$ & 6.7 \\
\bottomrule
\end{tabular*}
\caption{Row Similarity Search: hardware comparison across an A100 80GB, a V100 16GB, and CPU-only (2$\times$ Intel Xeon Gold 5120, no GPU, 504GB RAM available). OOM in a VRAM column indicates the approach ran out of VRAM on that hardware, either before completing any dataset or partway through the largest datasets; in both cases the corresponding speedup column shows --- since a full-run comparison against the A100 isn't available. \emph{impractical} indicates the approach was still running after several hours and was not run to completion. \emph{Peak RAM CPU} is the highest system-memory usage observed on the CPU-only run, across all 9 datasets}
\label{tab:row_sim_hardware_comparison}
\vspace{-2em}
\end{table*}

\subsubsection{Case Study: Row Similarity Search on More Realistic Hardware}
\label{sec:hardware_v100}
To complement the resource measurements in Section \ref{sec:triplet_row_evaluation}, which were collected on a single NVIDIA A100 80GB, we repeat Row Similarity Search on two more constrained hardware settings, described in Section \ref{sec:appendix_hardware}: a 16GB V100 and a CPU-only configuration on the same host. We choose Row Similarity Search since it is the task for which we already report execution time and GPU memory in the main paper (Figure \ref{fig:quadrant_row_sim_vram}, \ref{fig:quadrant_row_sim_time}), allowing a direct comparison. 
The CPU-only host provides 56 threads and 504GB RAM, well above what minimal or budget hardware typically offers, so execution time results on this host represent a best case rather than a worst case for CPU-only inference. 
To assess whether this host's large RAM capacity masks memory constraints a smaller machine would hit, we report peak RAM usage per approach separately.

\paragraph{V100: Execution time and VRAM.}
GritLM and TabuLa-8B, the two largest models by parameter count, fail to complete the V100 run entirely: their model weights alone approach (~14GB GritLM) and exceed (~32GB TabuLa-8B) the V100's 16GB capacity, leaving no room for the activations required at inference. HyTrel and SAP-RPT-1 complete their smaller datasets on the V100 but run out of memory on the same largest datasets that already required close to or above 16GB of peak VRAM on the A100.

\paragraph{CPU-only.}
Execution time increases between 1.65× (TabICL v2) and 23.95× (TUTA) relative to the A100 for the approaches that complete the full CPU run. TabPFN v2.5 does not run on CPU with more than 1,000 rows by default, so it completes only the smallest dataset, and its reported time ratio reflects that dataset alone. GritLM does not complete a run on the smallest dataset within several hours, so we report no CPU results for it. TabuLa-8B completes fewer datasets within our compute budget due to its per-row runtime on CPU. Peak RAM stays below 41.3GB for every approach across all datasets, well below what a machine with 64GB RAM would provide, showing that CPU-only feasibility for Row Similarity Search is bound by execution time rather than memory.

\subsubsection{Case Study: An End-to-End Pipeline Comparison
for Table Retrieval}
\label{sec:PNEUMA-gritlm}

\benchmark's model inclusion criterion (Section~\ref{sec:overview_approaches}) restricts the main comparison to approaches that produce general-purpose, reusable embeddings across multiple tasks. 
This deliberately excludes end-to-end systems built for a single task, such as PNEUMA \cite{balaka2025pneuma}, an LLM-based table discovery pipeline that combines hybrid BM25/ dense retrieval with an LLM re-ranking pass. 
Such systems answer a different question than \benchmark's main benchmark: not ``how well does this embedding transfer across tasks'', but ``how well does a pipeline purpose-built for one task perform at it''. 
Since Table Retrieval (Section~6.2) is a task in TEmBed for which such a specialized, LLM-based pipeline is readily available, we report a standalone comparison here between PNEUMA \cite{balaka2025pneuma} and GritLM \cite{muennighoff2024gritlm}, the best-performing approach on that task among our benchmarked models, as an exemplary case study of how a task-specialized pipeline stacks up against a general-purpose embedding on its native task.

\paragraph{Setup.} We run PNEUMA via its officially supported OpenAI-API backend (\texttt{gpt-4o-mini} for schema narration and LLM-Judge re-ranking, \texttt{text-embedding-3-small} for the vector index) with \texttt{table\_row\_limit=100}, matching GritLM's row-limited variant so both approaches see the same table content. 
Since PNEUMA embeds table content internally as multiple per-block vectors (schema narration and sampled-row summaries) rather than a single fixed-size table-level embedding, and exposes only a final ranked table list as output rather than these embeddings directly, it does not fit our harness's retrieval interface and is instead run as a standalone script against the same corpora, queries, and gold labels as three datasets from Table Retrieval (Section~\ref{sec:table_retrieval}): \texttt{bird-validation}, \texttt{spider-validation}, and \texttt{ottqa}.

\begin{table}
\centering
\resizebox{\columnwidth}{!}{%
\begin{tabular}{llccccc}
\toprule
Dataset & Approach & MRR@1 & MRR@10 & MAP@10 & Recall@10 & Precision@10 \\
\midrule
\multirow{2}{*}{bird-validation}
 & PNEUMA & \textbf{0.757} & 0.848 & 0.709 & 0.896 & \textbf{0.232} \\
 & GritLM & 0.728 & \textbf{0.829} & \textbf{0.716} & \textbf{0.933} & 0.180 \\
\midrule
\multirow{2}{*}{spider-validation}
 & PNEUMA & \textbf{0.807} & \textbf{0.883} & \textbf{0.813} & \textbf{0.956} & \textbf{0.192} \\
 & GritLM & 0.712 & 0.816 & 0.750 & 0.936 & 0.142 \\
\midrule
\multirow{2}{*}{ottqa}
 & PNEUMA & 0.367 & 0.501 & 0.501 & 0.814 & 0.093 \\
 & GritLM & \textbf{0.780} & \textbf{0.861} & \textbf{0.861} & \textbf{0.981} & \textbf{0.098} \\
\bottomrule
\end{tabular}%
}
\caption{PNEUMA vs.\ GritLM on Table Retrieval, matched \texttt{table\_row\_limit=100} setting. Bold marks the better result within each pair.}
\label{tab:PNEUMA_gritlm}
\vspace{-2em}
\end{table}

\paragraph{Discussion.} 
The result (Table \ref{tab:PNEUMA_gritlm}) varies by query domain rather than favoring one approach across the board.
On the two text-to-SQL datasets (\texttt{bird-validation}, \texttt{spider-validation}), where questions are tightly coupled to schema and column structure, PNEUMA matches or clearly outperforms GritLM, consistent with its LLM-based schema-and-row reasoning suiting that query style.
On \texttt{ottqa}, an open-domain QA setting over Wikipedia tables, GritLM performs decisively better across every metric.
We take this as evidence that a task-specialized, end-to-end pipeline does not straightforwardly outperform a general-purpose embedding on the pipeline's own native task; its advantage is domain-specific rather than general.
We note that the API-based PNEUMA configuration was considerably slower to run than GritLM (roughly 200--500$\times$ across the three datasets), driven by its LLM-Judge issuing one sequential re-ranking call per candidate.
PNEUMA's own reported efficiency (it is shown to be faster than comparable baselines in the original paper) uses a different, locally batched LLM configuration that we did not reproduce.

\begin{figure}
    \centering
    \includegraphics[width=\linewidth]{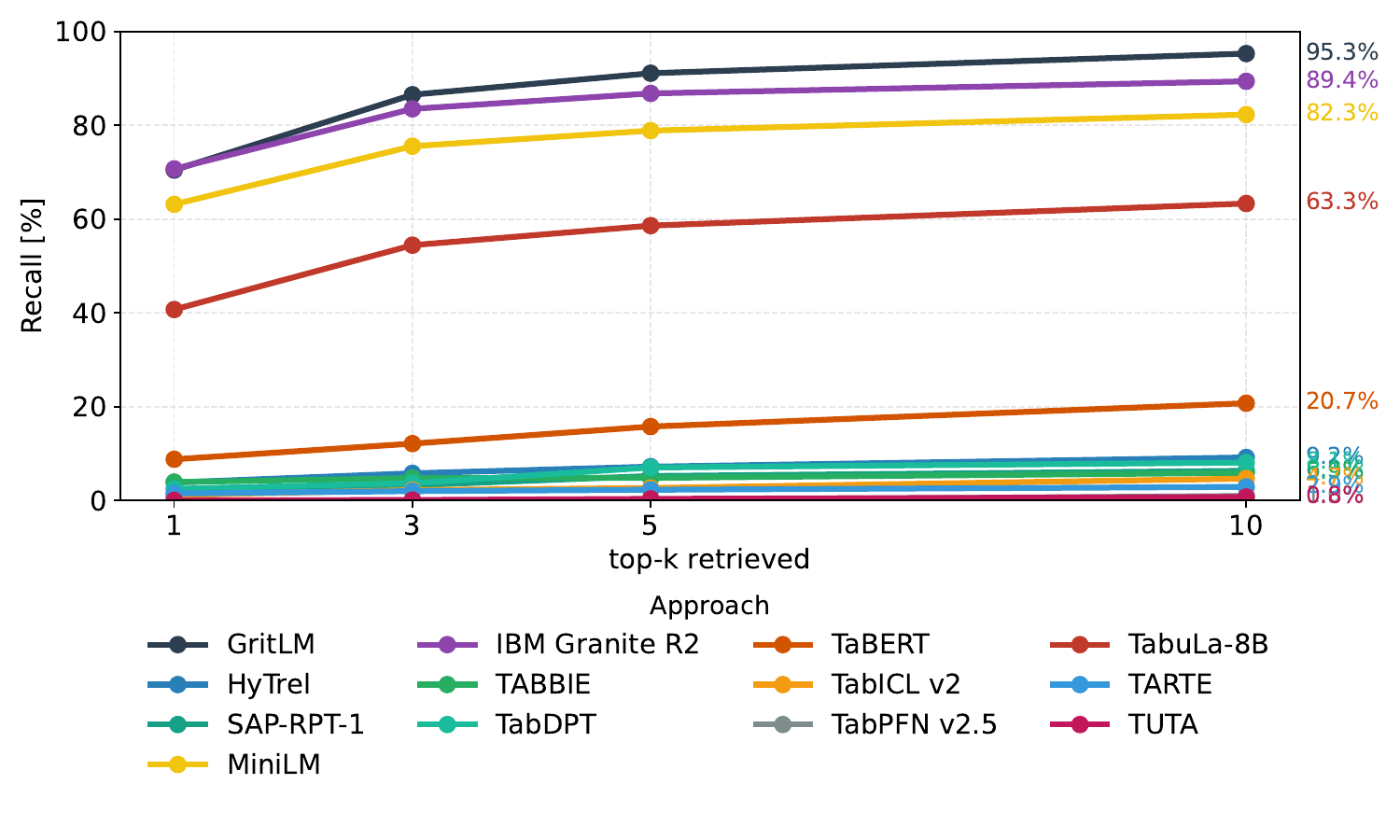}
    \caption{Row Similarity Search: Top-k recall results, aggregated over seven datasets.}
    \label{fig:exp_row_sim_topk}
    \vspace{-1em}
\end{figure}


\begin{table}[t]
\centering
\small
\setlength{\tabcolsep}{4pt}
\resizebox{\columnwidth}{!}{%
\begin{tabular}{p{0.45\columnwidth}rrr}
\toprule
Model & ELO Score & $\Delta$ & \#Comp \\
\midrule
TabICL v2 & 1818 & 318 & 14620 \\
TabPFN v2.5 & 1797 & 297 & 13840 \\
TabDPT & 1758 & 258 & 14840 \\
SAP-RPT-1 & 1694 & 194 & 14840 \\
Baseline & 1620 & 120 & 14840 \\
TABBIE & 1618 & 118 & 14840 \\
TUTA & 1605 & 105 & 14380 \\
TabuLa-8B* & 1495 & -5 & 14400 \\
TARTE & 1456 & -44 & 14340 \\
GritLM & 1453 & -47 & 14840 \\
TaBERT & 1414 & -86 & 14840 \\
IBM Granite R2 & 1342 & -158 & 14840 \\
TabDPT* & 1335 & -165 & 14840 \\
MiniLM & 1314 & -186 & 14840 \\
TabICL v2* & 1113 & -387 & 14620 \\
HyTrel & 1107 & -393 & 12000 \\
\bottomrule
\end{tabular}
}
\caption{Tabular Prediction: ELO scores calculated from pairwise comparisons, where higher values indicate better performance. All models are initialized at 1500 and updated over 20 rounds. $\Delta$ reports the change from the initial rating, and we also list the number of comparisons, which differs since not all models ran on every dataset.}
\label{tab:tabular_prediction_elo}
\vspace{-2em}
\end{table}

\subsection{Approaches - Evaluation Details}
\label{sec:appendix_approach_details}

\paragraph{HyTrel} \cite{chen2023hytrel} 
\newline
\textit{Hyperparameters.} For HyTrel, we use the model checkpoint trained on contrastive loss, made available with the source code on \href{https://github.com/awslabs/hypergraph-tabular-lm}{Github}, which provides us with a 768 dimensional output embedding for rows, columns, tables and cells. Since HYTREL relies on sentence embedding inputs for the cell content as input to the hypergraph, we keep its default pre-processing pipeline for text serialization and tokenization before the model forward pass. This process cleans up cells which are textual to extract content and converts numerical inputs to scientific notation before text tokenization. Input column hyper-edge embeddings are the column names, and the table hyper-edge embeddings are obtained from tokenzing the table caption/header. Row embeddings need to be randomly initialized and acts as conduits for information filtering during the forward pass. Embedding representations for HYTREL do not require access to labels during inference or in-context learning.

\medskip
\noindent
\textit{Extracting Embeddings.} HYTREL is based on SetTransformer operations, applied to nodes (table cells) and hyperedges (column, row, and table identities of each cell) obtained from the table grid. HYTREL takes as input the entire table, and outputs filtered node embeddings (for cell content) and hyper-edge embeddings (row, column, table) of the same size as the input sentence transformer embeddings. At inference, the final output embeddings are designed to be context-aware (i.e aware of the underlying grid structure in the table) and are permutation equivariant (agnostic to ordering of information in the rows/columns). During evaluation, row, table, and column embedding based tasks perform a single forward pass through the HYTREL model.

The base HYTREL model was not directly fine-tuned for predictive ML tasks due to which we evaluate the ability of the row embeddings to serve as representations needed for prediction. Therefore, one forward pass through the model is performed during the training to obtain the row-embedding input for the training data-points used to fit XGBoost (and other predictors in the benchmark). A second forward pass is performed on the table where test data points are appended to the training data to obtain test row embeddings for inference. This two-phase procedure ensures consistency in comparison with the in-context learning setting for the transformer based foundation models such as TabICL, TabPFN, SAP-RPT-OSS, and TabuLA.

\paragraph{TABBIE \cite{iida2021tabbie}}
\newline
\textit{Hyperparameters.} 
For TABBIE, we use the released \emph{mix} checkpoint,
pretrained on a mixture of Wikipedia tables and WDC web tables and made available with the source code on \href{https://github.com/SFIG611/tabbie}{Github}.
TABBIE passes a table through two alternating BERT-base transformers: one attending across columns within each row (the \emph{row transformer}), one
attending across rows within each column (the \emph{column transformer}), each producing a 768-dimensional output.
Each cell is tokenized to at most 16 BERT tokens.
The model's positional embeddings impose hard per-forward-pass limits of 29 data rows and 20 columns; tables that exceed either limit are handled by non-overlapping row or column windows whose outputs are aggregated, as described below.
Embedding extraction requires no labels or additional fine-tuning.

\medskip
\noindent
\textit{Extracting Embeddings.}
TABBIE augments the input with two sets of learnable \texttt{[CLS]} tokens: one per row (placed at column position 0) and one per column (placed at row position 0), so every forward pass jointly produces a row-level and a column-level summary vector for each row and column in the window.
Concretely, the row transformer's output tensor has shape $(n_\text{rows}{+}2)\times(n_\text{cols}{+}1)\times 768$, where the first two
row slots hold the table-level CLS and the header row, and the first column slot holds the per-row CLS token.
The column transformer produces a tensor of the same shape with analogous semantics. 
\emph{Row embeddings} are the per-row CLS vectors from the row transformer (\texttt{row\_embs[i+2, 0, :]} for data row $i$), which capture the full
cross-column context of that row.
For tables wider than 20 columns, we adopt TABBIE's own default behavior of truncating to the first 20 columns when computing row embeddings. 
\emph{Column embeddings} are the per-column CLS vectors from the column transformer (\texttt{col\_embs[0, j+1, :]} for column $j$), which capture the full cross-row context of that column.
For tables taller than 29 rows, column embeddings from each non-overlapping row window are averaged. 
\emph{Table embeddings} are the mean of all row embeddings, L2-normalised; for table retrieval, a natural-language query is embedded by wrapping it in a single-column, single-row table and passing it through the same pipeline. \emph{Cell embeddings} concatenate the row- and column-transformer outputs at each cell position, yielding 1536-dimensional vectors.
Specifically, the embedding for cell $(i, j)$ is $[\texttt{row\_embs}[i{+}1, j{+}1,~:]\; \|\; \texttt{col\_embs}[i{+}1, j{+}1, :]]$, where the $+1$ offsets skip the table-CLS row/column.
The header row (row 0 of the output matrix) is produced by averaging the header position's contribution across all row windows that cover the table; data rows belong to exactly one row window and require no averaging.
Cell embeddings therefore have output shape $(n_\text{rows}{+}1)\times n_\text{cols} \times 1536$, with row 0 providing column-level representations and rows $1\ldots n_\text{rows}$ providing per-cell representations.

\paragraph{TaBERT \cite{yin2020tabert}}
\newline
\textit{Hyperparameters.} 
We use the released \emph{TaBERT\_Base\_(K=1)} checkpoint, pretrained jointly on natural language utterances and a large corpus of web tables drawn from CommonCrawl and Wikipedia, made available with the source code on \href{https://github.com/facebookresearch/TaBERT}{Github}, which provides us with a 768-dimensional (BERT-base sized) output embedding for rows, columns, tables, and cells. 
TaBERT represents a table as a fixed schema (column names and an inferred type, \texttt{real} for numeric columns and \texttt{text} otherwise) together with a \emph{content snapshot}: one representative cell value per column. 
Since our embedding tasks provide no accompanying natural-language utterance, we condition every forward pass on an empty context string, so only the schema and content snapshot drive the representation. Embedding extraction requires no labels or additional fine-tuning.

\medskip
\noindent
\textit{Extracting Embeddings.} 
TaBERT's core operation returns one contextualized embedding per column (\texttt{column\_encoding}) for whatever table (schema plus content snapshot) is passed through a forward pass; row, table, and cell embeddings are all derived from this single primitive by varying what is used as the content snapshot and how the resulting column encodings are aggregated.
\emph{Column embeddings} are obtained with a single forward pass over the full table, using each column's first non-null value as its content snapshot; the returned per-column vectors are used directly. 
\emph{Row embeddings} require one forward pass per row: for row $i$, the content snapshot substitutes every column's value from row $i$, and the resulting column encodings are mean-pooled into a single row vector. 
Because each row is encoded from its own values and the fixed schema alone, without TaBERT ever seeing the other rows, row embeddings are independent across rows, unlike hypergraph- or attention-based table encoders whose row representations are conditioned on the rest of the table. \emph{Table embeddings} are the mean of the column embeddings from the column-embedding forward pass followed by scaling to unit norm.
For table retrieval, the natural-language query is embedded through the same pipeline by treating it as a single-column, single-row table. \emph{Cell embeddings} reuse the per-row forward passes computed for row embeddings, without the final mean-pool: the column encoding for column $j$ from row $i$'s forward pass is taken as the embedding of cell $(i,j)$, with the column-embedding forward pass over the full table providing a header/column-level embedding for row $0$.

As with HyTrel, the base TaBERT model was not fine-tuned for predictive ML tasks, so we evaluate its row embeddings as the input representation for the downstream predictor (XGBoost and others in the benchmark).
Because row embeddings depend only on a row's own values and the fixed table
schema, train and test row embeddings can be extracted via separate forward
passes over each split without any risk of train-test leakage.

\medskip
\noindent\paragraph{TUTA \cite{wang2021tuta}}
\newline
\textit{Hyperparameters} 
TUTA~\citep{wang2021tuta} is a tree-based transformer pre-trained on generally structured tables (including spreadsheets with hierarchical headers) using three self-supervised objectives: masked language modeling, cell-value prediction, and table context retrieval. 
The backbone follows the BERT-base architecture (12 transformer layers, $d_\text{model} = 768$, 12 attention heads, 3{,}072-dim feedforward) but replaces standard positional encodings with tree-based structural encodings that capture relative row/column distance within a 4-level tree hierarchy
(node degree $[32, 32, 64, 256]$, total tree nodes $= 384$).
Attention is biased additively by tree distance rather than multiplicatively, allowing the model to distinguish topologically proximate cells (same row, same column, same block) even in irregularly shaped tables. 
Tables are serialised into a flat token sequence: a [CLS] token is prepended, followed by one [SEP] separator token and up to $B$ body tokens per cell ($B = 8$ in our experiments), tokenised with \texttt{bert-base-uncased}.
Each token carries four integer-valued numerical features (magnitude, precision, top digit, low digit) that let the model distinguish numeric magnitudes without requiring explicit number tokenisation. 
TUTA additionally defines an 11-dimensional format feature vector, which we leave at zero in our benchmark since we do not extract cell-level formatting metadata from the source tables.
The full sequence is capped at 512 tokens.

\medskip
\noindent \textit{Embedding extraction.}
We use the \emph{TUTA} checkpoint variant (implicit tree positional encoding).
For tables with more than $R = 50$ rows, the table is processed in chunks of $R$ data rows each; the header row is included in every chunk. \emph{Cell embeddings} are obtained by mean-pooling the backbone hidden states over the body token positions of each cell, yielding a tensor of shape
$(N_\text{rows} + 1) \times N_\text{cols} \times 768$, where row~0 contains the column-header embeddings from the first chunk. 
\emph{Row embeddings} are the column-wise mean of the cell embeddings for each data row, producing a matrix of shape $N_\text{rows} \times 768$. \emph{Table embeddings} are taken from the \texttt{[CLS]} hidden state at sequence position 0, using only the first $R$-row chunk. 
For predictive ML tasks, row embeddings are passed as fixed features to the downstream classifier without fine-tuning the TUTA backbone.


\paragraph{TabuLa}\cite{gardner2024tabula} 
\newline
\textit{Hyperparameters.} 
TabuLa-8B processes inputs with a maximum sequence length of 2048 tokens, matching the context length recommended by TabuLa-8B's own evaluation code. 
For predictive modeling tasks, we use 8 randomly selected training examples as few-shot demonstrations, the shot count reported in the TabuLa-8B paper for a context length of 2048 tokens.
Text generation uses sampling with a temperature of 0.1 and generates up to 50 new tokens per prediction. The model leverages AutoModelForCausalLM architecture, which provides both embedding extraction capabilities through its base transformer layers and text generation for direct prediction tasks.

\medskip
\noindent
\textit{Extracting Embeddings.} For row embeddings, each table row is converted to a structured string format where column-value pairs are concatenated with separators (e.g., "col1: val1 | col2: val2"). The model tokenizes these strings and extracts embeddings from the last hidden state of the base transformer using average pooling across sequence positions, producing fixed-dimensional row representations, which can be used for row-based tasks. 
TabuLa-8B does not provide a mechanism to expose cell or table embeddings, due to which we skip those tasks.

For predictive ML tasks, TabuLA-8B operates in few-shot mode: it constructs prompts containing 8 randomly sampled training examples with their labels, followed by the test instance requiring prediction. The model generates predictions autoregressively, producing class labels for classification tasks or numerical values for regression. This approach leverages the model's pre-trained knowledge of tabular data patterns without requiring task-specific fine-tuning, enabling zero-shot generalization across diverse tabular datasets while maintaining computational efficiency through batched processing and GPU acceleration. However, we noticed that the output of the model often fails to adhere to the required format needed to parse labels reliably, or produces skewed outputs (always predicts majority class due to sampling bias) leading to poor performance for downstream evaluation. Such issues with instruction following and bias were also noted in other studies~\cite{gorla2026illusion} on TabuLA-8b. Therefore, we chose to report predictive performance based on row embeddings and not few-shot prompting, as it was more reliable in experiment.

\paragraph{TabPFN v2.5}\cite{hollmann2025tabpfn} 
\newline
\textit{Hyperparameters.} TabPFN (Tabular Prior-Fitted Networks) is a transformer model pre-trained on synthetic tabular datasets that performs in-context learning for tabular prediction tasks. Our implementation leverages TabPFN for both row embedding extraction and direct predictive modeling. For both tasks, we keep the default hyperparameters which use multiple ensembles (n estimators = 32) during the forward pass. During pre-processing, categorical columns are converted into numerical codes via label encoding with missing values filled by -1 and NaNs set to 0. TabPFNv2.5 also has a constraint in terms of allowing a maximum of 10000 training samples and 100 columns, due to which some evaluations cannot be completed.

\medskip
\noindent
\textit{Extracting Embeddings.} 
For row-embedding tasks such as entity matching and the triplet based evaluation (where no ground truth label exists), row embeddings are extracted through the models get-embeddings() API. Since this requires specifying labels which are not available for this task, we set row labels to 0s. Due to its alternating row-column grid attentional structure, TabPFN does not implement methods for obtaining column, table or cell embeddings in its implementation, due to which these tasks were skipped.

Predictive modeling with the TabPFNv2.5 code is straightforward, with a call to model.fit() on the training data, followed by model.predict() on the test samples. For row embedding extraction, TabPFN's API requires that labels be specified for both train and test data. We also experiment using row-embeddings from TabPFN chained with predictors such as XGBoost for predictive ML. In this experiment, to prevent data-leakage from test labels, we tried two settings. In the first, real labels are given for the training data and zeros are provided for the testing examples. In the second setting, both training and testing data are given dummy zero labels (constant zero embedding bias during encoding). We observed that using row embeddings from the latter results is better performance-wise, likely due to the posterior predictive distribution being less distorted by the differential label encoding for training and test. We report only these results in the main manuscript.


\paragraph{TabICL}\cite{qu2025tabicl} 
\newline
\textit{Hyperparameters.} TabICL also uses an ensemble of 32 estimators (default) with the tabicl-classifier-v2-20260212.ckpt checkpoint for classification tasks and tabicl-regressor-v2-20260212.ckpt for regression tasks. Our benchmark also allows for configuring the v1 checkpoint (which supports all tasks available with v2 except regression) with a single change to the model yaml file. TabICLv2 processes tabular data after converting categorical and string columns to numerical codes via label encoding after applying outlier removal and standard scaling. All missing values are filled with zeros. The model architecture includes a column embedder that generates contextualized representations for each cell and a row interactor that aggregates column embeddings into row-level representations. All tasks except regression use the classifier checkpoint for embedding extraction. 
For memory efficiency, the implementation supports batched processing with configurable batch sizes (default: 8)

\medskip
\noindent
\textit{Extracting Embeddings.} TabICL requires a call to model.fit() with provided labels (or dummy 0 labels similar to tabpfn). It then extracts embeddings through a two-stage process: the column embedder generates cell-level representations by processing the input table as tensors. This produces embeddings of shape (batch, rows, columns, embedding dim) using its column embedder. Next, TabICL's row interactor aggregates these column embeddings across the feature dimension to produce final row embeddings. Thus, cell and column representations from the column embedder before and after mean pooling are used for cell and column level tasks respectively, while row-embedding based tasks can use the outputs of the row extractor. As with other transformer based architectures, labels need to be provided to the model to extract embeddings due to which we provide 0 as the dummy label for non-predictive ML tasks. Table-level embeddings are not supported by TabICL as its in-context learning is table specific.  

For predictive ML tasks, TabICL operates in direct prediction mode where the model as fitted against the training labels generates class probabilities for classification (returning logits) or continuous values for regression using its predictproba() and predict() methods respectively. The v2 checkpoints support both classification and regression, while v1 checkpoints only support classification. For the experiments evaluating row-embeddings for prediction, during inference, a train size parameter needs to be provided before embedding extraction. The model then fits on the first train size rows (masking attentions between test-points and labels) and extracts embeddings  for the remaining test rows, enabling proper train-test separation during in-context learning.  As with TabPFN, we experimented by providing real and dummy 0 labels for the row-embedding extraction for prediction and observed the same representational collapse with the mismatched setting for train and test. The results we report on row-embedding based prediction supply 0 as the label for both test and train data.

\paragraph{SAP-RPT-1}\cite{spinaci2025contexttab} 
\newline
\textit{Hyperparameters.} The SAP-RPT-1-OSS (formerly ConTextTab) implementation uses a semantics-aware in-context learning approach with configurable bagging and context size parameters. The model operates bagging=8 (ensemble of 8 models) and supports context sizes of 8192 tokens (for optimal performance on high-memory GPUs). All input data is preprocessed by converting all columns to string format to preserve semantic information, as SAP-RPT-1-OSS is designed to work with textual representations of tabular data. The model maintains separate classifier and regressor instances, each initialized with the specified bagging and max-context-size parameters. For all embedding extraction cases, labels (real or dummy) need to be provided to the model API. 
\newline

\noindent
\textit{Extracting Embeddings.} For row embeddings, SAP-RPT-1-OSS employs an in-context learning paradigm where the model tokenizes the input table into context (training rows with labels) and query (test rows) portions. The tokenizer processes both portions through the model's embedding layer, followed by a multi-layer in-context encoder with attention mechanisms that allow query rows to attend to context rows. Cell embeddings are extracted from the encoder's output by taking the query portion's representations (excluding the target column token), averaging across feature column tokens produces fixed-dimensional row embedding vectors for row embedding tasks. Due to the symmetric nature of the architecture, an analogous procedure can be used to extract query column embeddings. Dummy 0 labels are provided to the API for each of these tasks. Table embeddings are not supported by SAP-RPT-1-OSS.

For predictive ML tasks, the model operates in direct prediction mode: after fitting on training data with provided labels, it generates class probabilities for classification via predictproba() or continuous predictions for regression via predict(). At train time, all rows serve as both context and query with real labels, while during inference, the training rows provide labeled context for embedding the remaining test rows. For the row-embedding based evaluation for predictive ml tasks, we followed the same experimental protocol as TabICL and TabPFN and observed better performance when dummy labels were provided consistently to train and test.

\medskip
\noindent
\paragraph{TabDPT \cite{ma2025tabdpt}}
\newline
\textit{Hyperparameters.} For TabDPT, we use the \emph{tabdpt1\_2.safetensors} checkpoint, downloaded automatically from the \href{https://huggingface.co/Layer6/TabDPT}{HuggingFace repository \texttt{Layer6/TabDPT}} together with the \href{https://github.com/layer6ai-labs/TabDPT-inference}{source code}, which provides us with an output embedding of width 512 (the pretrained model's embedding dimension, read from the checkpoint's config) for rows and tables. TabDPT is an in-context learner: at inference time a set of context rows is passed as key/value source, and query rows attend to this context via cross-row attention, with the context labels fused into the value vectors at every one of the model's transformer layers through a per-layer label encoder. We use up to \texttt{context\_size=1024} context rows per forward pass. 
Because our embedding tasks provide no accompanying supervised split, unsupervised extraction uses up to \texttt{context\_size} rows of the table itself as context, paired with a near-constant dummy label: all context rows are labelled zero except a single row flipped to one, the minimal label
variation that satisfies TabDPTClassifier's requirement of at least two classes. 
The model still produces useful contextual row representations from its learned feature interactions, since the context rows condition the attention pattern independently of whether their labels carry real information.
When true training labels are available (e.g. for row-embedding evaluation with a train/test split), the training rows and their real labels are used as context instead. Embedding extraction requires no fine-tuning and reuses TabDPT's own preprocessing pipeline (imputation, scaling, and PCA-based feature reduction for wide tables) exactly as at inference time.

\medskip
\noindent
\textit{Extracting Embeddings.} TabDPT has no separate embedding extraction mode in its public API — it exposes only classification/regression logits — so we patch the loaded model's forward pass to also return the pre-head hidden state for every query row, replicating the model's own forward computation (feature encoding, outlier clipping and normalisation, optional "thinking" tokens, and all transformer layers with per-layer label fusion) but returning the final-layer representations \emph{before} the output head instead of after it. \emph{Row embeddings} are obtained by first fitting TabDPT's estimator pipeline on the context rows to populate its imputer, scaler, and (for wide tables) PCA projection, then applying the same fitted transforms to the query rows; context and query features are padded to the model's maximum feature width, concatenated along the sequence axis, and passed through the patched forward pass, with query rows processed in chunks of \texttt{context\_size} to bound memory. The pre-head hidden state of each query row is taken directly as its row embedding. \emph{Table embeddings} are the mean of the row embeddings across all rows of the table; TabDPT has no CLS token or other table-level token, so mean-pooling is the natural aggregation, mirroring the approach used for other architectures without a dedicated table token. \emph{Cell and column embeddings are not available}: TabDPT's encoder projects every feature of a row into a single 512-dimensional vector at the input (a single \texttt{Linear(num\_features} $\to$ 512 layer) before any attention is computed, so no per-feature or per-column representation exists anywhere in the network; tasks requiring cell- or column-level embeddings are excluded from the TabDPT configuration. \emph{Predictive-ML} evaluation bypasses embedding extraction entirely and instead uses the standard \texttt{TabDPTClassifier}/\texttt{TabDPTRegressor} scikit-learn interface directly (\texttt{fit} on the training split, \texttt{predict\_proba}/\texttt{predict} on the test split), since TabDPT's in-context-learning mechanism already handles prediction natively without requiring any modification to the model.

\medskip 
\noindent
\paragraph{TARTE \cite{kim2025tarte}}
\newline
\textit{Hyperparameters.}  For TARTE, we use the released pretrained checkpoint (\emph{tarte\_pretrained\_weights.pt}), pretrained on knowledge-graph-derived table entries from Wikidata and YAGO4.5 and made available on \href{https://huggingface.co/inria-soda/tarte}{HuggingFace} together with the \href{https://github.com/soda-inria/tarte-ai}{source code}, which provides us with a 768-dimensional output embedding for rows, columns, tables, and cells. TARTE represents each row independently as a small graph: every column name and categorical cell value is embedded with a 300-dimensional FastText model (\emph{cc.en.300.bin}), while numerical and datetime cells are represented as the scalar value scaling the FastText embedding of their column name; a learnable readout token is prepended to this per-row sequence, and a 3-layer, 24-head transformer performs cross-column self-attention within the row (rows are never attended to jointly). We average the readout token's hidden state across all 3 transformer layers to obtain the row embedding, following the default \texttt{layer\_index="all"} setting. Embedding extraction requires no labels or additional fine-tuning; consistent with the dummy-label convention used for other in-context models in our benchmark, TARTE is always called with $y=\text{None}$, since its encoder does not condition on labels regardless.

\medskip
\noindent
\textit{Extracting Embeddings.} 
TARTE's core operation returns the full per-row hidden-state sequence (readout token plus one token per non-null cell) for whatever rows are passed through a forward pass.
row, table, cell, and column embeddings are all derived from this single primitive. \emph{Row embeddings} are obtained by fitting a fresh table-specific pre-processor (which fits a power transform on numerical columns and builds a FastText lookup for that table's schema) and running the transformer over the row sequences in batches of 128 rows to bound the $O(N^2)$ attention cost; the readout token's layer-averaged hidden state is taken directly as each row's embedding. \emph{Table embeddings} are the mean of the row embeddings; because TARTE has no dedicated table-level or CLS token, mean-pooling over rows is the natural aggregation, and TARTE cannot embed natural-language queries, so retrieval tasks requiring query encoding are excluded. \emph{Cell embeddings} reuse the full hidden-state sequence rather than only the readout token: for each row, we recover the sequence position assigned to each column (categorical columns first, then numerical, then datetime, following TARTE's fixed node ordering) and take the transformer's hidden state at that position as the cell's embedding, with missing/NaN cells assigned a zero vector; a header/column-level embedding for row $0$ is obtained by averaging each column's cell embeddings over all data rows, since TARTE likewise has no separate header token. \emph{Column embeddings} follow the same per-position extraction, but pool over a sample of up to 50 rows instead of all rows for efficiency, averaging the transformer outputs at each column's sequence position across the sampled rows.

\paragraph{all-MiniLM-L6-v2}\cite{wang2020minilm},  \paragraph{IBM Granite R2}\cite{awasthy2025granite-r2}  and \paragraph{GritLM}\cite{muennighoff2024gritlm}: 
As language model encoders, these models process textual input. 
We therefore convert tabular data into text prior to encoding. 

\vspace{0.5em}\noindent
\textit{Hyperparameters.}
We use the models \emph{all-MiniLM-L6-v2} and \emph{ibm-granite/granite-embedding-english-r2} via the sentence-transformer library \cite{reimers2019sentencetransformer} and GritLM via its Python package\footnote{https://pypi.org/project/gritlm/}.
Default configurations are employed for all models, with no additional hyperparameter tuning.

\vspace{0.5em}\noindent
\textit{Extracting Embeddings.}
Textual inputs for each embedding level are constructed as follows:

\textit{Row Embeddings:}
Each row is embedded individually by concatenating column headers and cell values into a single string
:
\begin{quote}
\texttt{'<header\_text\_0>: <column\_value\_0>; ...; \\  <header\_text\_n>: <column\_value\_n>'}. 
\end{quote}

\textit{Column Embeddings:}
For each table, we embed every column individually by concatenating the column header with all distinct column values in the following way: 
\begin{quote}
\texttt{'<header\_text>: <column\_value\_0> | ... | <column\_value\_n>'}
\end{quote}

\textit{Table Embeddings:}
To create table embeddings, we linearize each table into a markdown string, similar to \cite{ji2025target_benchmark}.
For the variants that include example table rows, we include the first 100 rows from the table.  

\textit{Cell Embeddings:}
For each table, we embed every cell individually by converting the cell value to a string and combining it with the column header in the following way:
\emph{'<header\_text>: <cell\_value>'}.


\subsection{Result Tables.}
\label{sec:appendix_result_tables}
In this section, we include the full result tables for all experiments from the main paper. 

\clearpage
\begin{table*}[t]
\centering
\resizebox{\textwidth}{!}{%
\begin{tabular}{lccccccccccccc}
\toprule
Dataset & GritLM & HyTrel & IBM Granite R2 & MiniLM & SAP-RPT-1 & TABBIE & TabDPT & TaBERT & TabICL v2 & TabPFN v2.5 & TabuLa-8B & TARTE & TUTA \\
\midrule
Amazon-Google & 0.57 & 0.05 & \textbf{0.67} & \underline{0.58} & 0.02 & 0.00 & 0.02 & 0.11 & 0.00 & 0.00 & 0.35 & 0.01 & 0.00 \\
Beer & \textbf{1.00} & 0.07 & \textbf{1.00} & \underline{0.86} & 0.00 & 0.00 & 0.09 & 0.05 & 0.00 & 0.00 & 0.45 & 0.00 & 0.00 \\
DBLP-ACM & \textbf{0.98} & 0.16 & \textbf{0.98} & \underline{0.96} & 0.02 & 0.00 & 0.00 & 0.17 & 0.00 & 0.00 & 0.95 & 0.01 & 0.00 \\
DBLP-GoogleScholar & \underline{0.61} & 0.03 & \textbf{0.62} & 0.60 & - & 0.00 & 0.16 & 0.18 & 0.19 & - & 0.48 & 0.00 & 0.00 \\
Fodors-Zagats & \textbf{1.00} & 0.04 & \textbf{1.00} & \underline{0.94} & 0.09 & 0.31 & 0.12 & 0.37 & 0.04 & 0.01 & 0.87 & 0.00 & 0.02 \\
MusicBrainz & \textbf{0.99} & 0.02 & \textbf{0.99} & \underline{0.96} & 0.00 & 0.00 & 0.00 & 0.01 & 0.00 & 0.00 & 0.36 & 0.00 & 0.00 \\
Walmart-Amazon & \underline{0.75} & 0.05 & \textbf{0.79} & 0.72 & 0.10 & 0.00 & 0.06 & 0.12 & 0.00 & 0.00 & 0.50 & 0.11 & 0.00 \\
geological-settlements & \textbf{0.65} & 0.00 & \underline{0.23} & 0.07 & 0.00 & 0.01 & 0.01 & 0.03 & 0.17 & 0.00 & 0.00 & 0.01 & 0.01 \\
iTunes-Amazon & \textbf{0.29} & 0.00 & \underline{0.27} & 0.23 & - & 0.00 & 0.01 & 0.00 & 0.00 & - & 0.03 & 0.01 & 0.00 \\
\midrule
Mean & \textbf{0.76} & 0.05 & \underline{0.73} & 0.66 & 0.03 & 0.04 & 0.05 & 0.12 & 0.04 & 0.00 & 0.44 & 0.02 & 0.00 \\
\bottomrule
\end{tabular}%
}
\caption{Row Similarity Search (Section \ref{sec:row_embeddings_row_similarity_search}): MAP results per dataset for all approaches. - indicates that the approach could not be run on the dataset, mostly due to memory constraints.}
\label{tab:row_sim_per_dataset}
\end{table*}

\begin{table*}[h]
\centering
\resizebox{\textwidth}{!}{%
\begin{tabular}{lrccccccccccccc}
\toprule
Dataset & \#Cols & GritLM & HyTrel & IBM Granite R2 & MiniLM & SAP-RPT-1 & TABBIE & TabDPT & TaBERT & TabICL v2 & TabPFN v2.5 & TabuLa-8B & TARTE & TUTA \\
\midrule
Wikidata Astronomical Objects & 41 & \underline{0.93} & 0.75 & \textbf{0.94} & 0.89 & 0.80 & 0.68 & 0.65 & 0.86 & 0.77 & 0.67 & 0.92 & 0.83 & 0.80 \\
\ensuremath{\hookrightarrow} \textsubscript{TEXT ONLY} & 14 & \underline{0.92} & 0.79 & \textbf{0.94} & 0.91 & 0.89 & 0.70 & 0.66 & 0.88 & 0.74 & 0.76 & 0.90 & 0.88 & 0.63 \\
\midrule
Wikidata Books & 281 & \textbf{0.84} & 0.57 & 0.68 & 0.77 & 0.59 & 0.69 & 0.59 & \underline{0.81} & 0.64 & 0.59 & 0.76 & 0.79 & 0.32 \\
\ensuremath{\hookrightarrow} \textsubscript{NO COLUMN NAMES} & 281 & \textbf{0.82} & 0.57 & 0.78 & \underline{0.81} & 0.60 & 0.68 & 0.59 & 0.78 & 0.64 & 0.59 & 0.79 & 0.76 & 0.42 \\
\ensuremath{\hookrightarrow} \textsubscript{SIMPLER COLUMN NAMES (NO PID)} & 281 & \textbf{0.82} & 0.57 & 0.72 & 0.73 & 0.60 & 0.69 & 0.60 & 0.78 & 0.64 & 0.59 & \underline{0.79} & 0.77 & 0.38 \\
\ensuremath{\hookrightarrow} \textsubscript{NO GENRE COLUMN} & 280 & \textbf{0.77} & 0.56 & 0.64 & 0.70 & 0.59 & 0.66 & 0.58 & \underline{0.75} & 0.61 & 0.60 & 0.74 & \underline{0.75} & 0.32 \\
\ensuremath{\hookrightarrow} \textsubscript{ONLY 5 COLUMNS} & 5 & \underline{0.89} & 0.55 & \textbf{0.90} & 0.83 & 0.83 & 0.68 & 0.55 & 0.87 & 0.54 & 0.57 & 0.81 & 0.85 & 0.59 \\
\midrule
Mean & - & \textbf{0.85} & 0.62 & 0.80 & 0.81 & 0.70 & 0.68 & 0.60 & \underline{0.82} & 0.65 & 0.63 & 0.81 & 0.80 & 0.49 \\
\bottomrule
\end{tabular}%
}
\caption{Triplet-Based Row Embedding Evaluations (Section \ref{sec:triplet_row_evaluation}): Accuracy per dataset for all approaches. - indicates that the approach could not be run on the dataset, mostly due to memory constraints.}
\label{tab:triplet_per_dataset}
\end{table*}


\begin{table*}[t]
\centering
\resizebox{\textwidth}{!}{%
\begin{tabular}{lcccccccccccccccc}
\toprule
Dataset & XGBoost & Granite* & GritLM* & HyTrel* & MiniLM* & SAP-RPT & TABBIE* & TabDPT & TabDPT* & TaBERT* & TabICL v2 & TabICL v2* & TabPFN v2.5 & TabuLa-8B* & TARTE* & TUTA* \\
\midrule
APSFailure & \underline{0.99} & 0.96 & 0.97 & - & 0.97 & \underline{0.99} & \textbf{1.00} & \underline{0.99} & 0.91 & 0.98 & \textbf{1.00} & 0.50 & - & - & \underline{0.99} & - \\
Amazon\_employee\_a. & 0.82 & 0.68 & 0.74 & 0.50 & 0.72 & \textbf{0.85} & 0.66 & \underline{0.84} & 0.58 & 0.76 & \textbf{0.85} & 0.50 & \underline{0.84} & 0.73 & 0.67 & 0.83 \\
Bank\_Customer\_Chu. & 0.85 & 0.76 & 0.81 & 0.50 & 0.73 & 0.87 & \underline{0.91} & 0.87 & 0.85 & 0.77 & 0.88 & 0.48 & 0.88 & 0.80 & 0.81 & \textbf{1.00} \\
Bioresponse & \underline{0.88} & 0.66 & 0.78 & - & 0.60 & 0.87 & 0.76 & 0.87 & 0.54 & 0.71 & 0.87 & 0.49 & \textbf{0.89} & 0.80 & - & 0.72 \\
Diabetes130US & 0.63 & 0.61 & 0.60 & - & 0.60 & 0.65 & \textbf{0.71} & \underline{0.67} & 0.57 & 0.59 & \underline{0.67} & 0.50 & \underline{0.67} & 0.61 & 0.56 & 0.60 \\
E-CommereShippingDa. & 0.74 & 0.64 & 0.71 & 0.51 & 0.67 & 0.75 & \underline{0.94} & 0.74 & 0.78 & 0.72 & 0.74 & 0.51 & 0.74 & 0.70 & 0.74 & \textbf{1.00} \\
Fitness\_Club & 0.76 & 0.74 & 0.72 & 0.53 & 0.68 & \underline{0.81} & 0.80 & \underline{0.81} & 0.64 & 0.75 & \underline{0.81} & 0.50 & \underline{0.81} & 0.77 & 0.77 & \textbf{1.00} \\
GiveMeSomeCredit & 0.86 & 0.79 & 0.83 & - & 0.77 & \underline{0.87} & \textbf{1.00} & \underline{0.87} & 0.71 & 0.81 & \underline{0.87} & 0.50 & - & 0.84 & 0.84 & \textbf{1.00} \\
HR\_Analytics\_Job\. & 0.80 & 0.77 & 0.77 & 0.51 & 0.77 & 0.81 & \underline{0.94} & 0.81 & 0.77 & 0.76 & 0.82 & 0.50 & 0.81 & 0.78 & 0.68 & \textbf{0.95} \\
Is-this-a-good-cust. & 0.67 & 0.67 & 0.70 & 0.50 & 0.69 & 0.75 & \underline{0.80} & 0.74 & 0.59 & 0.69 & 0.74 & 0.53 & 0.74 & 0.67 & 0.62 & \textbf{0.89} \\
Marketing\_Campaign & 0.92 & 0.74 & 0.81 & 0.51 & 0.75 & 0.89 & 0.92 & 0.93 & 0.75 & 0.82 & \underline{0.94} & 0.48 & \underline{0.94} & 0.86 & 0.89 & \textbf{0.96} \\
NATICUSdroid & \underline{0.98} & 0.97 & 0.94 & 0.48 & 0.88 & \textbf{0.99} & \textbf{0.99} & \textbf{0.99} & 0.97 & 0.95 & \textbf{0.99} & 0.48 & \textbf{0.99} & 0.97 & \underline{0.98} & 0.94 \\
bank-marketing & 0.75 & 0.71 & 0.72 & 0.51 & 0.70 & 0.77 & \underline{0.96} & 0.77 & 0.68 & 0.72 & 0.78 & 0.52 & 0.77 & 0.72 & 0.66 & \textbf{0.99} \\
blood-transfusion-s. & 0.65 & 0.65 & 0.63 & 0.45 & 0.62 & 0.68 & \underline{0.74} & 0.73 & 0.51 & 0.71 & \underline{0.74} & 0.53 & \underline{0.74} & 0.69 & 0.67 & \textbf{0.96} \\
churn & 0.93 & 0.62 & 0.73 & 0.52 & 0.66 & 0.93 & 0.73 & \underline{0.94} & \textbf{0.95} & 0.71 & \underline{0.94} & 0.49 & \underline{0.94} & 0.78 & 0.89 & 0.86 \\
coil2000\_insurance. & 0.71 & 0.63 & 0.67 & - & 0.61 & 0.73 & 0.67 & \underline{0.74} & 0.66 & 0.62 & \textbf{0.76} & 0.49 & \textbf{0.76} & 0.70 & 0.67 & 0.69 \\
credit-g & 0.77 & 0.73 & 0.78 & 0.53 & 0.74 & 0.80 & \underline{0.81} & 0.78 & 0.52 & 0.74 & 0.79 & 0.50 & 0.79 & 0.75 & 0.63 & \textbf{0.95} \\
credit\_card\_clien. & 0.77 & 0.72 & 0.75 & 0.49 & 0.67 & 0.79 & \textbf{0.99} & 0.79 & 0.72 & 0.73 & 0.79 & 0.50 & 0.79 & 0.75 & 0.75 & \underline{0.94} \\
customer\_satisfact. & \underline{0.99} & 0.93 & 0.97 & - & 0.92 & \underline{0.99} & \textbf{1.00} & \underline{0.99} & 0.96 & 0.96 & \textbf{1.00} & 0.51 & - & \underline{0.99} & 0.97 & \textbf{1.00} \\
diabetes & 0.81 & 0.70 & 0.82 & 0.44 & 0.70 & 0.84 & 0.82 & \underline{0.85} & 0.70 & 0.76 & \underline{0.85} & 0.42 & \underline{0.85} & 0.80 & 0.82 & \textbf{0.99} \\
hazelnut-spread-con. & \underline{0.97} & 0.71 & 0.80 & 0.48 & 0.75 & \textbf{0.99} & 0.88 & \textbf{0.99} & 0.88 & 0.82 & \textbf{0.99} & 0.48 & \textbf{0.99} & 0.82 & 0.96 & 0.85 \\
heloc & 0.77 & 0.75 & 0.75 & 0.51 & 0.72 & 0.80 & \textbf{0.96} & 0.80 & 0.86 & 0.74 & 0.80 & 0.52 & 0.80 & 0.76 & 0.76 & \underline{0.93} \\
in\_vehicle\_coupon. & 0.83 & 0.75 & 0.77 & 0.49 & 0.77 & 0.77 & 0.82 & 0.84 & 0.78 & 0.79 & \underline{0.85} & 0.50 & \underline{0.85} & 0.78 & 0.71 & \textbf{0.86} \\
jm1 & 0.73 & 0.70 & 0.71 & 0.51 & 0.70 & 0.75 & \textbf{0.98} & 0.79 & 0.74 & 0.71 & 0.79 & 0.51 & 0.77 & 0.72 & 0.72 & \underline{0.86} \\
kddcup09\_appetency & 0.77 & 0.65 & 0.54 & - & 0.55 & \underline{0.81} & 0.59 & 0.72 & 0.53 & 0.54 & - & - & \textbf{0.82} & - & 0.69 & 0.65 \\
online\_shoppers\_i. & 0.92 & 0.85 & 0.90 & 0.49 & 0.86 & 0.93 & \textbf{1.00} & 0.93 & 0.97 & 0.90 & 0.93 & 0.49 & 0.93 & 0.89 & 0.88 & \underline{0.99} \\
polish\_companies\. & 0.96 & 0.82 & 0.87 & 0.51 & 0.71 & \underline{0.97} & \textbf{0.98} & 0.92 & 0.90 & 0.86 & 0.95 & 0.52 & \textbf{0.98} & 0.91 & 0.82 & 0.84 \\
qsar-biodeg & 0.91 & 0.84 & 0.90 & 0.47 & 0.77 & 0.94 & \textbf{0.98} & 0.94 & 0.60 & 0.87 & 0.94 & 0.51 & 0.94 & 0.90 & 0.93 & \underline{0.95} \\
seismic-bumps & 0.74 & 0.73 & 0.73 & 0.47 & 0.66 & 0.78 & \textbf{0.97} & 0.78 & 0.67 & 0.76 & 0.80 & 0.49 & 0.79 & 0.75 & 0.74 & \underline{0.94} \\
taiwanese\_bankrupt. & 0.94 & 0.74 & 0.84 & 0.53 & 0.65 & 0.94 & \textbf{0.98} & 0.94 & 0.90 & 0.83 & \underline{0.95} & 0.51 & \underline{0.95} & 0.84 & 0.91 & 0.74 \\
\midrule
Mean & 0.83 & 0.74 & 0.78 & 0.50 & 0.72 & 0.84 & \underline{0.88} & 0.85 & 0.74 & 0.77 & 0.86 & 0.50 & 0.84 & 0.79 & 0.78 & \textbf{0.89} \\
\bottomrule
\end{tabular}%
}
\caption{Tabular Prediction (Section \ref{sec:tabular_prediction}): Binary Classification Results per Dataset measured by ROC AUC Score (higher is better). * indicates that the results were obtained by training XGBoost on top of row embeddings.}
\label{tab:tabular_prediction_binary}
\end{table*}


\begin{table*}[t]
\centering
\resizebox{\textwidth}{!}{%
\begin{tabular}{lcccccccccccccccc}
\toprule
Dataset & XGBoost & Granite* & GritLM* & HyTrel* & MiniLM* & SAP-RPT & TABBIE* & TabDPT & TabDPT* & TaBERT* & TabICL v2 & TabICL v2* & TabPFN v2.5 & TabuLa-8B* & TARTE* & TUTA* \\
\midrule
MIC & 0.64 & 0.91 & 0.81 & 1.71 & 0.97 & \textbf{0.46} & 0.71 & \underline{0.47} & 1.01 & 0.82 & \underline{0.47} & 1.02 & \textbf{0.46} & 0.73 & 0.68 & 0.75 \\
SDSS17 & 0.09 & 0.39 & 0.12 & - & 0.33 & 0.08 & 0.23 & \underline{0.07} & 0.43 & 0.21 & \underline{0.07} & 1.91 & - & 0.13 & 0.12 & \textbf{0.06} \\
anneal & 0.04 & 0.09 & 0.18 & 2.58 & 0.17 & \textbf{0.02} & 0.28 & 0.05 & 1.16 & 0.16 & \underline{0.03} & 1.58 & \textbf{0.02} & 0.08 & 1.01 & 0.13 \\
hiva\_agnostic & 0.27 & 0.28 & 0.24 & - & 0.21 & \textbf{0.18} & 0.27 & \textbf{0.18} & 0.26 & 0.28 & \underline{0.20} & 0.21 & \textbf{0.18} & 0.30 & - & - \\
maternal\_health\_r. & 0.41 & 0.40 & 0.41 & 3.49 & 0.40 & 0.50 & 0.46 & 0.37 & 1.53 & 0.48 & \underline{0.36} & 2.84 & 0.42 & 0.43 & 0.59 & \textbf{0.24} \\
splice & 0.10 & 0.62 & 0.46 & 2.48 & 0.49 & 0.09 & 0.51 & 0.19 & 1.16 & 0.52 & \textbf{0.07} & 2.74 & \underline{0.08} & 0.42 & 1.03 & 0.31 \\
students\_dropout\. & 0.64 & 1.13 & 0.93 & 2.96 & 1.07 & \underline{0.55} & 0.64 & 0.57 & 0.78 & 0.97 & \underline{0.55} & 1.42 & \textbf{0.54} & 0.93 & 0.92 & 0.74 \\
website\_phishing & 0.31 & 0.45 & 0.48 & 3.64 & 0.46 & 0.54 & 0.30 & \underline{0.21} & 0.81 & 0.46 & \textbf{0.20} & 1.52 & \underline{0.21} & 0.43 & 0.90 & 0.33 \\
\midrule
Mean & 0.31 & 0.54 & 0.45 & 2.81 & 0.51 & 0.30 & 0.42 & \underline{0.26} & 0.89 & 0.49 & \textbf{0.24} & 1.65 & 0.27 & 0.43 & 0.75 & 0.37 \\
\bottomrule
\end{tabular}%
}
\caption{Tabular Prediction (Section \ref{sec:tabular_prediction}): Multiclass Classification Results per Dataset measured by Log Loss Score (lower is better). * indicates that the results were obtained by training XGBoost on top of row embeddings.}
\label{tab:tabular_prediction_multiclass}
\end{table*}

\begin{table*}[t]
\centering
\resizebox{\textwidth}{!}{%
\begin{tabular}{lrrrrrrrrrrrrrrrr}
\toprule
Dataset & XGBoost & Granite* & GritLM* & HyTrel* & MiniLM* & SAP-RPT & TABBIE* & TabDPT & TabDPT* & TaBERT* & TabICL v2 & TabICL v2* & TabPFN v2.5 & TabuLa-8B* & TARTE* & TUTA* \\
\midrule
Another-Dataset-on. & 785.58 & 1425.24 & 1157.47 & 2357.67 & 1453.25 & 701.13 & 1059.64 & 688.66 & 1655.06 & 1138.33 & \underline{684.24} & 2169.51 & \textbf{683.78} & 1084.08 & 826.71 & 1608.56 \\
Food\_Delivery\_Tim. & 7.61 & 9.21 & 8.46 & 10.65 & 9.01 & 7.98 & 8.92 & \textbf{7.51} & 9.37 & 8.65 & 7.65 & 12.51 & \underline{7.60} & 8.25 & 8.29 & 9.05 \\
QSAR-TID-11 & 0.90 & 1.17 & 1.44 & - & 1.48 & 0.90 & 1.70 & \textbf{0.82} & 1.67 & 1.44 & 0.86 & 2.09 & \underline{0.83} & 1.26 & 1.08 & 1.40 \\
QSAR\_fish\_toxicit. & 0.98 & 1.29 & 1.08 & 1.75 & 1.30 & \underline{0.90} & 1.25 & 0.91 & 1.42 & 1.17 & \textbf{0.89} & 1.67 & \underline{0.90} & 1.10 & 0.99 & 1.43 \\
airfoil\_self\_nois. & 1.59 & 4.89 & 3.92 & 8.93 & 5.29 & 1.17 & 5.84 & 1.05 & 6.30 & 3.95 & \underline{0.99} & 11.78 & \textbf{0.97} & 3.70 & 1.91 & 6.38 \\
concrete\_compressi. & 4.80 & 11.10 & 9.00 & 22.25 & 12.84 & 4.05 & 11.06 & 4.06 & 15.74 & 8.76 & \textbf{3.67} & 21.08 & \underline{3.82} & 9.79 & 5.80 & 15.55 \\
diamonds & 568.03 & 1474.13 & 902.78 & 5441.49 & 1319.28 & 574.47 & 1455.75 & 515.44 & 2953.18 & 1113.29 & \underline{504.49} & 6982.71 & \textbf{503.14} & 761.96 & 1429.80 & 2220.10 \\
healthcare\_insuran. & 5016.73 & 7696.37 & 5565.27 & 15026.74 & 6698.41 & 4173.69 & 7603.12 & 4100.29 & 14659.47 & 5911.84 & \textbf{4052.97} & 22768.97 & \underline{4083.82} & 5739.85 & 13259.43 & 8405.87 \\
houses & \underline{0.23} & 0.48 & 0.37 & 0.70 & 0.48 & \textbf{0.20} & 0.46 & \textbf{0.20} & 0.39 & 0.44 & \textbf{0.20} & 0.73 & \textbf{0.20} & 0.34 & 0.33 & 0.56 \\
miami\_housing & 96721.82 & 279499.16 & 221239.94 & 395649.83 & 304003.34 & 87855.92 & 253593.27 & 81323.41 & 242931.13 & 260783.83 & \textbf{79390.12} & 2153038.76 & \underline{80753.88} & 199150.39 & 108957.11 & 315039.52 \\
physiochemical\_pro. & 3.89 & 5.98 & 5.15 & 6.79 & 5.99 & 3.53 & 5.98 & \textbf{2.97} & 5.74 & 5.71 & \underline{3.02} & 7.77 & 3.05 & 5.18 & 3.93 & 6.23 \\
superconductivity & 9.90 & 18.10 & 16.74 & - & 19.25 & 9.87 & 21.99 & \underline{9.16} & 25.00 & 18.46 & \textbf{8.94} & 36.42 & 9.20 & 15.13 & 12.37 & 18.55 \\
wine\_quality & 0.65 & 0.73 & 0.69 & 1.12 & 0.74 & 0.60 & 0.82 & \textbf{0.58} & 0.92 & 0.70 & \underline{0.59} & 1.40 & 0.62 & 0.71 & 0.85 & 0.87 \\
\midrule
Mean & 7932.52 & 22319.06 & 17608.64 & 38047.99 & 24117.74 & 7179.57 & 20289.98 & 6665.77 & 20174.26 & 20692.04 & \textbf{6512.20} & 168081.18 & \underline{6619.37} & 15906.29 & 9577.58 & 25179.54 \\
\bottomrule
\end{tabular}%
}
\caption{Tabular Prediction (Section \ref{sec:tabular_prediction}): Regression results per Dataset measured by RMSE (lower is better). * indicates that the results were obtained by training XGBoost on top of row embeddings.}
\label{tab:tabular_prediction_regression}
\end{table*}

\begin{table*}[t]
\centering
\begin{tabular*}{\textwidth}{@{\extracolsep{\fill}} l c c c c c c c c c c @{}}
\toprule
Dataset & GritLM & HyTrel & IBM Granite R2 & MiniLM & SAP-RPT-1 & TABBIE & TaBERT & TabICL v2 & TARTE & TUTA \\
\midrule
NextiaJD & \textbf{0.28} & 0.25 & 0.23 & \underline{0.26} & 0.13 & 0.24 & 0.13 & 0.08 & \textbf{0.28} & 0.18 \\
OpenData & 0.46 & 0.23 & \textbf{0.51} & 0.47 & 0.20 & 0.33 & 0.37 & 0.02 & 0.37 & \underline{0.48} \\
Valentine & \textbf{0.80} & 0.26 & \underline{0.65} & 0.59 & 0.32 & 0.62 & 0.27 & 0.42 & 0.27 & 0.51 \\
WikiJoin-Small & \textbf{0.84} & 0.65 & \underline{0.81} & 0.76 & 0.11 & 0.24 & 0.62 & 0.02 & 0.72 & 0.56 \\
\midrule
Mean & \textbf{0.48} & 0.28 & \underline{0.44} & 0.41 & 0.15 & 0.29 & 0.28 & 0.11 & 0.33 & 0.34 \\
\bottomrule
\end{tabular*}
\caption{Column Similarity Search (Section \ref{sec:col_sim_search}): MAP results per dataset for all approaches.}
\label{tab:col_sim_map}
\end{table*}


\begin{table*}[t]
\centering
\begin{tabular*}{\textwidth}{@{\extracolsep{\fill}} l c c c c c c c c c c @{}}
\toprule
Dataset & GritLM & HyTrel & IBM Granite R2 & MiniLM & SAP-RPT-1 & TABBIE & TaBERT & TabICL v2 & TARTE & TUTA \\
\midrule
GitTables & 0.39 & 0.38 & 0.37 & 0.38 & 0.34 & \underline{0.41} & \textbf{0.46} & 0.02 & 0.34 & 0.40 \\
SOTAB & \textbf{0.81} & 0.52 & \underline{0.75} & 0.73 & 0.57 & \textbf{0.81} & 0.71 & 0.00 & \underline{0.75} & 0.71 \\
\midrule
Mean & \underline{0.60} & 0.45 & 0.56 & 0.55 & 0.46 & \textbf{0.61} & 0.59 & 0.01 & 0.54 & 0.56 \\
\bottomrule
\end{tabular*}
\caption{Column Type Annotation (Section \ref{sec:col_type_annotation}): Macro F1 results per dataset for all approaches.}
\label{tab:cta}
\end{table*}

\begin{table*}[t]
\centering
\begin{tabular*}{\textwidth}{@{\extracolsep{\fill}} l c c c c c c c c @{}}
\toprule
Dataset & GritLM & HyTrel & IBM Granite R2 & MiniLM & SAP-RPT-1 & TABBIE & TaBERT & TUTA \\
\midrule
BIRD Column Schema & \textbf{0.41} & 0.13 & \underline{0.39} & 0.37 & 0.14 & 0.14 & 0.19 & 0.18 \\
\bottomrule
\end{tabular*}
\caption{Schema Linking (Section \ref{sec:col_schema_linking}): MAP results per dataset for all approaches.}
\label{tab:schema_linking_map}
\end{table*}

\begin{table*}[t]
\centering
\begin{tabular*}{\textwidth}{@{\extracolsep{\fill}} l c c c c c c c @{}}
\toprule
Dataset & GritLM & HyTrel & IBM Granite R2 & MiniLM & TABBIE & TaBERT & TUTA \\
\midrule
bird-validation & \underline{0.72} & 0.03 & \textbf{0.72} & 0.57 & 0.02 & 0.10 & 0.06 \\
fetaqa & \textbf{0.61} & 0.00 & \underline{0.51} & 0.28 & 0.00 & 0.01 & 0.01 \\
ottqa & \textbf{0.86} & 0.01 & \underline{0.78} & 0.50 & 0.00 & 0.04 & 0.02 \\
spider-test & \underline{0.66} & 0.01 & \textbf{0.68} & 0.58 & 0.00 & 0.08 & 0.03 \\
spider-train & \underline{0.47} & 0.00 & \textbf{0.50} & 0.40 & 0.00 & 0.03 & 0.01 \\
spider-validation & \underline{0.75} & 0.02 & \textbf{0.82} & 0.70 & 0.01 & 0.13 & 0.05 \\
tabfact & \textbf{0.62} & 0.00 & \underline{0.57} & 0.41 & 0.00 & 0.04 & 0.02 \\
\midrule
Mean & \textbf{0.67} & 0.01 & \underline{0.66} & 0.49 & 0.01 & 0.06 & 0.03 \\
\bottomrule
\end{tabular*}
\caption{Table Retrieval MAP@10 per dataset (markdown serialization). Best per dataset in bold, second-best underlined.}
\label{tab:retrieval_main_map10}
\end{table*}

\begin{table*}[t]
\centering
\begin{tabular*}{\textwidth}{@{\extracolsep{\fill}} l c c c c c c c c c @{}}
\toprule
Dataset & GritLM & HyTrel & IBM Granite R2 & MiniLM & TABBIE & TabDPT & TaBERT & TARTE & TUTA \\
\midrule
CKAN Subset & 0.16 & 0.38 & 0.06 & 0.22 & \underline{0.45} & \textbf{0.69} & 0.10 & 0.37 & 0.14 \\
ECB & \underline{0.00} & \textbf{1.00} & \underline{0.00} & \underline{0.00} & \underline{0.00} & \underline{0.00} & \underline{0.00} & \textbf{1.00} & \underline{0.00} \\
FeTaQA & 0.37 & \textbf{0.87} & 0.23 & 0.28 & 0.64 & 0.54 & 0.26 & \underline{0.82} & 0.10 \\
OTT-QA & 0.28 & \textbf{0.99} & 0.20 & 0.31 & 0.52 & 0.55 & 0.24 & \underline{0.99} & 0.12 \\
Spider & 0.12 & \textbf{1.00} & 0.12 & 0.16 & 0.46 & 0.56 & 0.15 & \underline{0.99} & 0.08 \\
TabFact & 0.16 & \textbf{1.00} & 0.16 & 0.21 & 0.53 & 0.56 & 0.15 & \underline{0.99} & 0.06 \\
\midrule
Mean & 0.18 & \textbf{0.87} & 0.13 & 0.20 & 0.43 & 0.48 & 0.15 & \underline{0.86} & 0.08 \\
\bottomrule
\end{tabular*}
\caption{Table Shuffling accuracy per dataset (v0: hi-pos/hi-neg, both row+col perturbation, default window). Best per dataset in bold, second-best underlined.}
\label{tab:shuffling_accuracy}
\end{table*}

\begin{table*}[t]
\centering
\begin{tabular*}{\textwidth}{@{\extracolsep{\fill}} l c c c c c c c c c c @{}}
\toprule
Dataset & GritLM & HyTrel & IBM Granite R2 & MiniLM & SAP-RPT-1 & TABBIE & TaBERT & TabICL v2 & TARTE & TUTA \\
\midrule
s2abel@dirty & \underline{0.63} & 0.34 & \textbf{0.71} & 0.60 & 0.34 & 0.31 & 0.37 & 0.13 & 0.35 & 0.40 \\
s2abel@clean & 0.66 & 0.34 & \textbf{0.75} & \underline{0.71} & 0.35 & 0.34 & 0.40 & 0.18 & 0.36 & 0.41 \\
\midrule
Mean & 0.64 & 0.34 & \textbf{0.73} & \underline{0.65} & 0.35 & 0.32 & 0.38 & 0.16 & 0.35 & 0.40 \\
\bottomrule
\end{tabular*}
\caption{Cell Level Semantic Retrieval (Section \ref{sec:cell_similarity_search}: MAP per dataset for all approaches.}
\label{tab:cell_results_map}
\end{table*}

\begin{table*}[t]
\centering
\begin{tabular*}{\textwidth}{@{\extracolsep{\fill}} l c c c c c c c c @{}}
\toprule
Dataset & GritLM & HyTrel & IBM Granite R2 & MiniLM & SAP-RPT-1 & TABBIE & TaBERT & TUTA \\
\midrule
BIRD Exact Match & 0.55 & 0.19 & \textbf{0.58} & 0.54 & 0.18 & 0.18 & 0.39 & \underline{0.56} \\
BIRD Fuzzy Match & \textbf{0.75} & 0.13 & \underline{0.67} & 0.58 & 0.16 & 0.14 & 0.40 & 0.45 \\
\midrule
Mean & \textbf{0.65} & 0.16 & \underline{0.62} & 0.56 & 0.17 & 0.16 & 0.39 & 0.51 \\
\bottomrule
\end{tabular*}
\caption{Value Linking (Section \ref{sec:value_linking}): MAP results per dataset for all approaches.}
\label{tab:value_linking_map}
\end{table*}